\documentclass[runningheads]{llncs}

\usepackage{eccv}

\usepackage{eccvabbrv}
\usepackage{graphicx}
\usepackage{booktabs}
\usepackage{tikz}
\usepackage[latin1]{inputenc}
\usepackage{pgfplots}
\usetikzlibrary{backgrounds,scopes}   
\pgfplotsset{compat=1.3}
\usetikzlibrary{shadings,spy,calc}
\usetikzlibrary {arrows.meta} 
\usepackage{bm}
\usepackage{colortbl}
\usepackage{booktabs} 

\usepackage[accsupp]{axessibility}  

\usepackage{hyperref}

\usepackage{orcidlink}

\begin{document}

\title{GeoGaussian: Geometry-aware Gaussian Splatting for Scene Rendering} 

\titlerunning{GeoGaussian}

\author{Yanyan Li\inst{1,2}\orcidlink{0000-0001-7292-9175} 
\and Chenyu Lyu\inst{3}  
\and Yan Di\inst{2} 
\and Guangyao Zhai\inst{2}
\and Gim Hee Lee*\inst{1}\orcidlink{0000-0002-1583-0475}\index{Lee, Gim Hee}
\and Federico Tombari*\inst{2,4}\orcidlink{0000-0001-5598-5212}}

\authorrunning{Y. Li et al.}

\institute{National University of Singapore, Singapore \and 
Technical University of Munich, Germany\and
Tianjin University, China \and
Google, Zurich, Switzerland 
\\ 
\url{https://yanyan-li.github.io/project/gs/geogaussian}}


\renewcommand{\thefootnote}{\fnsymbol{footnote}}
\footnotetext[1]{Equal senior author} 

\maketitle

\begin{abstract}
During the Gaussian Splatting optimization process, the scene 
geometry can gradually deteriorate if its structure is not deliberately preserved, especially in non-textured regions such as walls, ceilings, and furniture surfaces. This degradation significantly affects the rendering quality of novel views that deviate significantly from the viewpoints in the training data. 
To mitigate this issue, we propose a novel approach called GeoGaussian. Based on the smoothly connected areas observed from point clouds, this method introduces a novel pipeline to initialize thin Gaussians aligned with the surfaces, where the characteristic can be transferred to new generations through a carefully designed densification strategy. Finally, the pipeline ensures that the scene 
geometry and texture are maintained through constrained optimization processes with explicit geometry constraints. Benefiting from the proposed architecture, the generative ability of 3D Gaussians is enhanced, especially in structured regions.
Our proposed pipeline achieves state-of-the-art performance in novel view synthesis and geometric reconstruction, as evaluated qualitatively and quantitatively on public datasets.

\vspace{-2mm}
\keywords{Gaussian Splatting \and Geometry-aware Densification \and Geometric Consistency}
\vspace{-2mm}
\end{abstract}

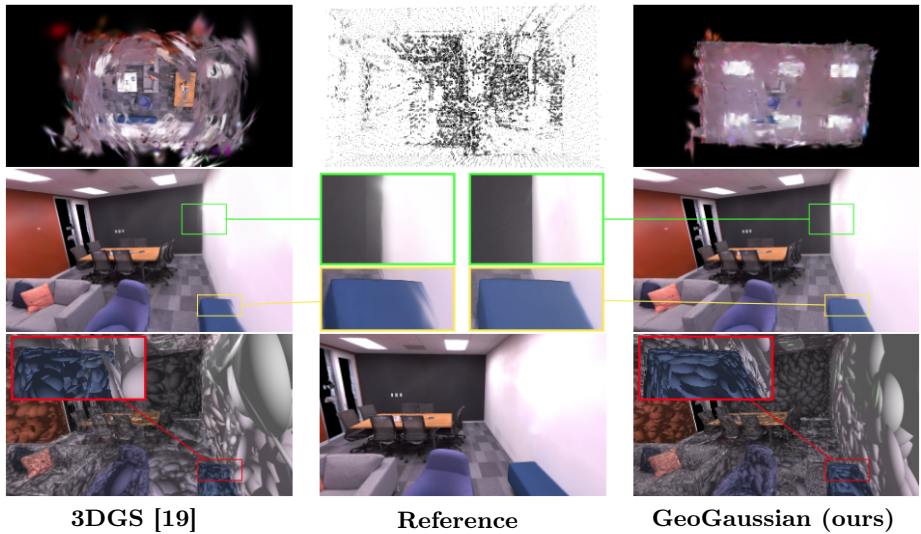
\begin{figure}
\centering
\includegraphics[width=\linewidth]{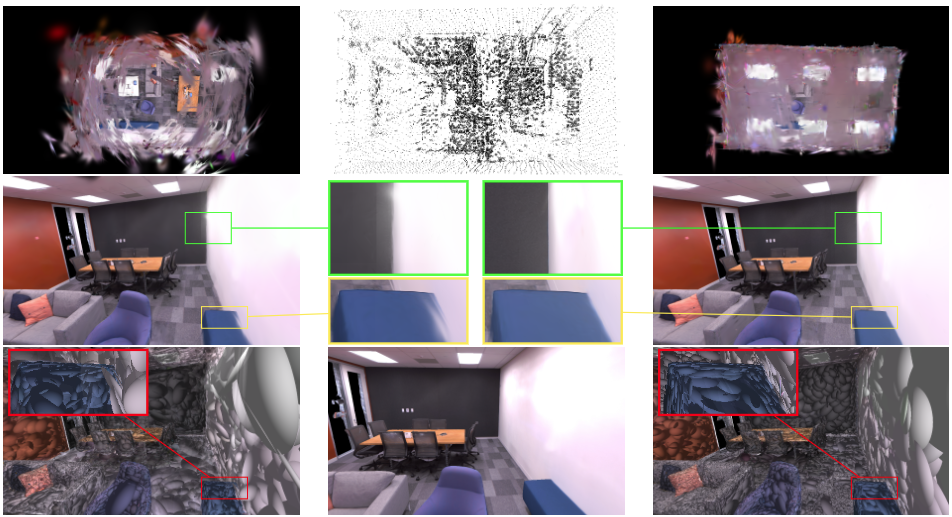}
\begin{tikzpicture}[spy using outlines={red, magnification=3, size=2cm, connect spies}] 
\node[font=\bf] at (10,-3.5) {GeoGaussian (ours)};
\node[font=\bf] at (5.8,-3.5) {Reference};
\node[font=\bf] at (1.5,-3.5) {3DGS~\cite{kerbl3Dgaussians}};
\node[font=\bf] at (0,-3.5) {};
\end{tikzpicture}
\label{fig:teaser}
\caption{
Comparisons of novel view rendering and 3D Gaussian model on the Replica Datasets. As highlighted in the second row, the proposed method shows a very clear boundary between two low-textured walls, but 3DGS has blurring issues since the geometry of its 3D Gaussian model is not accurate in this area.}
\end{figure}

\section{Introduction}
Due to the impressive rendering quality of Neural Radiance Fields (NeRF)~\cite{mildenhall2021nerf}, the area of photo-realistic novel view synthesis (NVS) has become a popular research topic in the communities of computer vision, graphics, and robotics. While NeRFs offer high-quality rendering, 3D Gaussian Splatting~(\cite{kerbl3Dgaussians,fan2023lightgaussian}) shows 
better performance in terms of training speed and rendering quality. 
3D Gaussian Splatting is explicitly represented by a set of Gaussian points parameterized 
by its position, orientation, and spherical harmonics parameters. An additional rasterization step re-projects these 3D Gaussians back to training images to capture scene geometry and appearance by using alpha-blending. However, in the Gaussian Splatting optimization process, the geometry of 3D Gaussian models lacks sufficient constraints \textcolor{black}{especially in low-textured regions}, leading to significant degradation in rendering performance for novel views that deviate substantially from the training data. 

NVS methods typically represent 3D scenes implicitly~\cite{niemeyer2020differentiable,
xu2019disn} or explicitly~\cite{debevec2023modeling,kazhdan2006poisson} based on multiple 2D views and corresponding camera poses.
Previous solutions for NVS tasks have primarily relied on 3D surface reconstruction using technologies such as Structure-from-Motion~\cite{goesele2007multi}, multi-view stereo~\cite{seitz2006comparison}, TSDF~\cite{izadi2011kinectfusion}, Marching Cubes~\cite{lorensen1998marching}, Poisson~\cite{kazhdan2006poisson}, and texturing~\cite{waechter2014let}. While these methods enable the rendering of new views in texture and depth based on 3D surface models, achieving photo-realistic rendering quality remains challenging. \textcolor{black}{Since the impressive achievements of convolutional neural networks are made on many tasks, such as point detection~\cite{detone2018superpoint} and scene completion~\cite{wang2019forknet}}, deep learning-based mesh representations~\cite{feng2019meshnet} have been proposed to enhance view synthesis, while the challenges for these mesh-based methods are to capture accurate geometry and appearance in complex scenarios. Compared to learning-based mesh methods, NeRF~\cite{mildenhall2021nerf} proposes a continuous volumetric function representation using a multi-layer perceptron (MLP), which produces high-quality renderings with impressive details. To address the heavy computational burden and intensive memory consumption, recent improvements have been made using sparse volumes~\cite{liu2020neural}, hash tables~\cite{muller2022instant}, and hierarchical sampling methods~\cite{barron2023zip,wang2023f2}. 
Despite a significant reduction in training time with these methods, improving rendering \textcolor{black}{efficiency} is still a pressing requirement for applications such as SLAM. Recently, 3D Gaussian Splatting (3DGS)~\cite{kerbl3Dgaussians} has gained significant attention in the community, \textcolor{black}{which shows that the rendering speed in high-quality NVS tasks can be made up to 150 FPS~\cite{fan2023lightgaussian}}. In the Gaussian Splatting optimization process, approaches often prioritize image clarity over geometric fidelity. While some viewpoints may render well, issues arising from confusing geometry can impact rendering quality at certain viewpoints. \textcolor{black}{To extract mesh surface from 3D Gaussians, SuGaR~\cite{guedon2023sugar} serves as a refinement module to jointly optimize mesh and Gaussians. However, the method requires more GPU resources for training compared with 3DGS.}


In this paper, we propose a geometry-aware Gaussian Splatting method \textcolor{black}{emphasizing rendering fidelity and geometry structure simultaneously}.  
Initially, normal vectors are extracted from input point clouds, \textcolor{black}{and then smoothly connected areas are detected based on normals. For general regions, we follow the traditional initialization process~\cite{kerbl3Dgaussians} that represents every point as a sphere. 
However, each point is parameterized as a thin ellipsoid with explicit geometric information for smoothly connected areas. 
Specifically, the third value of the scale vector $\mathbf{S}$ is fixed to make the ellipsoid thin. Additionally, the third column of rotation matrix $\mathbf{R}$ decoupled from the covariance matrix $\mathbf{\Sigma}$ is initialized by the normal vector, as shown in Figure~\ref{fig:archi}. Based on the design, we encourage these thin ellipsoids to lie on the surface of smooth regions.} Leveraging the thin ellipsoid representation, our densification strategy \textcolor{black}{containing split and clone steps} operates within a well-constrained process. On the one hand, a large Gaussian point can be split into two \textcolor{black}{smaller co-planar Gaussians lying on the plane established by the position and normal vectors of the large Gaussian}, whereas traditional approaches~\cite{fan2023lightgaussian,kerbl3Dgaussians,Matsuki:Murai:etal:CVPR2024} randomly position new Gaussians within the ellipsoid. On the other hand, our Gaussian clone approach ensures that new generations \textcolor{black}{have to lie on the same plane with the original since the new Gaussian is also required to align with the surface of the smooth region. The step is supported by accumulating the gradient 
descent direction of the origin's position $\bm{\mu}$, and then the component of the direction that is perpendicular to the normal vector of origin is decoupled to guide the position for the cloned Gaussian.}  
Through the learning rate and direction, the Gaussian map undergoes densification for continuous training. \textcolor{black}{In the optimization module, we propose a new geometrically consistent constraint for thin ellipsoids lying on the smooth areas by encouraging the nearest neighbors to be co-planar, which are jointly optimized with the widely used photometric residuals~\cite{kerbl3Dgaussians} in an iterative process.}
The contributions of the paper are summarized as: 
\begin{itemize}
    \item \textcolor{black}{A parameterization with explicit geometry meaning for thin 3D Gaussians is employed in our carefully designed initialization and densification strategies} to establish reasonable 3D Gaussian models.  
    \item \textcolor{black}{A geometrically consistent constraint is proposed to encourage thin Gaussians to align with the smooth surfaces.}
    \item \textcolor{black}{Evaluations on the public datasets demonstrate that the proposed method improves rendering quality compared to state-of-the-art Gaussian Splatting methods.}
\end{itemize}

\section{Related Work}

\subsection{Map Reconstruction and Texturing}

Traditional approaches to view synthesis typically involve 3D model reconstruction and texture mapping modules. 3D surface reconstruction can be achieved using algorithm modules such as structure-from-motion~\cite{westoby2012structure}, simultaneous localization and mapping~\cite{kerl2013dense}, and depth map fusion~\cite{izadi2011kinectfusion,dai2017bundlefusion}. After obtaining surface meshes based on methods like Marching Cubes~\cite{lorensen1998marching} and Poisson surface reconstruction~\cite{kazhdan2006poisson}, another crucial step is to refine the texture of triangle faces based on input visual images~\cite{lempitsky2007seamless} or through a form of blending~\cite{allene2008seamless} between these images. Learning-based solutions for 3D reconstruction~\cite{mescheder2019occupancy,gao2020learning} and texturing~\cite{dong2018learning,oechsle2019texture} tasks have also been proposed, representing high-frequency texture and employing techniques for natural blending.


\subsection{Neural Rendering and Radiance Fields}
Instead of explicitly representing scenes, NeRF~\cite{mildenhall2021nerf} records environments using MLP neural networks, opening new possibilities for high-quality view synthesis based on volumetric ray tracing. However, limited by inference speed and convergence difficulties, explicit structures such as voxels~\cite{liu2020neural} or deep features are incorporated into ray-based representations to accelerate rendering. Relu fields~\cite{karnewar2022relu} investigate grid-based representations to maintain the high-fidelity rendering performances of MLPs while speeding up the reconstruction and reference processes through fixed non-linearity on interpolated grid parameters. Instant-NGP~\cite{muller2022instant} proposes a multi-resolution hash table of trainable feature vectors to improve neural networks for the same speeding-up goal of neural graphics primitives. Despite the significant improvements achieved by these methods~\cite{liu2020neural,rebain2021derf,deng2020jaxnerf}, current approaches may not fully satisfy the requirements of real-time rendering applications.

\begin{figure}[htbp]
\centering
\resizebox{\linewidth}{!}{
\begin{tikzpicture}[spy using outlines={red,magnification=3,size=2cm}, connect spies]
  \node (img1) at (0,-1.8){
    \includegraphics[width=0.3\linewidth]{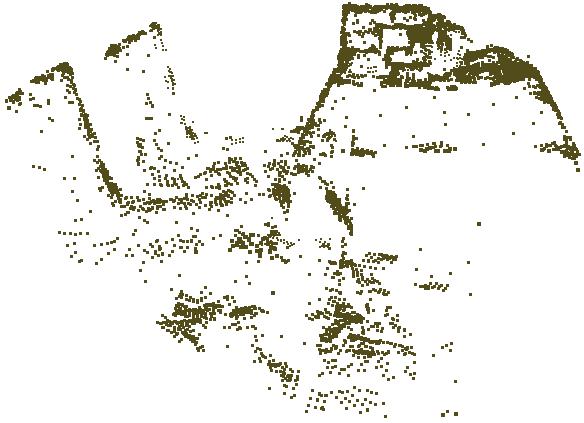}};
 \node (img2) at (-0.5,-3.5){
    \includegraphics[width=0.13\linewidth]{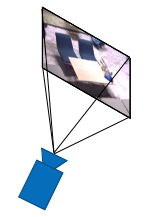}};  
    \spy[green, height = 0.5cm, width = 0.5cm] on (0.55,-1.2) in node at (2.0,-1.2);
    \begin{scope} 
    \fill[blue, opacity=0.1] (-1.8,-4.6) rectangle (4.9,0);
    \end{scope}
    \draw (2.2, -0.4) node {smooth area};
    \draw [->] (2.0,-1.2) to [out=-30,in=-100] (2.2, -0.6);
    \node [right] at (4.5,-1.5) {$y$};
    \node [above] at (3.5,-0.5) {$z$};
    \node [above] at (2.8, -2.0) {$x$};
    \draw[line width=1pt,-latex]  (3.5,-1.5) -- (4.5,-1.5);
    \draw[line width=1pt,-latex]  (3.5,-1.5) -- (3.5,-0.5);
    \draw[line width=1pt,-latex]  (3.5,-1.5) -- (2.8, -2.0);
    \shade[right color=orange,middle color=red,left color=blue,opacity=0.4,shading angle=-110] (3.5,-1.5) circle (1.3 and 0.5);
    \shade[ball color=orange,opacity=0.3] (3.5,-1.5) circle (1.3 and 0.5);
    \draw[dashed] (3.5,-1.5) ellipse[x radius=1.3/9, y radius=0.5];
    \draw [arrows = {-Computer Modern Rightarrow[line cap=butt]},line width=1mm] (1.5,-2.1) -- (2.0,-2.1);
    \node[draw,align=left] at (2.8,-3.5) {3D Position: $\bm{\mu}$\\ Scales: $\mathbf{S} = [s_1\;s_2\;0.001]^T$ \\
    Rotation: $\mathbf{R} = [\mathbf{r}_1 \; \mathbf{r}_2 \; \mathbf{n}]$ \\
    Spherical Harmonics: $\mathbf{C}$ \\ Opacity: $\alpha$};
    \draw (4.1/2,-5) node {\textbf{Parametrization of Thin Gaussians}};
    \draw [arrows = {-Computer Modern Rightarrow[line cap=butt]},line width=1mm] (5,-2.1) -- (5.5,-2.1);
    \node (img3) at (9-1.5,-1.5){
    \includegraphics[width=0.3\linewidth]{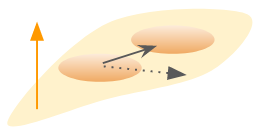}};
    \begin{scope} 
    \fill[green, opacity=0.1] (4.9,0) rectangle (10.3,-4.6);
    \end{scope}
    \draw (9-1.5, -2.5) node {\textbf{Clone on Tangent space}};
    \draw (8.5-1.5, -0.8) node[anchor=south east]  {\textbf{normal}};
    \draw[text width=2cm] (11-1.5, -2.0) node  {\small accumulated direction};
    \draw[text width=2cm] (10.5-1.5, -0.6) node {\small modified \\ direction};
    \shade[right color=orange,middle color=red,left color=blue,opacity=0.4,shading angle=-110] (8-1.5,-3.5) circle (1 and 0.2);
    \shade[ball color=orange,opacity=0.3] (8-1.5,-3.5) circle (1 and 0.2);
    \draw[dashed] (8-1.5,-3.5) ellipse[x radius=1/14, y radius=0.2]; 
    \coordinate (O) at (8-1.5,-3.5);
    \filldraw (O) circle (.5pt);
    \shade[right color=orange,middle color=red,left color=blue,opacity=0.4,shading angle=-110] (10-1.5,-3.5) circle (0.5 and 0.2);
    \draw[dashed] (10-1.5,-3.5) ellipse[x radius=0.5/14, y radius=0.2];
     \shade[right color=orange,middle color=red,left color=blue,opacity=0.4,shading angle=-110] (10+1-1.5,-3.5) circle (0.5 and 0.2);
    \draw[dashed] (10+1-1.5,-3.5) ellipse[x radius=0.5/14, y radius=0.2];
    \coordinate (x1) at (10.15-1.5,-2.8-0.3);
	\coordinate (x2) at (11.4-1.5,-2.8-0.3);
	\coordinate (x3) at (10.8-1.5, -3.4-0.5 );
	\coordinate (x4) at (9.45-1.5, -3.4-0.5);
    \node[shift={(-0.15,0.1)}] at (x1) {$\pi$};
    \path[draw=black, fill=black!20, thick, opacity = 0.4] (x1) -- (x2) -- (x3) -- (x4) -- (x1);
    \coordinate (O1) at (11-1.5,-3.5);
    \filldraw (O1) circle (.5pt) node {};
    \coordinate (O2) at (10-1.5,-3.5);
    \filldraw (O2) circle (.5pt) node{};
    \draw (9-1.5, -4.3) node {\textbf{Split on Tangent space}};
    \draw [->] (8-1.5,-3.7) to [out=-30,in=-170] (11-1.6,-3.7);
    \draw [->] (8-1.5,-3.7) to [out=-30,in=-150] (10-1.6,-3.7);
    \draw [arrows = {-Computer Modern Rightarrow[line cap=butt]},line width=1mm]
        (10.5,-2.1) -- (11,-2.1);
    \draw (4.9+5.4/2,-5) node {\textbf{Densification}};
    \node at (13,-2.3) {\includegraphics[width=0.3\linewidth]{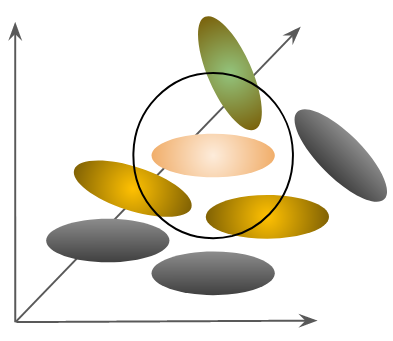}};
    \node [right] at (10.7,-1.1) {$Y$};
    \node [above] at (14.3,-1.2) {$Z$};
    \node [above] at (14.3,-3.8) {$X$};
    \draw[text width=2cm] (12.5,-1.0) node {\small nearest \\ neighbors};
    \draw [->] (12.5,-1.3) to [out=-30,in=-170] (12.9,-1.6);
    \draw (13, -4.2) node {\textbf{Co-planar Constraint}};
    \begin{scope}  
    \fill[red, opacity=0.1] (10.3,-4.6) rectangle (15,0);
    \end{scope}
    \draw (10.3+4.7/2,-5) node {\textbf{Smooth Constraint}};
\end{tikzpicture}}
    \caption{Geometry-aware strategies of our GeoGaussian. In smoothly connected areas, the \textbf{parameterization of thin Gaussians} contains clear geometry meanings in the mean vector and covariance matrix. Furthermore, the \textbf{densification} operation for these thin Gaussians encourages the new generations to lie in the tangent space established by the position and normal vectors of the original Gaussian. Finally, these thin Gaussians, measured by a training view, are used to establish \textbf{smooth constraints} with photometric constraints in the optimization process.}
    \label{fig:archi}
\end{figure}
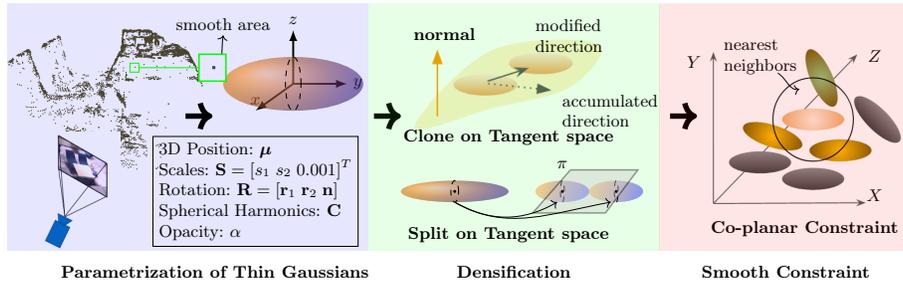

\subsection{Gaussian Splatting in Rendering and Reconstruction}
Recently, 3DGS~\cite{kerbl3Dgaussians}, combining point-based splatting and blending techniques for rendering, has achieved real-time speed with photo-realistic rendering quality. To make the representation more compact, LightGS~\cite{fan2023lightgaussian} proposes a strategy to remove unimportant Gaussians based on 
the volume and opacity of each Gaussian. Building on the results of 3DGS, SuGaR~\cite{guedon2023sugar} extracts mesh faces from unorganized 3D Gaussian maps, serving as an optional refinement module to jointly optimize surface and Gaussians. However, this method requires significantly more GPU resources compared to 3DGS. \textcolor{black}{In contrast to the architecture of Colmap+GS solutions}, 3D Gaussians working as the only representation are used in incremental tracking and mapping systems~\cite{yugay2023gaussian, Matsuki:Murai:etal:CVPR2024,keetha2023splatam}. Gaussian-SLAM~\cite{yugay2023gaussian} selects every 5th frame as a keyframe for training 3D Gaussians based on RGB-D images.
Compared with 3DGS and related refinements, our method focuses on modeling geometry and achieving photo-realistic rendering performance through redesigned parameterization and densification processes, as well as geometry-aware constraints.

\section{Our Methodology}
\subsection{3D Gaussian Splatting}
In the widely used 3D Gaussian Splatting representation, the scene is built by 
a set of anisotropic Gaussians. 
Each Gaussian point is characterized by attributes including the mean $\bm{\mu}_w^i$ and covariance $\Sigma_w^i$ in the world coordinate frame, where the mean vector and covariance matrix also represent the position and shape of the $i^{th}$ Gaussian ellipsoid $\mathcal{G}^i$. 
To ensure semi-positive definiteness, the covariance matrix 
is represented by diagonal scaling matrix $\mathbf{S}_w^i= Diag[s_1^i\;s_2^i\;s_3^i] 
\in \mathbb{R}^{3\times 3}$ and rotation matrix $\mathbf{R}_w^i=[\mathbf{r}_1^i\;\mathbf{r}_2^i\;\mathbf{r}_3^i] 
\in \operatorname{SO}(3)$ 
as follows:
\begin{equation}
    \Sigma_w^i = \mathbf{R}_w^i\mathbf{S}_w^i{\mathbf{S}_w^i}^{\intercal}{\mathbf{R}_w^i}^{\intercal},
\end{equation}
where $\operatorname{SO}$(3) is the special orthogonal group.

In addition to the position and shape parameters, spherical harmonics coefficients $\mathbf{C}_w^i \in \mathbb{R}^{(m+1)^2\times 3}$, where $m$ is the degrees of the freedom, and opacity $\alpha^i_w \in \mathbb{R}$ also 
play important roles in 
rendering the colored image. The color of the target pixel can be synthesized by splatting and blending these $N$ organized Gaussian points that have overlaps with the pixel. First, the splatting operation is to form 2D Gaussians $\mathcal{N}(\mu_I, \Sigma_I )$ on the image planes from the 3D Gaussians $\mathcal{N}(\mu_w, \Sigma_w )$ in the world coordinates based on camera poses, \ie: 
\begin{equation}
     \mu_{I} =\Pi(\mathbf{T}_{cw} \mu_w) ,\;
    \Sigma_{I} = \mathbf{J}\mathbf{W}_{cw}\Sigma_{w}\mathbf{W}_{cw}^{\intercal}\mathbf{J}^{\intercal},
\end{equation}
where  $\mathbf{T}_{cw} = \left[\begin{array}{cc}
      \mathbf{W}_{cw}& \mathbf{t}_{cw}  \\
      \mathbf{0}& 1 
 \end{array}\right] 
 \in \operatorname{SE}$(3) is the camera pose from the world to the camera coordinate in the special Euclidean group, and $\mathbf{W}_{cw}$ and $\mathbf{t}_{cw}$ are rotational and translational components, respectively. $\mathbf{J}$ is the Jacobian matrix of the projective transformation. 
 The blending operation is then given as follows:
\begin{equation}
    \mathcal{C}_{\mathbf{p}}=\sum_{i\in N} c^i\alpha_w^i \prod_{j=1}^{i-1}(1-\alpha_w^i),
\end{equation}
where $c_i$ and $\alpha_w^i$ represent the color and opacity of the point, and the 3D Gaussian Splatting and blending operations are differentiable. Based on optimization solvers, Gaussian parameters can be trained gradually and supervised via photometric residuals.

\subsection{Gaussian Initialization and Densification}
\noindent \textbf{Initialization.} Surface normals are first extracted from point clouds using a method~\cite{muja2009fast}, wherein unreliable normals are detected based on two rules: 1) points with long distances from others, and 2) points whose neighbors within a distance threshold have different normals with large angles with the point. This leads to the creation of two groups of points: $\mathcal{S}_{co}$ and $\mathcal{S}_{ind}$, where $\mathcal{S}_{co}$ contains points located on smooth surfaces while $\mathcal{S}_{ind}$ comprises more individual points.

These sparse points are then fed to the Gaussian Initialization module. For Gaussians initialized by $\mathcal{S}_{ind}$, we adopt an approach similar to that used in 3DGS 
by using the distance between the point and its nearest neighbor to set the scale. For Gaussians $\mathcal{G}$ related to $\mathcal{S}_{co}$, we encode more geometric information in the parameters as shown in Figure~\ref{fig:archi}.
Specifically, the scale vector has only 2 DoFs since the value in the z-axis $s_3$ (in the Gaussian coordinate frame) is fixed to a small value to create a thin ellipsoid 
aligned with the surfaces. 
Additionally, we assume that the third column $\mathbf{r}_3$ of the rotation matrix 
from the covariance matrix $\Sigma$ represents the normal 
while the other two columns are set as a group of normalized basis vectors perpendicular to $\mathbf{r}_3$.

\vspace{3mm}
\noindent \textbf{Densification in the tangential space.}
The general Gaussians initialized by points from $\mathcal{S}_{ind}$ are densified in the traditional manner, which involves splitting a large ellipsoid into two smaller ones located randomly inside the ellipsoid. However, the approach differs for thin Gaussians denoted as $\mathcal{G}_{\theta}$. 
We do the following to ensure that the new clones and splits are aligned with the surface: 
\textbf{1) \textit{Cloning.}} The gradient of the position is calculated over 10 iterations~\cite{kerbl3Dgaussians}. 
The position $\bm{\mu}^{i+1}$ of the new Gaussian is obtained when the accumulated gradient $(\delta \bm{\mu}^{i})$ exceeds the threshold $\gamma$.
However, instead of being set along the direction of the gradient, the new position of the clone Gaussian is determined as follows:
\begin{equation}
\bm{\mu}^{i+1} = \bm{\mu}^{i} + \delta \bm{\mu}^{i} - \mathbf{r}_3^\intercal \delta \bm{\mu}^{i},
\end{equation}
where $\bm{\mu}^{i}$ is the position of the original Gaussian $\mathcal{G}_{\theta}^{i}$. Additionally, parameters such as $\mathbf{r}_3$ are also cloned to the new Gaussian. 
\textbf{2) \textit{Splitting.}} The split process also follows the rule of being co-planar. Due to our thin Gaussian representation, the new position $\bm{\mu}^{i+1}$ lies on the plane established by the normal and position vectors of $\mathcal{G}_{\theta}^{i}$, satisfying the equation $[\bm{\mu}^{i+1} \mid {\bm{\mu}^{i+1}}^\intercal\mathbf{r}_3^{i+1} = {\bm{\mu}^{i}}^\intercal\mathbf{r}_3^{i},~ \mathcal{G}_{\theta}^{i+1} \in \mathcal{S}_{co}]$. Based on this method, the Gaussian clone occurs in the tangential space near the original Gaussian point, and the split operation also ensures that the new Gaussians lie in the tangential space.

\subsection{ View-dependent Optimization} 
The iterative optimization process is necessary to ensure that these Gaussians have the ability to render photo-realistic images at given viewpoints. Following the strategy of 3DGS~\cite{kerbl3Dgaussians}, the goal of our designed loss functions is to create correct geometry and adjust incorrectly positioned Gaussians. The photometric residual between the rendered pixel $c_i$ and the ground truth pixel $\bar{c}_i$ is represented as:
\begin{equation}
    E_{pho}^i = \| c_i - \bar{c}_i  \|
    = \| f(\mathcal{G}, \mathbf{T}_{cw}) - \bar{c}_i  \|,
\end{equation}
where $f(\mathcal{G}, \mathbf{T}_{cw})$ rasterizes and blends the relevant Gaussians (regular and thin ellipsoids) to produce the color.

After selecting view-dependent thin Gaussians $\mathcal{G}_{\theta}$ which are encouraged to preserve smooth connections with their nearest neighbors, we propose a smoothness loss function to further improve the geometry of the model. 
Only Gaussians detected by the current view are considered in this function. We then utilize the K-NN algorithm to detect the eight nearest neighbors around $\mathcal{G}_{\theta}$, which are passed through a filter to remove outliers that have large angle differences by comparing their normal vectors with that of $\mathcal{G}_{\theta}$. 
Consequently, the proposed smooth constrain for optimizing position and normal vectors can be represented as:
\begin{equation}
   E_{geo} = 
        \beta_1\| \mathbf{r}_3^\intercal\bm{\mu} - \frac{1}{m}\sum_{j\in m} {\mathbf{r}_3^j}^\intercal\bm{\mu}^j \| \\ +
        \beta_2\| \mathbf{r}_3^j - \mathbf{r}_3 \|_{2},
\end{equation}
where $\beta_1$ and $\beta_2$ are weights, and $\mathbf{r}_3^\intercal\bm{\mu}$ shows the distance between the origin of the world coordinate and the plane established by normal $\mathbf{r}_3$ and position $\bm{\mu}$ vectors of $\mathcal{G}_{\theta}$, while $\frac{1}{m}\sum_{j\in m} {\mathbf{r}_3^j}^{\intercal}\bm{\mu}^j$ shows the average point-plane distance of the neighborhood. 

By jointly optimizing the normal alignment and depth consistency of the Gaussian point tangent space in the neighborhood, we can obtain Gaussian points with a smooth distribution and alignment on the scene surface.

Finally, the loss function $L$ combines both photometric $E_{pho}$ and geometric $E_{geo}$ residuals as: 
\begin{equation}
  L = \lambda_1 \sum_{\forall k \in H\times{W}} E_{pho}^k + \lambda_2\sum_{\mathcal{G}_{\theta}^i} E_{geo}^i,
\end{equation}
where $\lambda_1$ and $\lambda_2$ are hyper-parameters, and $H$ and $W$ represent the height and width of the image, respectively. 
$\mathcal{G}_{\theta}^i$ is detected by the image.

\section{Experiments}

\subsection{Implementation Details}
In this section, we detail the experimental settings. The proposed GeoGaussian approach is trained and evaluated on a desktop PC equipped with an Intel Core i9 12900K 3.50GHz processor and a single GeForce RTX 3090 GPU. Throughout all experiments, we maintain a consistent learning rate of 0.0002 for Gaussian optimization and $\gamma$ is set to $0.0002$. We also use a photometric loss weighting of $0.8$ and a geometric loss weighting of $0.3$, while weights $\beta_1$ and $\beta_2$ are $0.05$ and $0.01$, respectively. 

In the evaluation phase, we employ a standard procedure. 
Specifically, we follow 3DGS~\cite{kerbl3Dgaussians} to train our models on public datasets for 30,000 iterations. 
During the first 2,000 iterations, only photometric constraints are used in the optimization process. Densification operations cease after 10,000 iterations. Additionally, Gaussians with small opacity values (under 0.05) are removed from the map. The list of nearest neighbors is updated every 100 iterations before reaching 20,000 iterations. After 20,000 iterations, the update frequency is set to 1,000 iterations 
as no more Gaussians are added to the map at that time.

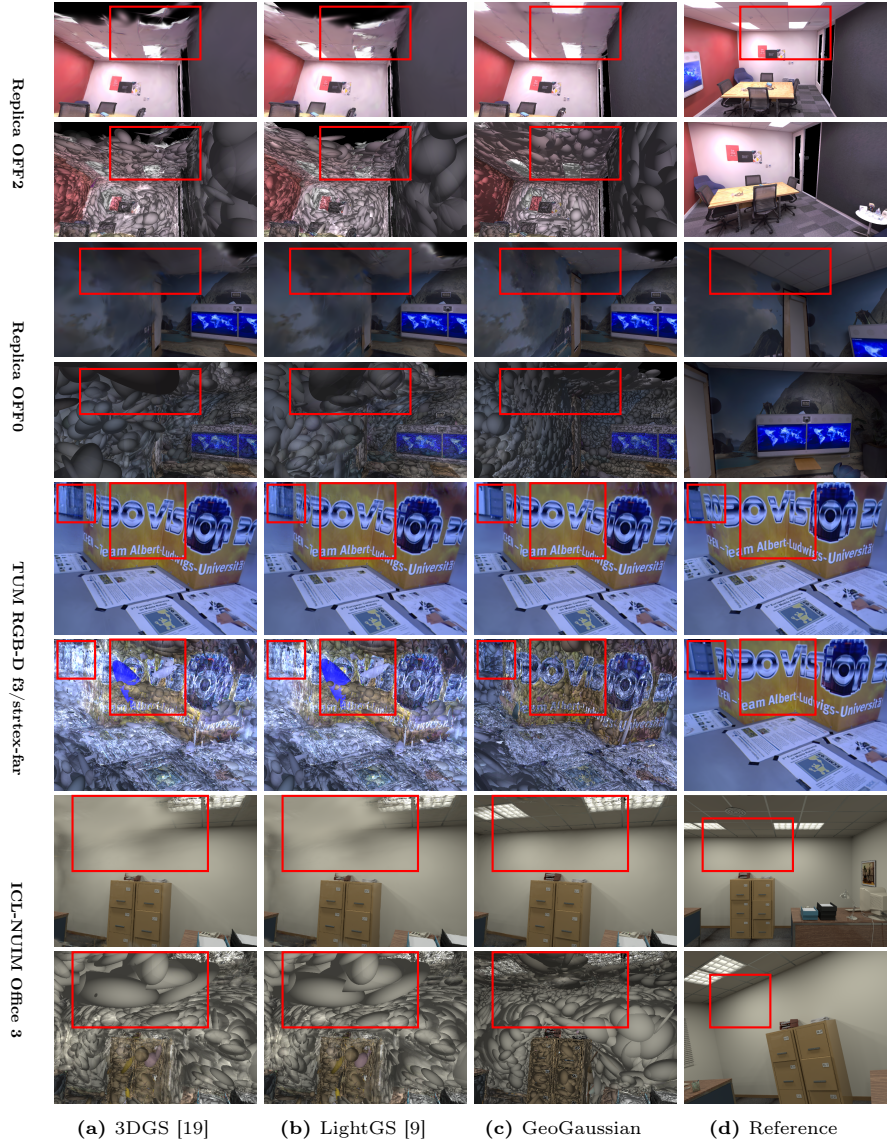
\begin{figure} 
	\centering

 \resizebox{0.034\linewidth}{!}{
     	\begin{tikzpicture}
        \node[rotate=270,font=\bf] at (0, 0) {};
        \node[rotate=270,font=\bf] at (0, 3) {ICL-NUIM Office 3};
        \node[rotate=270,font=\bf] at (0, 9) {TUM RGB-D f3/strtex-far};
        \node[rotate=270,font=\bf] at (0, 15) {Replica OFF0};
        \node[rotate=270,font=\bf] at (0, 20) {Replica OFF2};
    	\end{tikzpicture}}
\subfloat[3DGS~\cite{kerbl3Dgaussians}]{
 \begin{minipage}[b]{0.22\textwidth}
 \begin{tikzpicture}
     \node (img1) at (0,1.6*7) {\includegraphics[width=\linewidth]{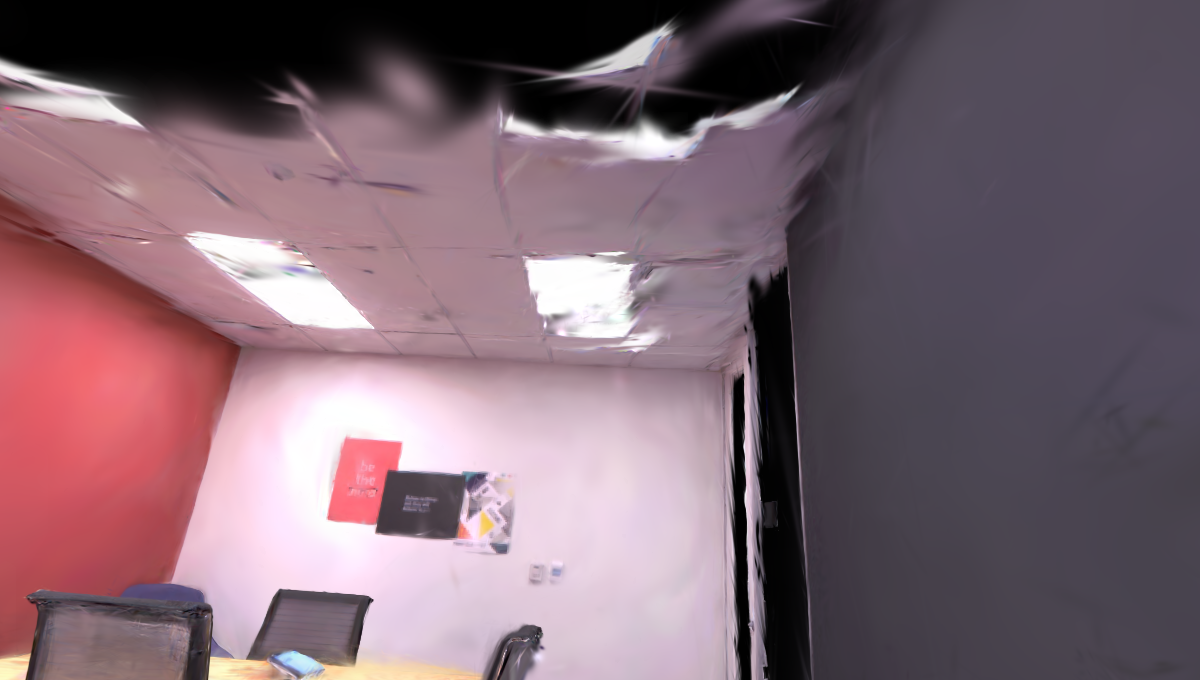}};
     \node (img1) at (0,1.6*6) {\includegraphics[width=\linewidth]{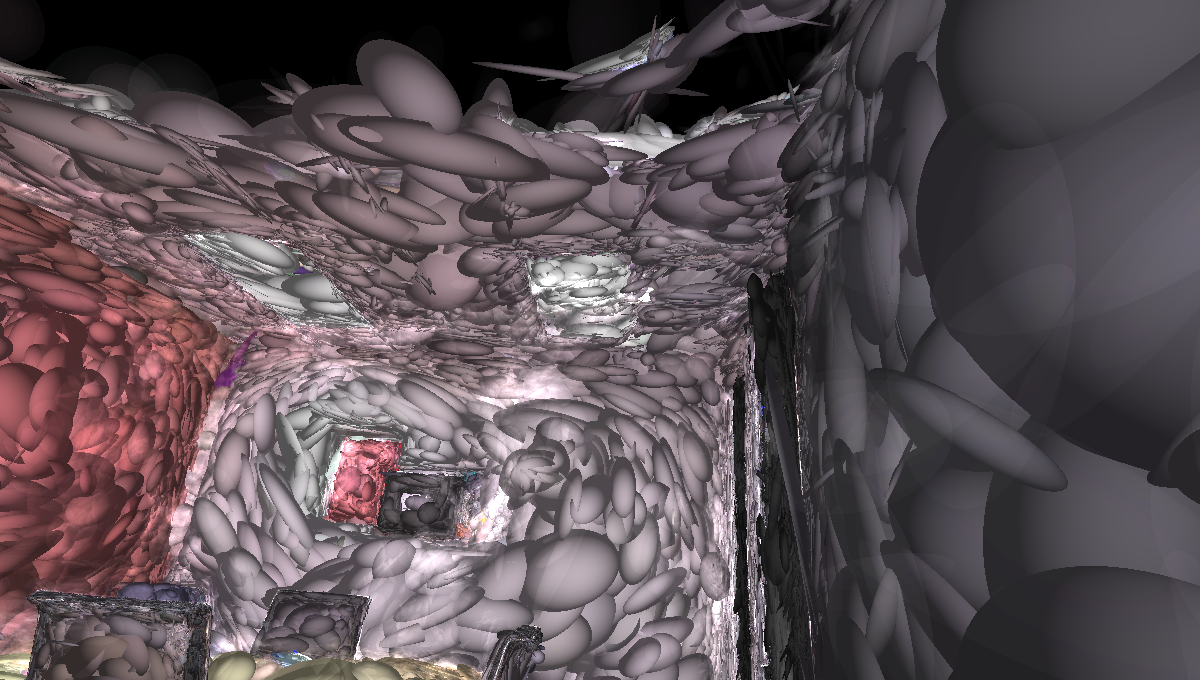}}; 
     \node (img1) at (0,1.6*5) {\includegraphics[width=\linewidth]{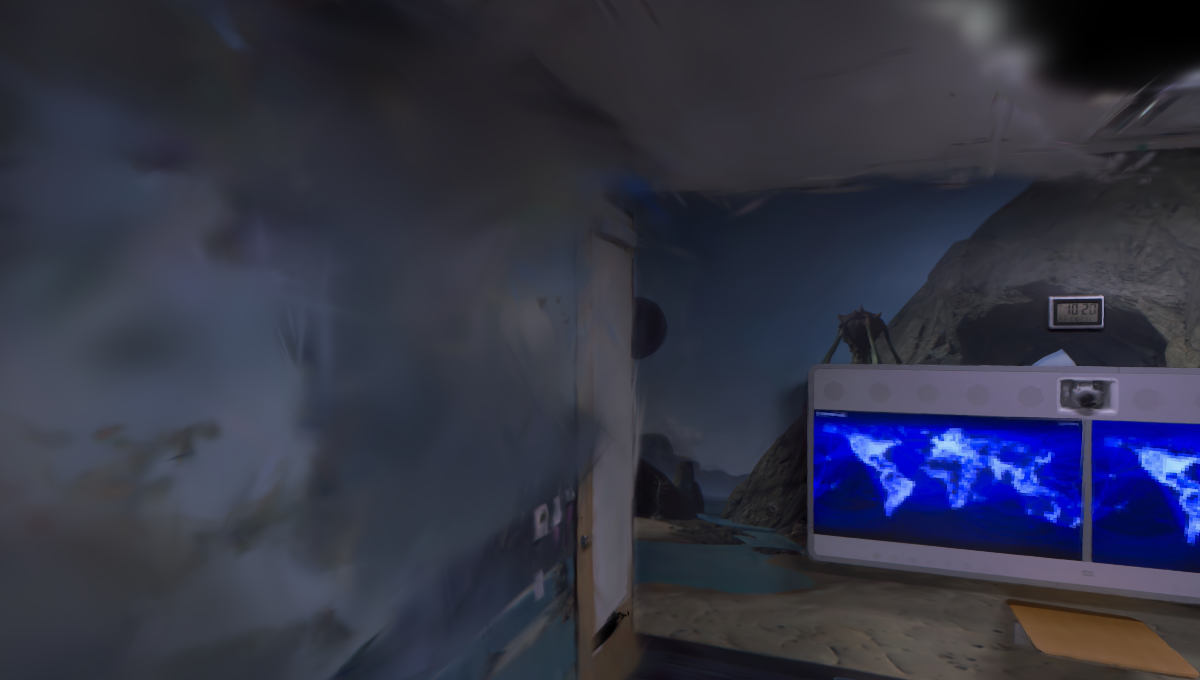}};
     \node (img1) at (0,1.6*4) {\includegraphics[width=\linewidth]{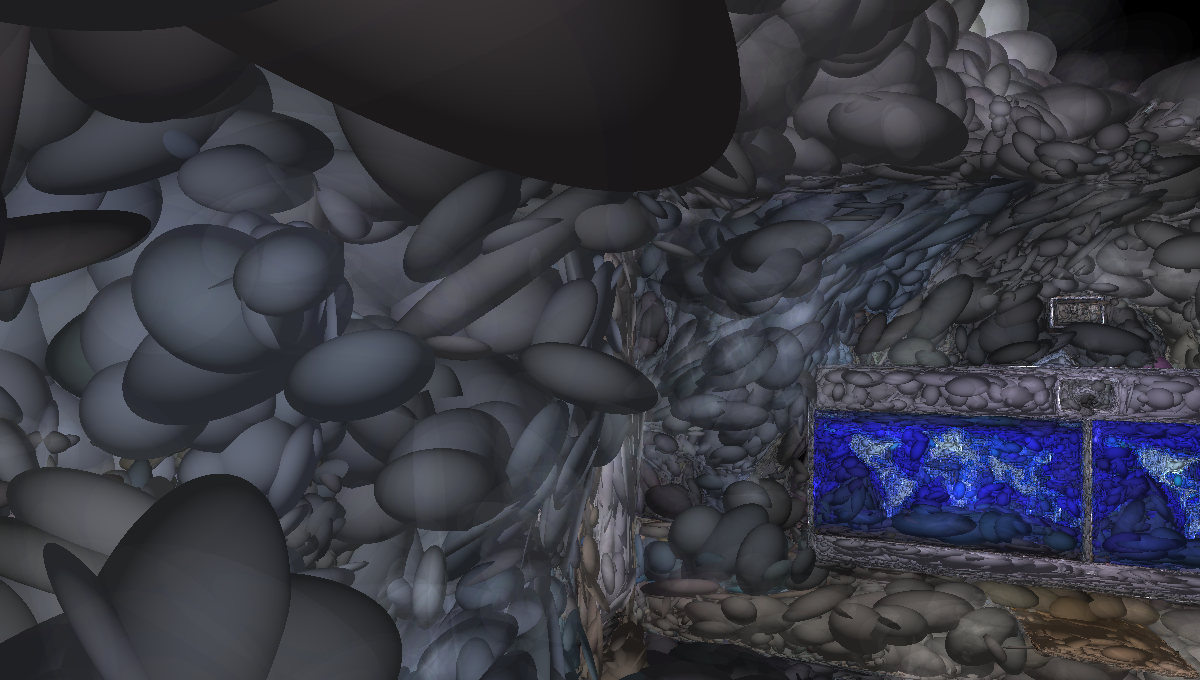}};
    \node(img5) at (0,1.6*2.85){\includegraphics[width=\linewidth]{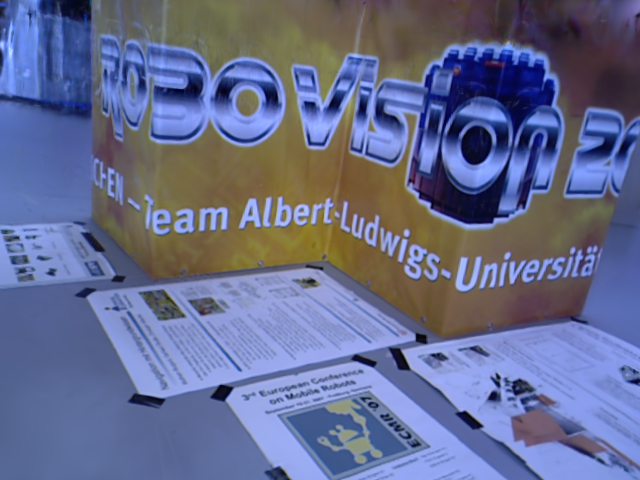}};
    \node(img6) at (0,1.6*1.55){\includegraphics[width=\linewidth]{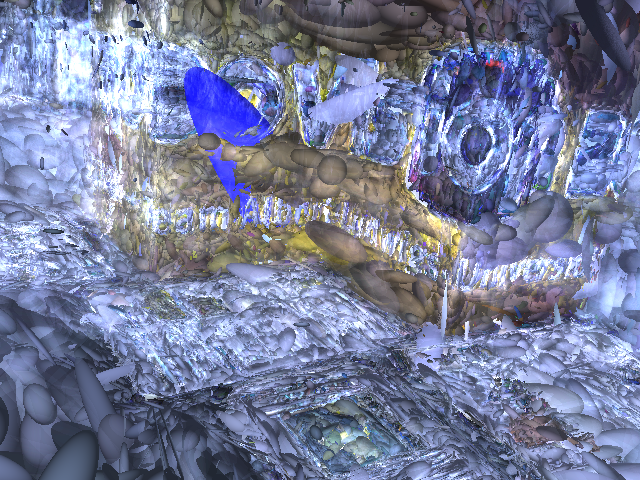}}; 
     \node (img1) at (0,1.6*0.25) {\includegraphics[width=\linewidth]{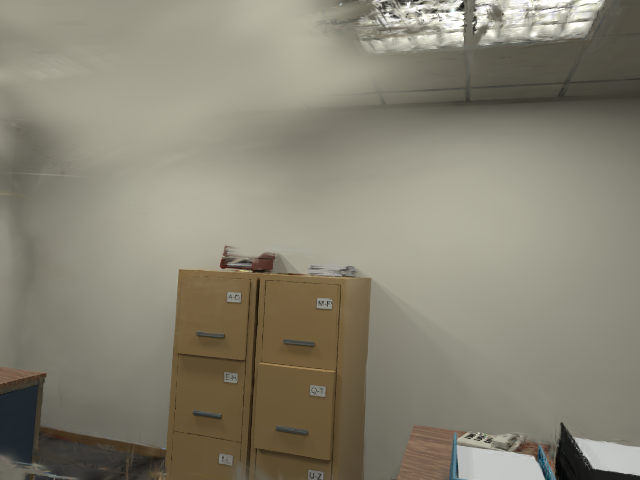}};
     \node (img1) at (0,-1.6*1.05) {\includegraphics[width=\linewidth]{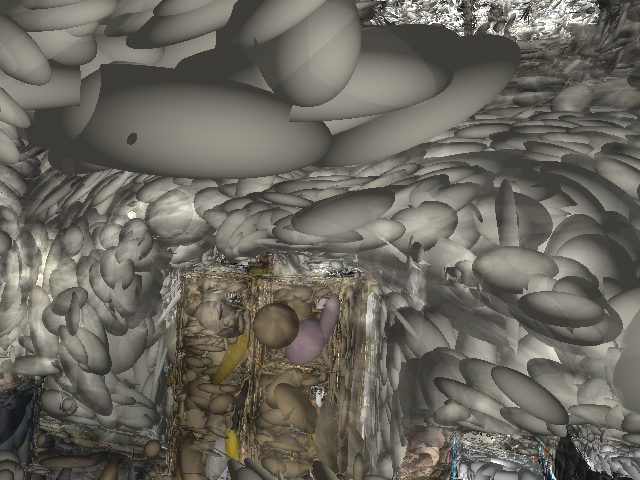}};

    \draw[red, thick] (-0.6,1.6*7) rectangle +(1.2,0.7);
    \draw[red, thick] (-0.6,1.6*6) rectangle +(1.2,0.7);

    \draw[red, thick] (-1,1.6*5.05) rectangle +(1.6,0.6);
    \draw[red, thick] (-1,1.6*4.05) rectangle +(1.6,0.6);
    
    \draw[red, thick] (-1.3,1.6*3.15) rectangle +(0.5,0.5);
    \draw[red, thick] (-1.3,1.6*1.85) rectangle +(0.5,0.5);
    \draw[red, thick] (-0.6,1.6*2.85) rectangle +(1,1);
    \draw[red, thick] (-0.6,1.6*1.55) rectangle +(1,1);
    
    \draw[red, thick] (-1.1,1.6*0.25) rectangle +(1.8,1);
    \draw[red, thick] (-1.1,-1.6*1.05) rectangle +(1.8,1);
   
 \end{tikzpicture}
\end{minipage}}
 \subfloat[LightGS~\cite{fan2023lightgaussian}]{
 \begin{minipage}[b]{0.22\textwidth}
 \begin{tikzpicture}
     
     \node (img1) at (0,1.6*7) {\includegraphics[width=\linewidth]{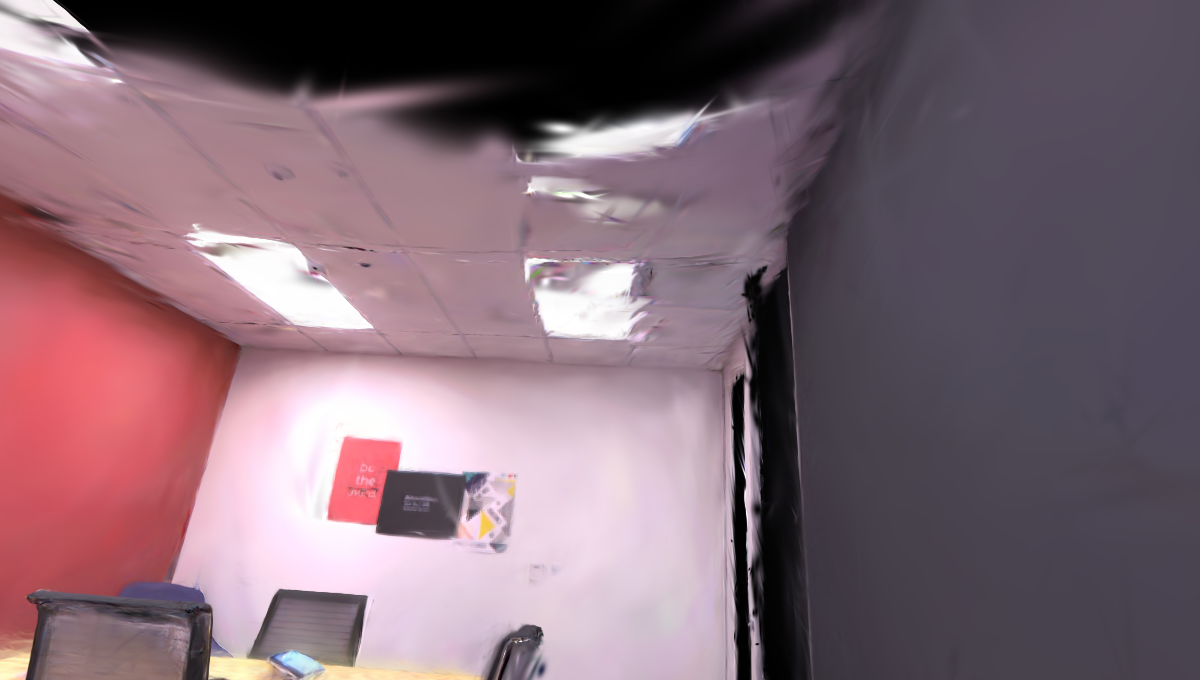}};
     \node (img1) at (0,1.6*6) {\includegraphics[width=\linewidth]{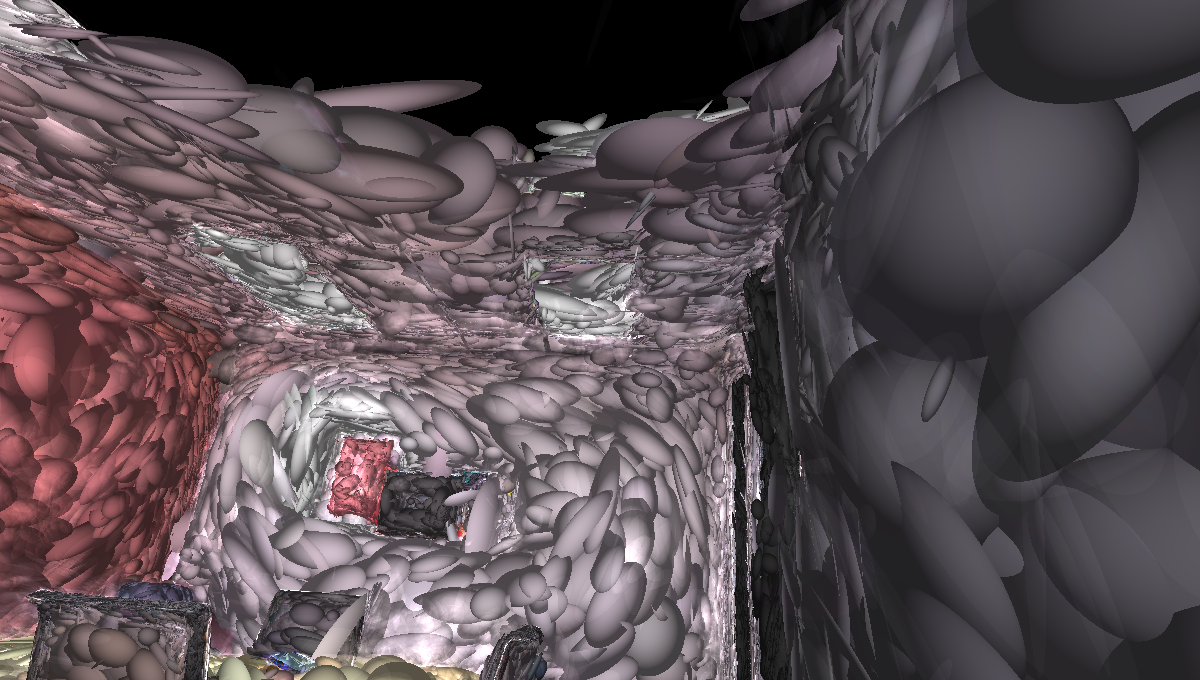}}; 
     \node (img1) at (0,1.6*5) {\includegraphics[width=\linewidth]{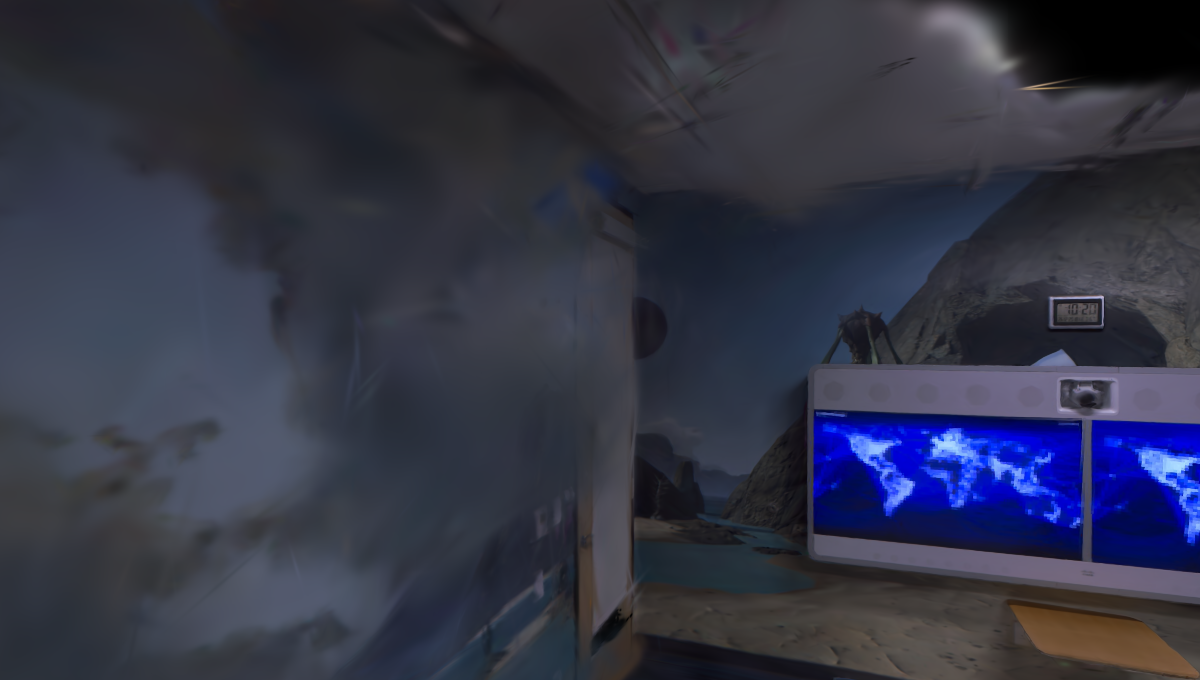}};
     \node (img1) at (0,1.6*4) {\includegraphics[width=\linewidth]{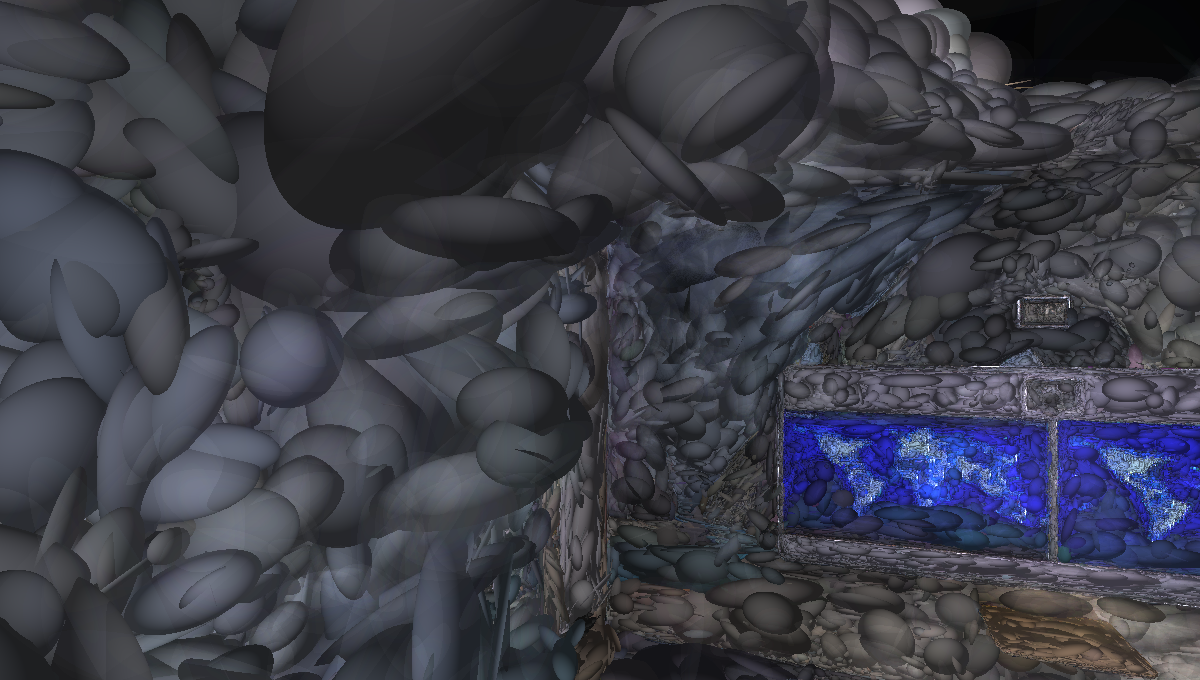}};
      \node(img5) at (0,1.6*2.85){\includegraphics[width=\linewidth]{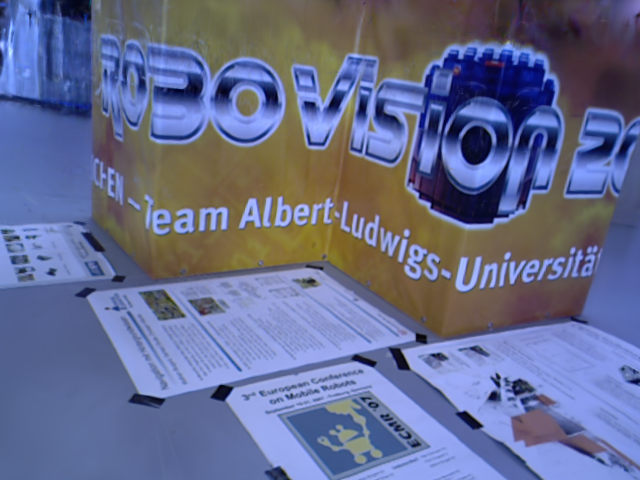}};
    \node(img6) at (0,1.6*1.55){\includegraphics[width=\linewidth]{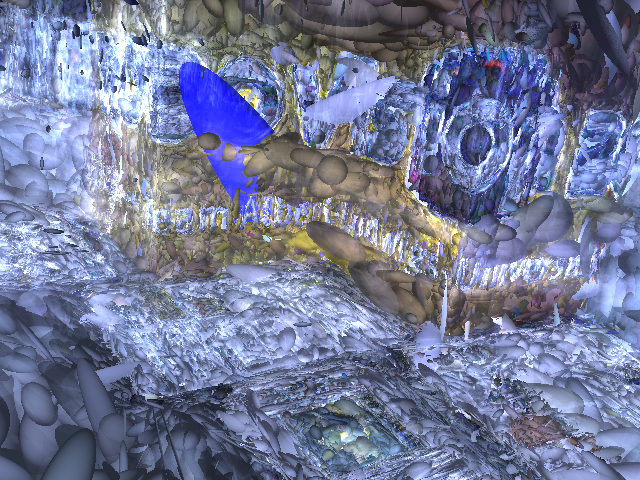}}; 
     \node (img1) at (0,1.6*0.25) {\includegraphics[width=\linewidth]{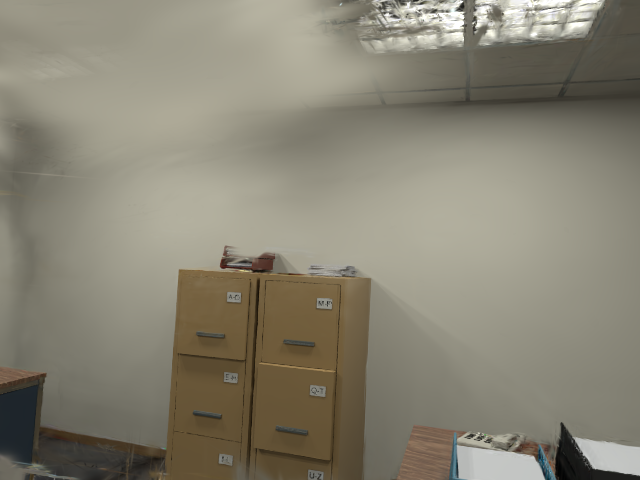}};
     \node (img1) at (0,-1.6*1.05) {\includegraphics[width=\linewidth]{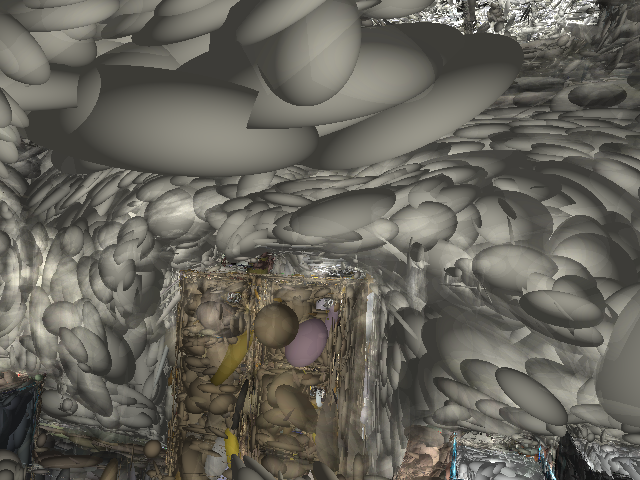}};

    \draw[red, thick] (-0.6,1.6*7) rectangle +(1.2,0.7);
    \draw[red, thick] (-0.6,1.6*6) rectangle +(1.2,0.7);

    \draw[red, thick] (-1,1.6*5.05) rectangle +(1.6,0.6);
    \draw[red, thick] (-1,1.6*4.05) rectangle +(1.6,0.6);
    
    \draw[red, thick] (-1.3,1.6*3.15) rectangle +(0.5,0.5);
    \draw[red, thick] (-1.3,1.6*1.85) rectangle +(0.5,0.5);
    \draw[red, thick] (-0.6,1.6*2.85) rectangle +(1,1);
    \draw[red, thick] (-0.6,1.6*1.55) rectangle +(1,1);
      \draw[red, thick] (-1.1,1.6*0.25) rectangle +(1.8,1);
    \draw[red, thick] (-1.1,-1.6*1.05) rectangle +(1.8,1);
  
 \end{tikzpicture}
\end{minipage}}
 \subfloat[GeoGaussian]{
 \begin{minipage}[b]{0.22\textwidth}
 \begin{tikzpicture}
    \node(img7) at (0,1.6*7){\includegraphics[width=\linewidth]{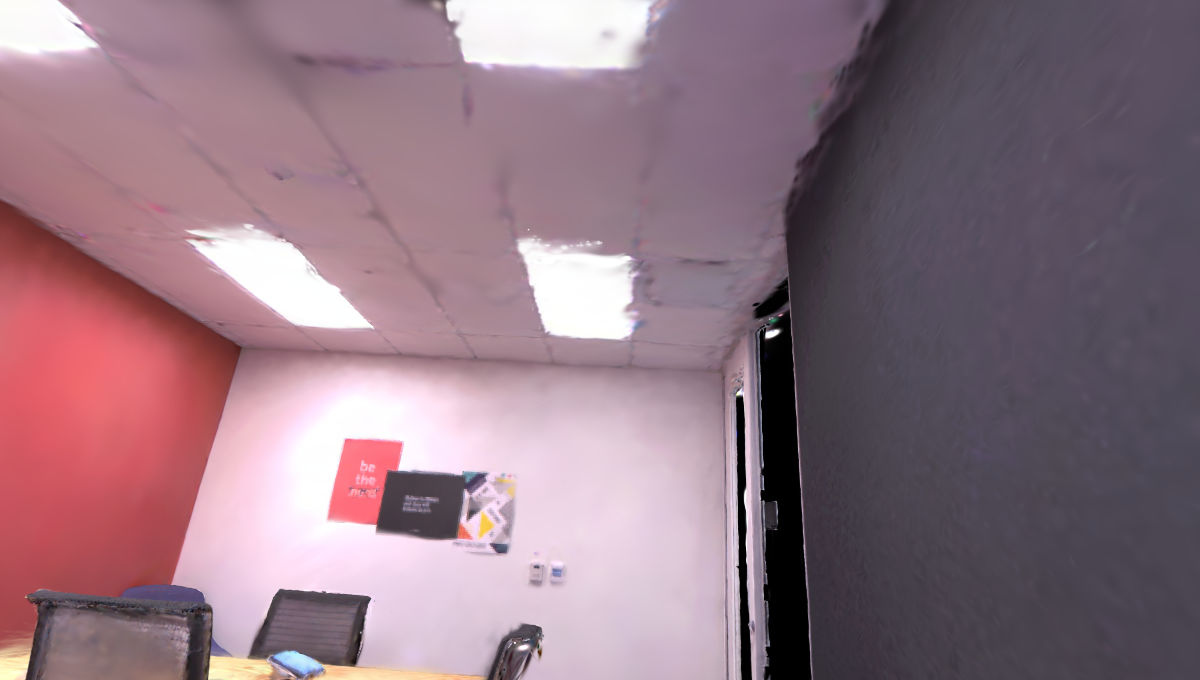}};
    \node(img8) at (0,1.6*6){\includegraphics[width=\linewidth]{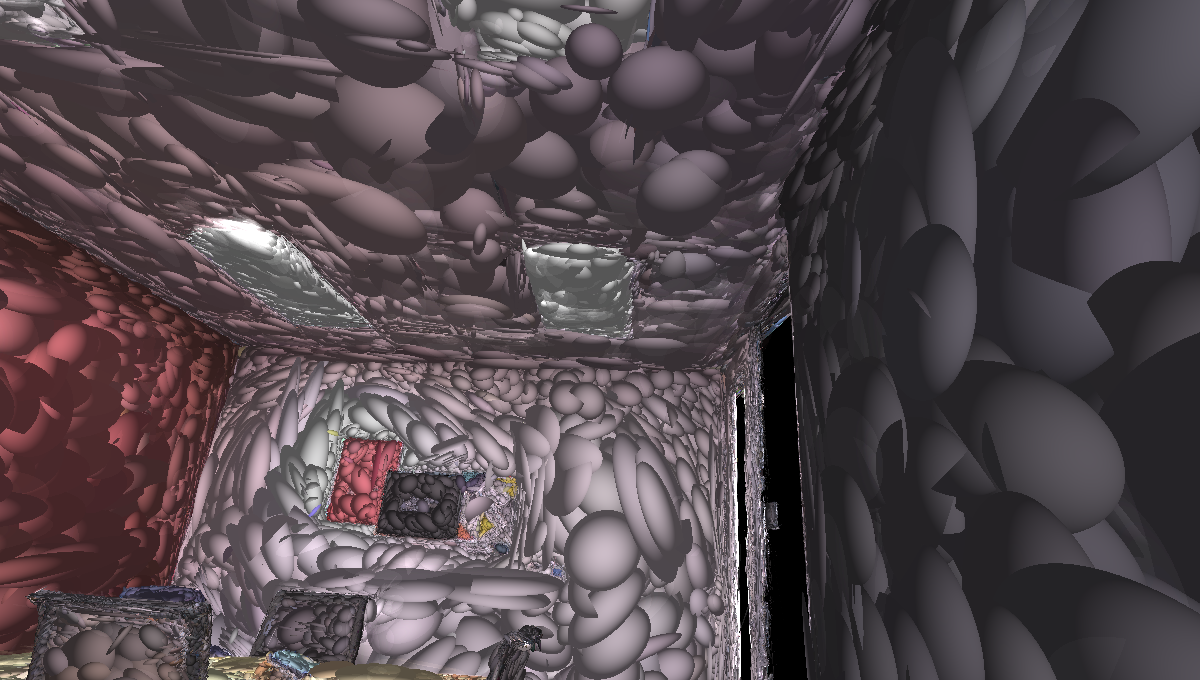}};
	\node(img3) at (0,1.6*5){ \includegraphics[width=\linewidth]{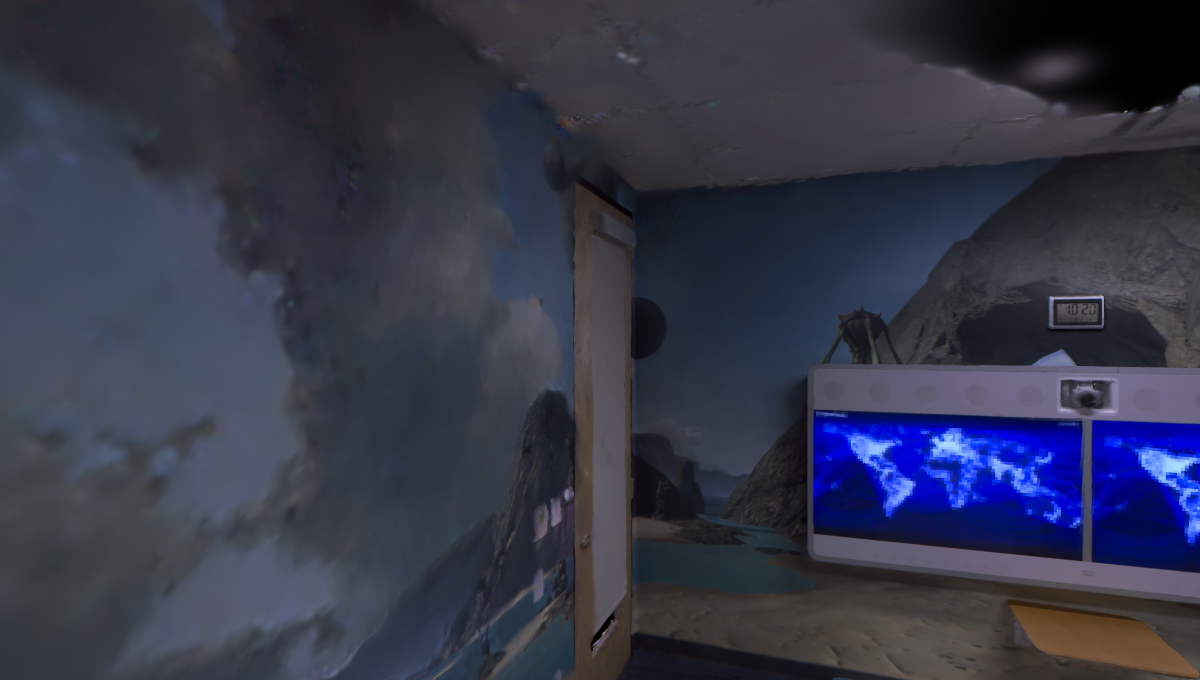}};
    \node(img4) at (0,1.6*4){\includegraphics[width=\linewidth]{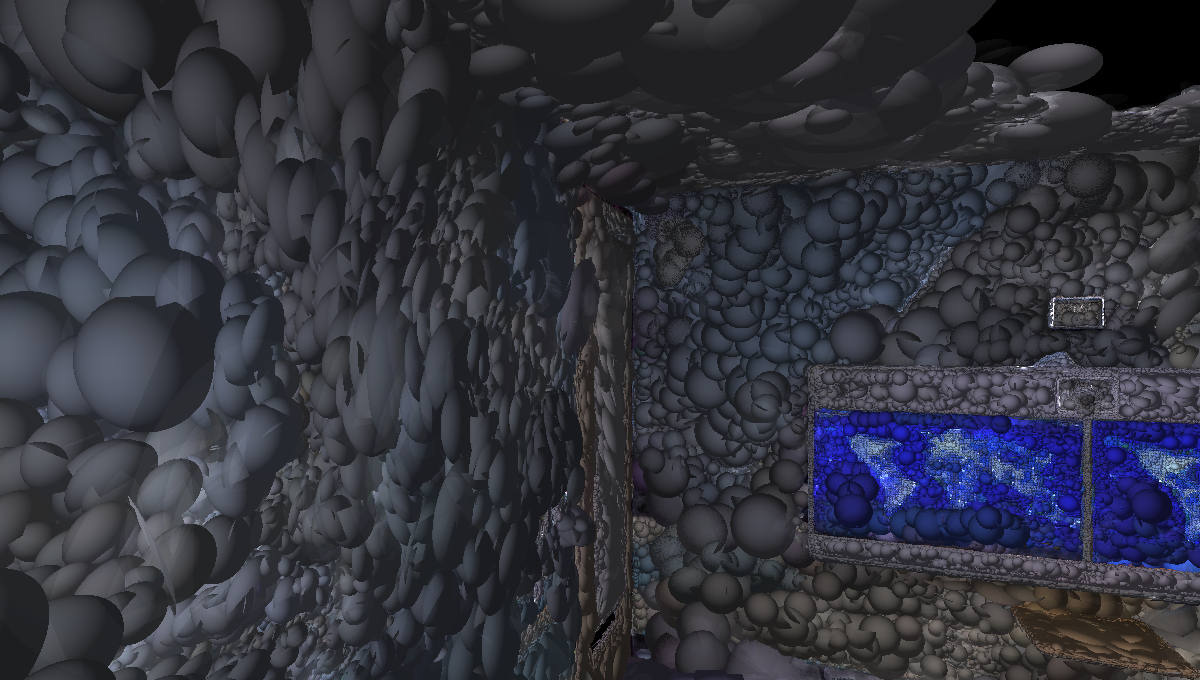}}; 
    \node(img5) at (0,1.6*2.85){\includegraphics[width=\linewidth]{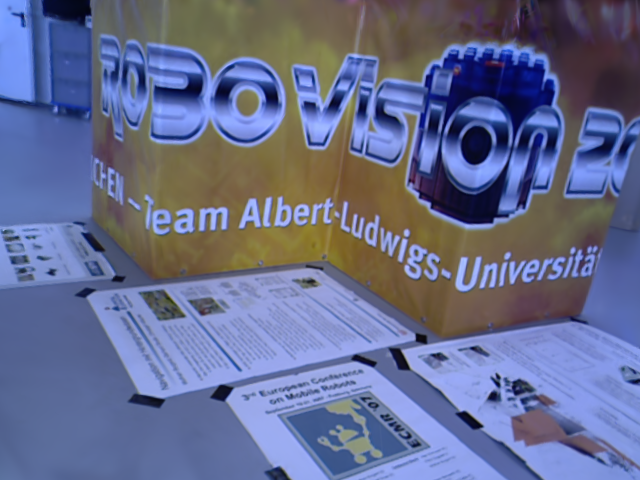}};
    \node(img6) at (0,1.6*1.55){\includegraphics[width=\linewidth]{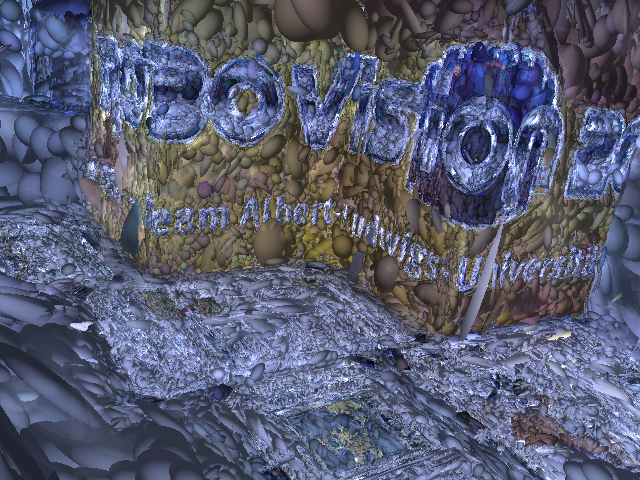}}; 
     \node (img1) at (0,1.6*0.25) {\includegraphics[width=\linewidth]{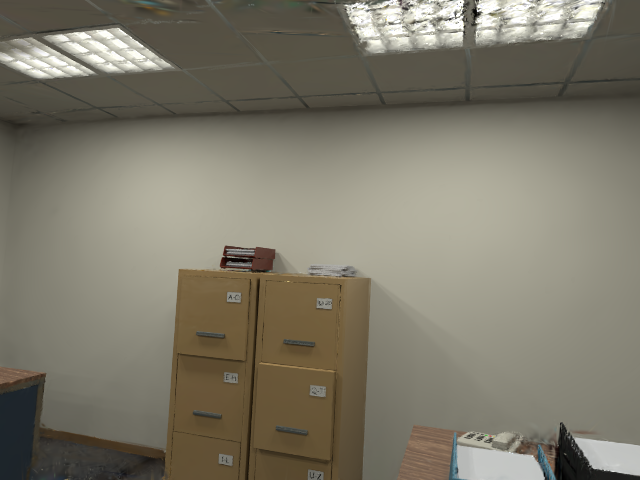}};
     \node (img1) at (0,-1.6*1.05) {\includegraphics[width=\linewidth]{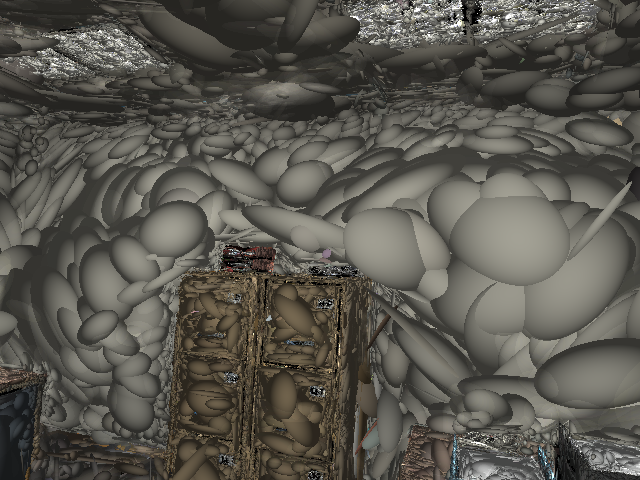}};
    
    \draw[red, thick] (-0.6,1.6*7) rectangle +(1.2,0.7);
    \draw[red, thick] (-0.6,1.6*6) rectangle +(1.2,0.7);

    \draw[red, thick] (-1,1.6*5.05) rectangle +(1.6,0.6);
    \draw[red, thick] (-1,1.6*4.05) rectangle +(1.6,0.6);
    
    \draw[red, thick] (-1.3,1.6*3.15) rectangle +(0.5,0.5);
    \draw[red, thick] (-1.3,1.6*1.85) rectangle +(0.5,0.5);
    \draw[red, thick] (-0.6,1.6*2.85) rectangle +(1,1);
    \draw[red, thick] (-0.6,1.6*1.55) rectangle +(1,1);
      \draw[red, thick] (-1.1,1.6*0.25) rectangle +(1.8,1);
    \draw[red, thick] (-1.1,-1.6*1.05) rectangle +(1.8,1);
  
 \end{tikzpicture}
	\end{minipage}}
\subfloat[Reference]{
	\begin{minipage}[b]{0.22\textwidth}
 \begin{tikzpicture}
     \node (img1) at (0,1.6*7){\includegraphics[width=\linewidth]{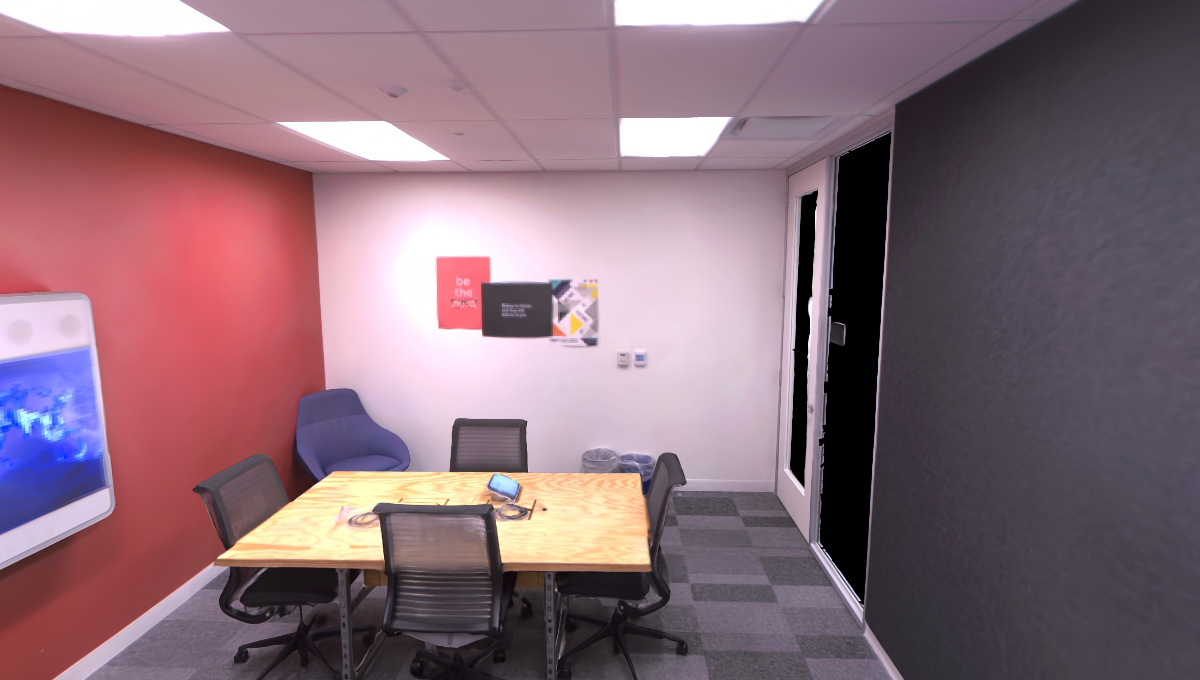}};
     \node (img1) at (0,1.6*6){\includegraphics[width=\linewidth]{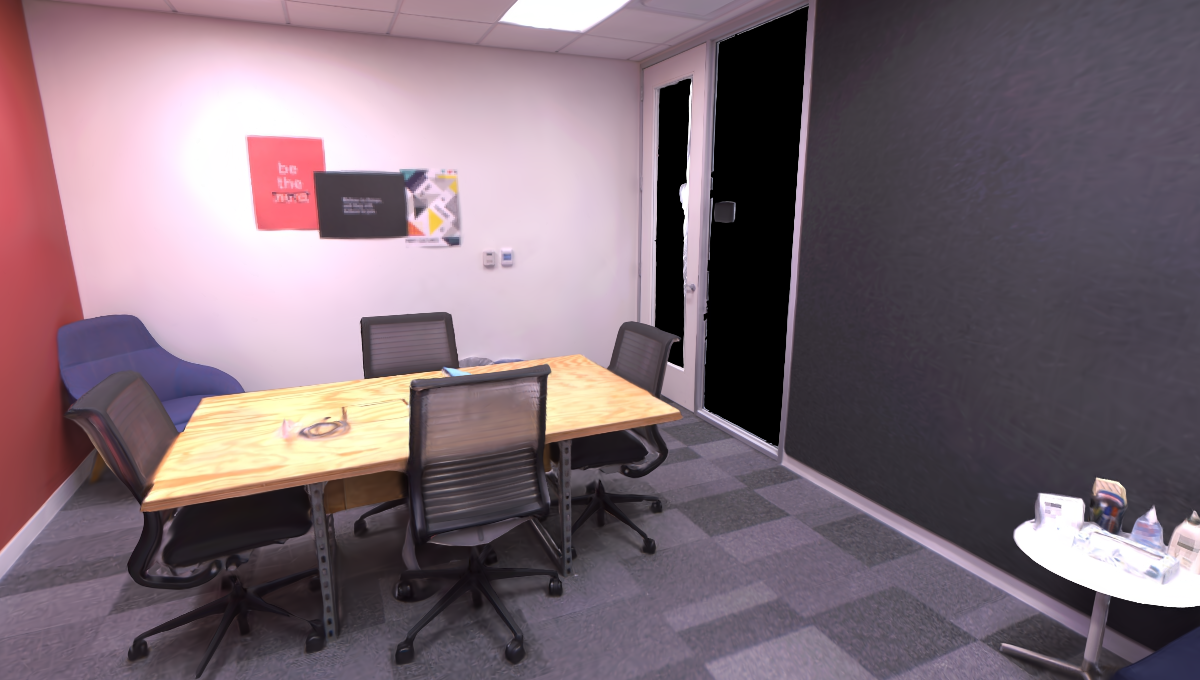}};
	\node (img1) at (0,1.6*5) {\includegraphics[width=\linewidth]{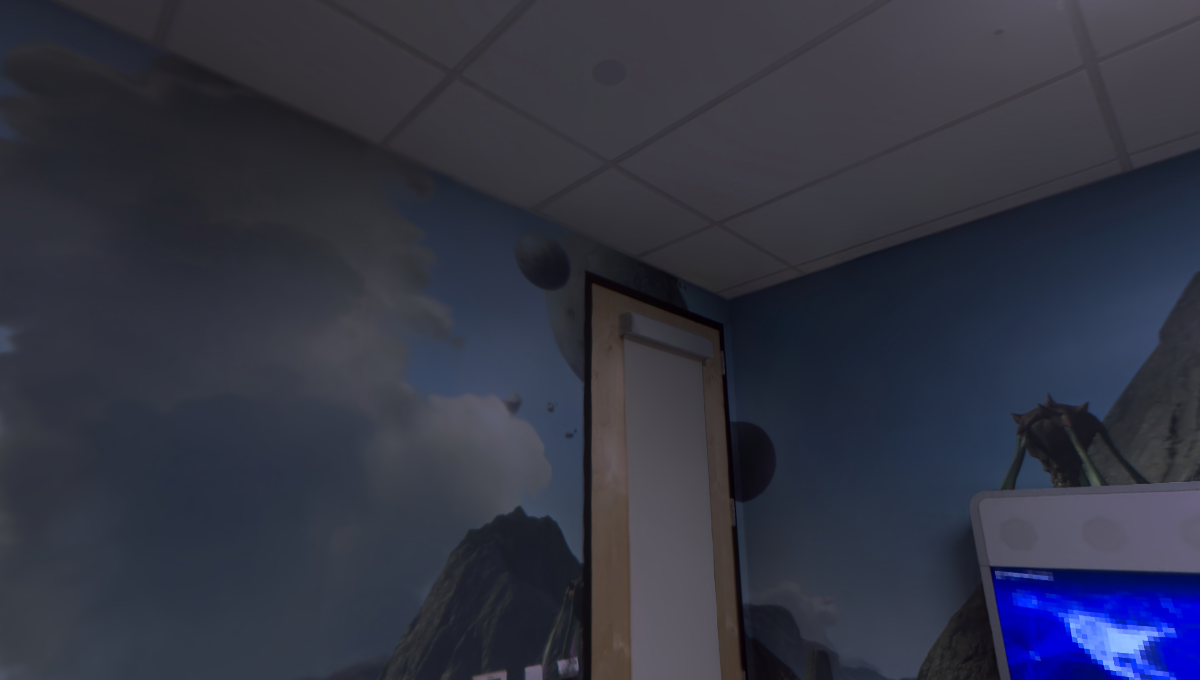}};
	\node (img1) at (0,1.6*4) {\includegraphics[width=\linewidth]{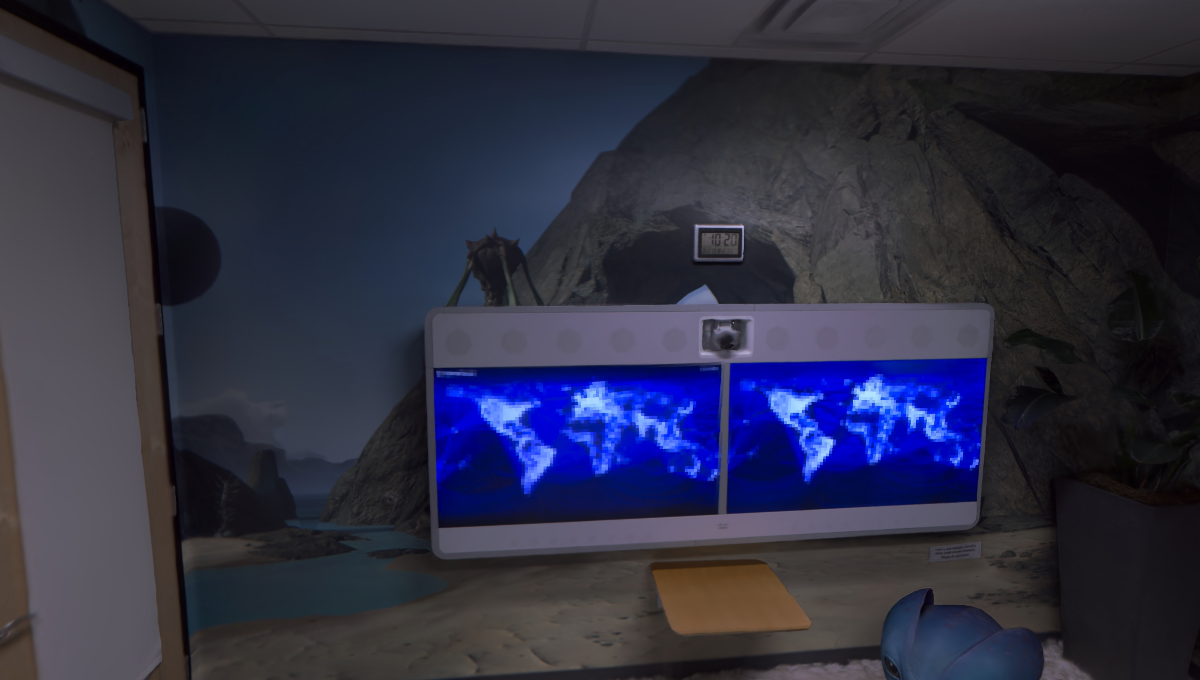}};
    \node(img5) at (0,1.6*2.85){\includegraphics[width=\linewidth]{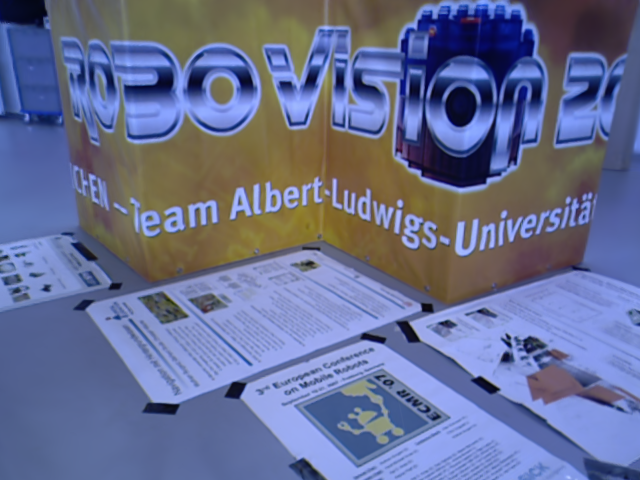}};
    \node(img6) at (0,1.6*1.55){\includegraphics[width=\linewidth]{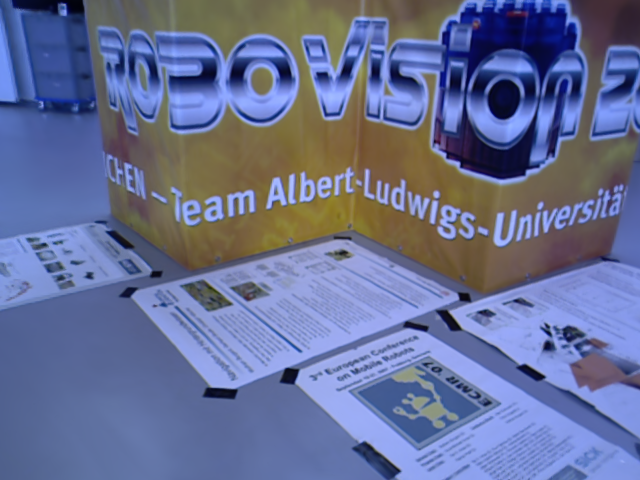}}; 
    \node (img1) at (0,1.6*0.25) {\includegraphics[width=\linewidth]{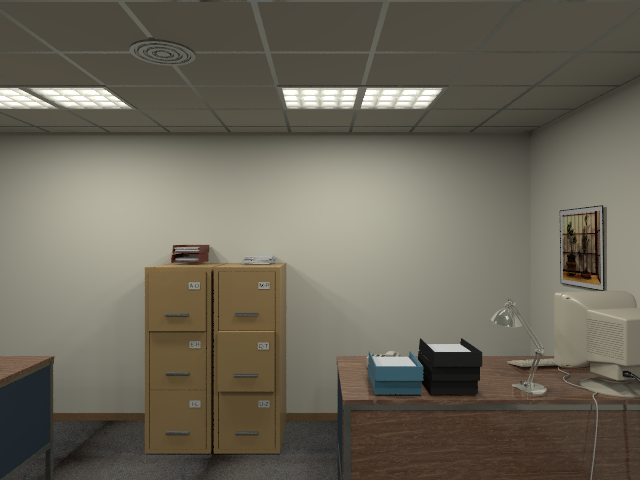}};
    \node (img1) at (0,-1.6*1.05) {\includegraphics[width=\linewidth]{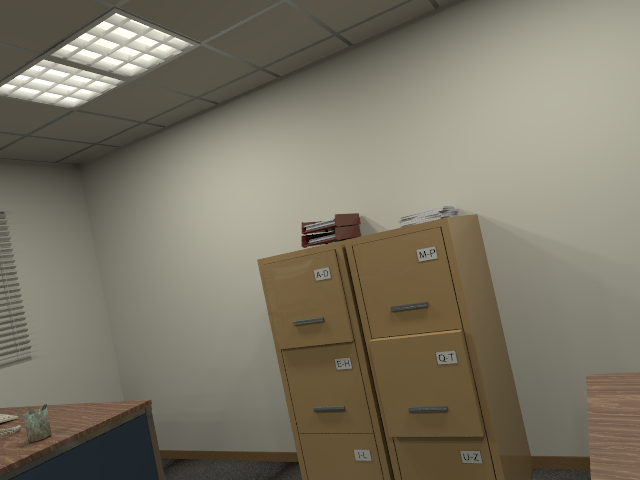}};

    \draw[red, thick] (-0.6,1.6*7) rectangle +(1.2,0.7);
    \draw[red, thick] (-1,1.6*5.05) rectangle +(1.6,0.6);
    \draw[red, thick] (-1.3,1.6*3.15) rectangle +(0.5,0.5);
    \draw[red, thick] (-1.3,1.6*1.85) rectangle +(0.5,0.5);
    \draw[red, thick] (-0.6,1.6*2.85) rectangle +(1,1);
    \draw[red, thick] (-0.6,1.6*1.55) rectangle +(1,1);
      \draw[red, thick] (-1.1,1.6*0.25) rectangle +(1.2,0.7);
    \draw[red, thick] (-1,-1.6*1.05) rectangle +(0.8,0.7);
     
     \end{tikzpicture}
	\end{minipage}}
\caption{Comparisons of novel view rendering on public datasets. At some challenging viewpoints having bigger differences in translation and orientation motions compared with training views, 
3DGS and LightGS have issues with photorealistic rendering. (d) shows the training view closest to the rendered one.}
\label{fig:qualitative}
\end{figure}

\subsection{Datasets and Metrics}
\noindent \textbf{Datasets.} Three public datasets, including Replica~\cite{straub2019replica}, TUM RGB-D~\cite{sturm2012benchmark}, and ICL-NUIM~\cite{handa2014benchmark}, are utilized in the evaluation process. The Replica dataset comprises $8$ sequences featuring living room and office scenarios. Similarly, ICL-NUIM is also a synthetic dataset that offers environments comparable to those in Replica. We also select $4$ sequences from the TUM RGB-D dataset, which were captured in real-world settings unlike Replica and ICL-NUIM. These TUM RGB-D sequences consist of distortion-free images, making them suitable as references for evaluation.

\vspace{3mm}
\noindent \textbf{Metrics.}
Following the popular evaluation protocol used in methods~\cite{kerbl3Dgaussians,straub2019replica}, standard photometric rendering quality metrics are employed in the experiment section to evaluate the quality of novel view rendering. These metrics include Peak Signal-to-Noise Ratio (PSNR), Structural Similarity Index Measure (SSIM), and Learned Perceptual Image Patch Similarity (LPIPS). 
Particularly, PSNR evaluates on a color-wise basis. 
SSIM measures the similarity between two images, which considers changes in structural information, luminance, and contrast that can occur with various types of distortion. 
LPIPS compares features of two images extracted by a pre-trained neural network such as VGG-Net~\cite{simonyan2014very} instead of comparing two images directly.

\begin{table}[]
\centering
\resizebox{.95\linewidth}{!}{%
\begin{tabular}{@{}ll|cccccccc|c@{}}
\toprule
Method & Metric & R0 & R1 & R2 & OFF0 & OFF1 & OFF2 & OFF3 & OFF4 & Avg. \\ \hline
\begin{tabular}[c]{@{}l@{}}Vox-\\ Fusion~\cite{voxfusion}\end{tabular} 
& \begin{tabular}[r]{@{}l@{}}PSNR$\uparrow$ \\ SSIM$\uparrow$ \\ LPIPS$\downarrow$\end{tabular} 
& \begin{tabular}[c]{@{}c@{}}22.39 \\ 0.683 \\ 0.303\end{tabular} 
& \begin{tabular}[c]{@{}c@{}}22.36 \\ 0.751 \\ 0.269\end{tabular} 
& \begin{tabular}[c]{@{}c@{}}23.92 \\ 0.798 \\ 0.234\end{tabular} 
& \begin{tabular}[c]{@{}c@{}}27.79 \\ 0.857 \\ 0.241\end{tabular} 
& \begin{tabular}[c]{@{}c@{}}29.83 \\ 0.876 \\ 0.184\end{tabular} 
& \begin{tabular}[c]{@{}c@{}}20.33 \\ 0.794 \\ 0.243\end{tabular} 
& \begin{tabular}[c]{@{}c@{}}23.47 \\ 0.803 \\ 0.213\end{tabular} 
& \begin{tabular}[c]{@{}c@{}}25.21 \\ 0.847 \\ 0.199\end{tabular} 
& \begin{tabular}[c]{@{}c@{}}24.41 \\ 0.801 \\ 0.236\end{tabular} \\ 
\hline
\begin{tabular}[c]{@{}l@{}}Point-\\ SLAM~\cite{Sandstrom_2023_ICCV}\end{tabular} 
& \begin{tabular}[r]{@{}l@{}}PSNR$\uparrow$ \\ SSIM$\uparrow$\\ LPIPS$\downarrow$\end{tabular}
& \begin{tabular}[c]{@{}c@{}}32.40 \\ \cellcolor[HTML]{57cc99}0.974 \\ 0.113\end{tabular} 
& \begin{tabular}[c]{@{}c@{}}34.08 \\ \cellcolor[HTML]{c7f9cc}0.977 \\ 0.116\end{tabular} 
& \begin{tabular}[c]{@{}c@{}}35.50 \\ \cellcolor[HTML]{57cc99}0.982 \\ 0.111\end{tabular} 
& \begin{tabular}[c]{@{}c@{}}38.26 \\ \cellcolor[HTML]{57cc99}0.983 \\ 0.100\end{tabular} 
& \begin{tabular}[c]{@{}c@{}}39.16 \\ \cellcolor[HTML]{57cc99}0.986 \\ 0.118\end{tabular} 
& \begin{tabular}[c]{@{}c@{}}33.99 \\ 0.960 \\ 0.156\end{tabular} 
& \begin{tabular}[c]{@{}c@{}}33.48 \\ 0.960 \\ 0.132\end{tabular} 
& \begin{tabular}[c]{@{}c@{}}33.49 \\ \cellcolor[HTML]{57cc99}0.979 \\ 0.142\end{tabular} 
& \begin{tabular}[c]{@{}c@{}}35.17 \\ \cellcolor[HTML]{57cc99}0.975 \\ 0.124\end{tabular} \\ 
\hline
\begin{tabular}[c]{@{}l@{}}Gaussian-\\ Splatting \\SLAM~\cite{Matsuki:Murai:etal:CVPR2024}\end{tabular} 
& \begin{tabular}[r]{@{}l@{}}PSNR$\uparrow$ \\ SSIM$\uparrow$ \\ LPIPS$\downarrow$\end{tabular} 
& \begin{tabular}[c]{@{}c@{}}\cellcolor[HTML]{c7f9cc}34.83 \\ \cellcolor[HTML]{c7f9cc}0.954 \\ \cellcolor[HTML]{c7f9cc}0.068\end{tabular} 
& \begin{tabular}[c]{@{}c@{}}\cellcolor[HTML]{c7f9cc}36.43 \\ 0.959 \\ \cellcolor[HTML]{c7f9cc}0.076\end{tabular} 
& \begin{tabular}[c]{@{}c@{}}\cellcolor[HTML]{c7f9cc}37.49 \\ 0.965 \\ \cellcolor[HTML]{c7f9cc}0.07\end{tabular} 
& \begin{tabular}[c]{@{}c@{}}\cellcolor[HTML]{c7f9cc}39.95 \\ 0.971 \\ \cellcolor[HTML]{c7f9cc}0.072\end{tabular} 
& \begin{tabular}[c]{@{}c@{}}\cellcolor[HTML]{c7f9cc}42.09 \\ \cellcolor[HTML]{c7f9cc}0.977 \\ \cellcolor[HTML]{c7f9cc}0.055\end{tabular} 
& \begin{tabular}[c]{@{}c@{}}\cellcolor[HTML]{c7f9cc}36.24 \\ \cellcolor[HTML]{c7f9cc}0.964 \\ \cellcolor[HTML]{c7f9cc}0.078\end{tabular} 
& \begin{tabular}[c]{@{}c@{}}\cellcolor[HTML]{57cc99}36.70 \\ \cellcolor[HTML]{c7f9cc}0.963 \\ \cellcolor[HTML]{c7f9cc}0.065\end{tabular} 
& \begin{tabular}[c]{@{}c@{}}\cellcolor[HTML]{c7f9cc}36.07 \\ 0.957 \\ \cellcolor[HTML]{c7f9cc}0.099\end{tabular} 
& \begin{tabular}[c]{@{}c@{}}\cellcolor[HTML]{c7f9cc}37.50 \\ 0.960 \\ \cellcolor[HTML]{c7f9cc}0.070\end{tabular} \\ 
\hline
\begin{tabular}[c]{@{}l@{}} GeoGaussian\\(ours)  \end{tabular} 
& \begin{tabular}[r]{@{}l@{}}PSNR $\uparrow$\\ SSIM$\uparrow$\\ LPIPS$\downarrow$\end{tabular} 
& \begin{tabular}[c]{@{}c@{}} \cellcolor[HTML]{57cc99}35.20\\ 0.952\\ \cellcolor[HTML]{57cc99}0.029 \end{tabular} 
& \begin{tabular}[c]{@{}c@{}} \cellcolor[HTML]{57cc99}38.24 \\ \cellcolor[HTML]{57cc99}0.979 \\ \cellcolor[HTML]{57cc99}0.021\end{tabular} 
& \begin{tabular}[c]{@{}c@{}} \cellcolor[HTML]{57cc99}39.14 \\ \cellcolor[HTML]{c7f9cc}0.970\\ \cellcolor[HTML]{57cc99}0.024\end{tabular} 
& \begin{tabular}[c]{@{}c@{}} \cellcolor[HTML]{57cc99}42.74 \\ \cellcolor[HTML]{c7f9cc}0.981\\ \cellcolor[HTML]{57cc99}0.016\end{tabular} 
& \begin{tabular}[c]{@{}c@{}} \cellcolor[HTML]{57cc99}42.20 \\ 0.970 \\ \cellcolor[HTML]{57cc99}0.040\end{tabular} 
& \begin{tabular}[c]{@{}c@{}} \cellcolor[HTML]{57cc99}37.31 \\ \cellcolor[HTML]{57cc99}0.970 \\ \cellcolor[HTML]{57cc99}0.029\end{tabular} 
& \begin{tabular}[c]{@{}c@{}} \cellcolor[HTML]{c7f9cc}36.66 \\ \cellcolor[HTML]{57cc99}0.964 \\ \cellcolor[HTML]{57cc99}0.029\end{tabular} 
& \begin{tabular}[c]{@{}c@{}} \cellcolor[HTML]{57cc99}38.74 \\ \cellcolor[HTML]{c7f9cc}0.967\\ \cellcolor[HTML]{57cc99}0.031\end{tabular} 
& \begin{tabular}[c]{@{}c@{}} \cellcolor[HTML]{57cc99}38.78 \\ \cellcolor[HTML]{c7f9cc}0.969 \\ \cellcolor[HTML]{57cc99}0.027\end{tabular} \\
\bottomrule
\end{tabular}}
\caption{Comparison of different solutions for novel view rendering on the Replica dataset, GeoGaussian is fed by point clouds, initial camera poses, and monocular images, while other methods work in real-time with RGB-D streams. Results of Gaussian-Splatting SLAM, Vox-Fusion, and Point-SLAM are taken from~\cite{Matsuki:Murai:etal:CVPR2024}.}

\label{tab:replica}
\end{table}

\subsection{Scene Rendering Solutions}
Solutions for NVS tasks can be generally classified into end-to-end and hybrid methods.
Gaussian Splatting SLAM~\cite{Matsuki:Murai:etal:CVPR2024} represents end-to-end methods that use Gaussians as the map for incremental localization, reconstruction, and rendering tasks. 
In contrast, Vox-Fusion~\cite{voxfusion} represents hybrid methods that integrate implicit neural representations into traditional volumetric methods, and estimate 6-DoF camera poses following traditional pose estimation without using neural embeddings saved in voxels. Similarly, our GeoGaussian starts from point clouds and camera poses which can be generated by SLAM~\cite{yunus2021manhattanslam} and Structure-from-Motion~\cite{schoenberger2016sfm} methods, and uses differentiable 3D Gaussian representations for rendering instead of neural implicit ones. 
Nonetheless, the big difference is that methods 
such as Vox-Fusion 
operates online, while Gaussian Splatting optimization functions 
similar to a global refinement module 
that requires more time and unable to perform real-time optimization.

We put aside the all other differences between these methods to focus on comparisons on the rendering task. Specifically, we 
compare the rendering results of our GeoGaussian against Gaussian-Splatting SLAM, Point-SLAM, and Vox-Fusion on the same 
Replica dataset. As listed in Table~\ref{tab:replica}, 
our proposed GeoGaussian using offline Gaussian Splatting achieves the best novel view rendering results, especially in PSNR and LPIPS metrics. Point-SLAM is also very robust, especially in SSIM, but the LPIPS metric is worse than Gaussian-Splatting SLAM. 
These results show that the Gaussian Splatting algorithm 
yield more impressive performance in photo-realistic rendering problems compared with neural representation methods used in Vox-Fusion and Point-SLAM. 

\subsection{Gaussian Splatting in NVS}
In this section, 
our GeoGaussian is benchmarked against Gaussian Splatting methods which are fed by the same inputs such as sparse point clouds, initial camera poses, and monocular images based on the preprocessing method~\cite{yunus2021manhattanslam}. 3DGS~\cite{kerbl3Dgaussians} is the most important baseline in this area, while LightGS~\cite{fan2023lightgaussian} is a refinement approach based on 3DGS 
that involves the pruning of insignificant Gaussian points and further fine-tuning. 

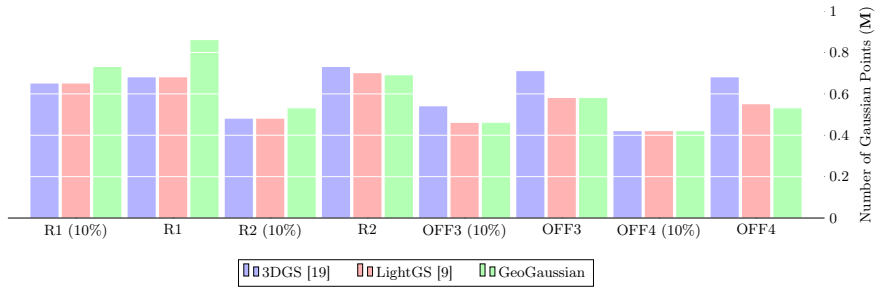
\begin{figure}
    \centering
    \resizebox{.95\linewidth}{!}{%
    \begin{tikzpicture}
    \begin{axis}[
        ybar, axis on top,
        height=6cm, width=19cm,
        bar width=0.6cm,
        ymajorgrids, tick align=inside,
        major grid style={draw=white},
        ymin=0, ymax=1.0,
        axis x line*=bottom,
        axis y line*=right,
        y axis line style={opacity=0},
        tickwidth=0pt,
        enlarge x limits=true,
        legend style={
            at={(0.5,-0.2)},
            anchor=north,
            legend columns=-1,
            /tikz/every even column/.append style={column sep=0.5cm}
        },
        ylabel={Number of Gaussian Points (\textbf{M})},
        symbolic x coords={
           R1 (10\%), R1, R2 (10\%), R2, OFF3 (10\%), OFF3, OFF4 (10\%), OFF4,},
       xtick=data,
       nodes near coords={
       }]
    \addplot [draw=none, fill=blue!30] coordinates {
      (R1 (10\%), 0.65) (R1, 0.68) (R2 (10\%), 0.48) (R2, 0.73) (OFF3 (10\%), 0.54) (OFF3, 0.71) (OFF4 (10\%), 0.42) (OFF4, 0.68) };
   \addplot [draw=none,fill=red!30] coordinates {
      (R1 (10\%), 0.65) (R1, 0.68) (R2 (10\%), 0.48) (R2, 0.70) (OFF3 (10\%), 0.46) (OFF3, 0.58) (OFF4 (10\%), 0.42) (OFF4, 0.55) };
   \addplot [draw=none, fill=green!30] coordinates {
      (R1 (10\%), 0.73) (R1, 0.86) (R2 (10\%), 0.53) (R2, 0.69) (OFF3 (10\%), 0.46) (OFF3, 0.58) (OFF4 (10\%), 0.42) (OFF4, 0.53) };
    \legend{3DGS~\cite{kerbl3Dgaussians}, LightGS~\cite{fan2023lightgaussian}, GeoGaussian}; 
  \end{axis}    
\end{tikzpicture}} 
    \caption{Statistics of the number of Gaussians in sequences of Replica. To make the comparison compact, more values are illustrated in Appendix.}
    \label{fig:model-size}
\end{figure}

For the experiments, we feed the three methods with the same inputs to validate their rendering performance as listed in Table~\ref{tab:sparse-replica}. Additionally, we evaluate these methods in sparse views, which is a significant metric for evaluating the generality of the trained Gaussian models. Evaluating on a dataset that is too similar to the training dataset can make it difficult to assess the 
generality of the model. To address this issue, we first select the evaluation dataset at a frequency of one frame every five frames. The remaining data is considered as the training dataset, referred to as R1 (100\%) as an example in the second row of Table~\ref{tab:sparse-replica} and Figure~\ref{fig:model-size}. Similarly, we use only 10\% of the training dataset for training, referred to as R1 (10\%). It is important to note that the removal process is uniformly implemented across all sequences.

\begin{table}[]
\centering
\resizebox{.90\linewidth}{!}{%
\begin{tabular}{ll|cccc|cccc|cccc}
\toprule
\multicolumn{2}{c|}{Methods} & \multicolumn{4}{c|}{3DGS~\cite{kerbl3Dgaussians}} & \multicolumn{4}{c|}{LightGS~\cite{fan2023lightgaussian}} & \multicolumn{4}{c}{GeoGaussian} \\
\multicolumn{2}{c|}{Data} & 10\% & 16.6\% & 50\% & 100\% & 10\% & 16.6\% & 50\% & 100\% & 10\% & 16.6\% & 50\% & 100\% \\ \hline

R1 & \begin{tabular}[r]{@{}l@{}}PSNR$\uparrow$\\ SSIM$\uparrow$\\ LPIPS$\downarrow$\end{tabular} 
& \begin{tabular}[c]{@{}c@{}} 30.49 \\ \cellcolor[HTML]{c7f9cc}0.932 \\ \cellcolor[HTML]{c7f9cc}0.051\end{tabular}
& \begin{tabular}[c]{@{}c@{}} 33.98 \\ \cellcolor[HTML]{c7f9cc}0.951 \\ 0.036\end{tabular}
& \begin{tabular}[c]{@{}c@{}} 37.45 \\ 0.964 \\ 0.029\end{tabular}
& \begin{tabular}[c]{@{}c@{}} 37.60 \\ 0.965 \\ 0.028 \end{tabular}

& \begin{tabular}[c]{@{}c@{}} \cellcolor[HTML]{c7f9cc}30.54 \\ \cellcolor[HTML]{c7f9cc}0.932 \\ \cellcolor[HTML]{c7f9cc}0.051\end{tabular}
& \begin{tabular}[c]{@{}c@{}} \cellcolor[HTML]{c7f9cc}34.06 \\ \cellcolor[HTML]{c7f9cc}0.951 \\ \cellcolor[HTML]{c7f9cc}0.035\end{tabular} 
& \begin{tabular}[c]{@{}c@{}} \cellcolor[HTML]{c7f9cc}37.72 \\ \cellcolor[HTML]{c7f9cc}0.965 \\ \cellcolor[HTML]{c7f9cc}0.028\end{tabular} 
& \begin{tabular}[c]{@{}c@{}} \cellcolor[HTML]{57cc99}38.44 \\ \cellcolor[HTML]{c7f9cc}0.967 \\ \cellcolor[HTML]{c7f9cc}0.025\end{tabular} 

& \begin{tabular}[c]{@{}c@{}} \cellcolor[HTML]{57cc99}31.65 \\ \cellcolor[HTML]{57cc99}0.937 \\ \cellcolor[HTML]{57cc99}0.041\end{tabular}
& \begin{tabular}[c]{@{}c@{}} \cellcolor[HTML]{57cc99}35.17 \\ \cellcolor[HTML]{57cc99}0.957 \\ \cellcolor[HTML]{57cc99}0.027\end{tabular}
& \begin{tabular}[c]{@{}c@{}} \cellcolor[HTML]{57cc99}38.00 \\ \cellcolor[HTML]{57cc99}0.968 \\ \cellcolor[HTML]{57cc99}0.022\end{tabular}  
& \begin{tabular}[c]{@{}c@{}} \cellcolor[HTML]{c7f9cc}38.24 \\ \cellcolor[HTML]{57cc99}0.979 \\ \cellcolor[HTML]{57cc99}0.021\end{tabular} \\
\hline

R2 & \begin{tabular}[r]{@{}l@{}}PSNR$\uparrow$\\ SSIM$\uparrow$ \\ LPIPS$\downarrow$\end{tabular} 
& \begin{tabular}[c]{@{}c@{}} 31.53 \\ \cellcolor[HTML]{c7f9cc}0.935 \\ 0.050 \end{tabular} 
& \begin{tabular}[c]{@{}c@{}} 35.82 \\ \cellcolor[HTML]{c7f9cc}0.959 \\ \cellcolor[HTML]{c7f9cc}0.031 \end{tabular} 
& \begin{tabular}[c]{@{}c@{}} 38.53 \\ \cellcolor[HTML]{c7f9cc}0.968 \\ 0.028 \end{tabular} 
& \begin{tabular}[c]{@{}c@{}} 38.70 \\ \cellcolor[HTML]{c7f9cc}0.968 \\ 0.029 \end{tabular} 

& \begin{tabular}[c]{@{}c@{}} \cellcolor[HTML]{c7f9cc}31.54 \\\cellcolor[HTML]{c7f9cc} 0.935\\ \cellcolor[HTML]{c7f9cc}0.049 \end{tabular}
& \begin{tabular}[c]{@{}c@{}} \cellcolor[HTML]{c7f9cc}35.93 \\ \cellcolor[HTML]{c7f9cc}0.959\\\cellcolor[HTML]{c7f9cc} 0.031 \end{tabular} 
& \begin{tabular}[c]{@{}c@{}} \cellcolor[HTML]{c7f9cc}38.78 \\ \cellcolor[HTML]{c7f9cc}0.968\\ \cellcolor[HTML]{c7f9cc}0.027 \end{tabular} 
& \begin{tabular}[c]{@{}c@{}} \cellcolor[HTML]{c7f9cc}39.07 \\ \cellcolor[HTML]{c7f9cc}0.968\\ \cellcolor[HTML]{c7f9cc}0.028 \end{tabular} 

& \begin{tabular}[c]{@{}c@{}} \cellcolor[HTML]{57cc99}32.13 \\ \cellcolor[HTML]{57cc99}0.943 \\ \cellcolor[HTML]{57cc99}0.041 \end{tabular} 
& \begin{tabular}[c]{@{}c@{}} \cellcolor[HTML]{57cc99}36.81 \\ \cellcolor[HTML]{57cc99}0.963 \\ \cellcolor[HTML]{57cc99}0.025 \end{tabular} 
& \begin{tabular}[c]{@{}c@{}} \cellcolor[HTML]{57cc99}38.84 \\ \cellcolor[HTML]{57cc99}0.969 \\ \cellcolor[HTML]{57cc99}0.024 \end{tabular} 
& \begin{tabular}[c]{@{}c@{}} \cellcolor[HTML]{57cc99}39.14 \\ \cellcolor[HTML]{57cc99}0.970 \\ \cellcolor[HTML]{57cc99}0.024 \end{tabular}  \\
\hline

OFF3 & \begin{tabular}[r]{@{}l@{}}PSNR$\uparrow$\\ SSIM$\uparrow$\\ LPIPS$\downarrow$\end{tabular} 
& \begin{tabular}[c]{@{}c@{}}30.90\\ \cellcolor[HTML]{c7f9cc}0.928 \\ \cellcolor[HTML]{c7f9cc}0.052\end{tabular} 
& \begin{tabular}[c]{@{}c@{}}33.86\\  0.946 \\  \cellcolor[HTML]{c7f9cc}0.040\end{tabular} 
& \begin{tabular}[c]{@{}c@{}}36.26\\  \cellcolor[HTML]{c7f9cc}0.958 \\  0.037\end{tabular} 
& \begin{tabular}[c]{@{}c@{}}36.56\\  \cellcolor[HTML]{c7f9cc}0.959 \\  0.036\end{tabular} 

& \begin{tabular}[c]{@{}c@{}}\cellcolor[HTML]{c7f9cc}30.93\\ \cellcolor[HTML]{c7f9cc}0.928 \\ \cellcolor[HTML]{c7f9cc}0.052\end{tabular} 
& \begin{tabular}[c]{@{}c@{}}\cellcolor[HTML]{c7f9cc}33.90\\ \cellcolor[HTML]{c7f9cc}0.947 \\ \cellcolor[HTML]{c7f9cc}0.040\end{tabular} 
& \begin{tabular}[c]{@{}c@{}}\cellcolor[HTML]{c7f9cc}36.38\\ \cellcolor[HTML]{c7f9cc}0.958 \\ \cellcolor[HTML]{c7f9cc}0.036\end{tabular} 
& \begin{tabular}[c]{@{}c@{}}\cellcolor[HTML]{c7f9cc}36.63\\ 0.958 \\ \cellcolor[HTML]{c7f9cc}0.037\end{tabular} 

& \begin{tabular}[c]{@{}c@{}} \cellcolor[HTML]{57cc99}31.62 \\ \cellcolor[HTML]{57cc99}0.938 \\ \cellcolor[HTML]{57cc99}0.040\end{tabular}
& \begin{tabular}[c]{@{}c@{}} \cellcolor[HTML]{57cc99}33.91 \\ \cellcolor[HTML]{57cc99}0.953 \\ \cellcolor[HTML]{57cc99}0.032\end{tabular}
& \begin{tabular}[c]{@{}c@{}} \cellcolor[HTML]{57cc99}36.42 \\ \cellcolor[HTML]{57cc99}0.963 \\ \cellcolor[HTML]{57cc99}0.029\end{tabular}
& \begin{tabular}[c]{@{}c@{}} \cellcolor[HTML]{57cc99}36.66 \\ \cellcolor[HTML]{57cc99}0.964 \\ \cellcolor[HTML]{57cc99}0.029\end{tabular}  \\
\hline

OFF4 & \begin{tabular}[r]{@{}l@{}}PSNR$\uparrow$\\ SSIM$\uparrow$\\ LPIPS$\downarrow$\end{tabular} 
& \begin{tabular}[c]{@{}c@{}}\cellcolor[HTML]{c7f9cc}29.55\\ 0.920\\  \cellcolor[HTML]{c7f9cc}0.070\end{tabular} 
& \begin{tabular}[c]{@{}c@{}}32.98\\ \cellcolor[HTML]{c7f9cc}0.941\\  \cellcolor[HTML]{c7f9cc}0.049\end{tabular} 
& \begin{tabular}[c]{@{}c@{}}37.70\\ \cellcolor[HTML]{c7f9cc}0.962\\  0.037\end{tabular} 
& \begin{tabular}[c]{@{}c@{}} 38.48 \\ \cellcolor[HTML]{c7f9cc}0.964\\  \cellcolor[HTML]{c7f9cc}0.035\end{tabular} 

&  \begin{tabular}[c]{@{}c@{}}29.51\\ 0.920\\  \cellcolor[HTML]{c7f9cc}0.070\end{tabular}
&  \begin{tabular}[c]{@{}c@{}}\cellcolor[HTML]{c7f9cc}32.97\\ \cellcolor[HTML]{c7f9cc}0.941\\  \cellcolor[HTML]{c7f9cc}0.049\end{tabular}
&  \begin{tabular}[c]{@{}c@{}}\cellcolor[HTML]{c7f9cc}37.95\\ \cellcolor[HTML]{c7f9cc}0.962\\  \cellcolor[HTML]{c7f9cc}0.036\end{tabular}
&  \begin{tabular}[c]{@{}c@{}}\cellcolor[HTML]{c7f9cc}38.59\\ \cellcolor[HTML]{c7f9cc}0.964\\  0.036\end{tabular} 
& \begin{tabular}[c]{@{}c@{}}\cellcolor[HTML]{57cc99}31.90\\ \cellcolor[HTML]{57cc99}0.936\\  \cellcolor[HTML]{57cc99}0.050\end{tabular}
& \begin{tabular}[c]{@{}c@{}}\cellcolor[HTML]{57cc99}34.61\\ \cellcolor[HTML]{57cc99}0.953\\  \cellcolor[HTML]{57cc99}0.036\end{tabular} 
& \begin{tabular}[c]{@{}c@{}}\cellcolor[HTML]{57cc99}38.30\\ \cellcolor[HTML]{57cc99}0.966\\  \cellcolor[HTML]{57cc99}0.030\end{tabular} 
& \begin{tabular}[c]{@{}c@{}}\cellcolor[HTML]{57cc99}38.74\\ \cellcolor[HTML]{57cc99}0.967\\ \cellcolor[HTML]{57cc99}0.031\end{tabular} \\ \hline
Avg. & \begin{tabular}[r]{@{}l@{}}PSNR$\uparrow$\\ SSIM$\uparrow$\\ LPIPS$\downarrow$\end{tabular} 
&  \begin{tabular}[c]{@{}c@{}}30.62\\ \cellcolor[HTML]{c7f9cc}0.929\\  \cellcolor[HTML]{c7f9cc}0.056\end{tabular} 
&  \begin{tabular}[c]{@{}c@{}}34.16\\ 0.949\\  \cellcolor[HTML]{c7f9cc}0.039\end{tabular} 
&  \begin{tabular}[c]{@{}c@{}}37.49\\ \cellcolor[HTML]{c7f9cc}0.964\\  \cellcolor[HTML]{c7f9cc}0.032\end{tabular} 
&  \begin{tabular}[c]{@{}c@{}} 37.84 \\ \cellcolor[HTML]{c7f9cc}0.964\\  \cellcolor[HTML]{c7f9cc}0.032\end{tabular} 
&  \begin{tabular}[c]{@{}c@{}}\cellcolor[HTML]{c7f9cc}30.63\\ \cellcolor[HTML]{c7f9cc}0.929\\  \cellcolor[HTML]{c7f9cc}0.056\end{tabular} 
&  \begin{tabular}[c]{@{}c@{}}\cellcolor[HTML]{c7f9cc}34.22\\ \cellcolor[HTML]{c7f9cc}0.950\\  \cellcolor[HTML]{c7f9cc}0.039\end{tabular} 
&  \begin{tabular}[c]{@{}c@{}}\cellcolor[HTML]{c7f9cc}37.71\\ \cellcolor[HTML]{c7f9cc}0.964\\  \cellcolor[HTML]{c7f9cc}0.032\end{tabular} 
&  \begin{tabular}[c]{@{}c@{}}\cellcolor[HTML]{c7f9cc}38.18\\ \cellcolor[HTML]{c7f9cc}0.964\\  \cellcolor[HTML]{c7f9cc}0.032\end{tabular} 
&  \begin{tabular}[c]{@{}c@{}}\cellcolor[HTML]{57cc99}31.83\\ \cellcolor[HTML]{57cc99}0.939\\  \cellcolor[HTML]{57cc99}0.043\end{tabular} 
&  \begin{tabular}[c]{@{}c@{}}\cellcolor[HTML]{57cc99}35.13\\ \cellcolor[HTML]{57cc99}0.957\\  \cellcolor[HTML]{57cc99}0.030\end{tabular} 
&  \begin{tabular}[c]{@{}c@{}}\cellcolor[HTML]{57cc99}38.18\\ \cellcolor[HTML]{57cc99}0.967\\  \cellcolor[HTML]{57cc99}0.026\end{tabular} 
& \begin{tabular}[c]{@{}c@{}}\cellcolor[HTML]{57cc99}38.20\\ \cellcolor[HTML]{57cc99}0.970\\ \cellcolor[HTML]{57cc99}0.026\end{tabular} \\
\bottomrule
\end{tabular}}
\caption{Comparison of rendering on the Replica dataset. The position and orientation of viewpoints used in training and evaluation are illustrated in Appendix.}
\label{tab:sparse-replica}
\end{table}

\noindent \textbf{Evaluation on Replica.}
Our proposed GeoGaussian demonstrates superior rendering performance compared to other Gaussian Splatting-based methods in the R1, R2, and OFF3 sequences, as shown in Table~\ref{tab:sparse-replica}. Particularly in the R1 and R2 sequences, our method exhibits comprehensive improvements across all three metrics. These environments consist mostly of less-textured regions, such as single-color walls, where 3DGS~\cite{kerbl3Dgaussians} and its refinement method LightGS~\cite{fan2023lightgaussian} lack sufficient photometric constraints for their optimization processes. This is evident in the first two rows of Figure~\ref{fig:qualitative}, where the ceiling appears blurry in these two methods but is well-preserved in our method. 

Furthermore, the distributions of 3D Gaussians illustrate that the geometry in our method is more reasonable compared to the results from 3DGS, with flatter walls and clearer junction lines highlighted in Figure~\ref{fig:qualitative}. 
As shown in the OFF3 sequence, our 3D Gaussian model shows more generality and robustness in rendering performance when the training data becomes sparser.
Specifically, the PSNR distance between 3DGS and ours is $0.1$ in OFF3 (100\%), while the distance becomes $0.7$ in OFF3 (10\%). 
Similar improvements are witnessed in R1, where the distance between these two methods in R1 (100\%) is around $0.6$, and this difference continues to increase in R1 (16.6\%) to $1.2$. Since our evaluation dataset includes rich scenarios, these improvements demonstrate that the proposed method has advantages in novel view rendering tasks by using geometry-aware strategies.

In Figure~\ref{fig:model-size}, we present the number of Gaussians used for each sequence. All methods 
are optimized for 30,000 iterations. LightGS received an additional 5,000 iterations based on the 25,000 models of 3DGS, following 
the official settings of LightGS. Generally, 3DGS requires more Gaussians in average compared to the other two methods. 
For example, 3DGS employs about one-fourth more Gaussians than the other methods in OFF4, yet the rendering performance remains similar across all three methods.


\vspace{3mm}
\noindent \textbf{Evaluation on TUM RGB-D.}
In Table~\ref{tab:tum-icl}, sequences from TUM RGB-D are used to evaluate Gaussian Splatting approaches. In sequences 
such as f3/cabinet and f3/strtex-far, our method outperforms 3DGS and LightGS significantly. For example, our method achieves a PSNR of $28.64$ 
while 3DGS obtains $27.85$ in f3/strtex-far. As shown in the last row of Figure~\ref{fig:qualitative}, the 3D Gaussian points in our method are well-organized in structure, whereas the geometry is not preserved in 3DGS, appearing as a big blue ellipsoid passing through the wall. In the cabinet sequence, characterized by a lack of textured regions, our proposed method demonstrates impressive improvements in both sparse (10\%) and full (100\%) training data scenarios.
However, in the lag-cabinet sequences where the camera moves around a cabinet placed in the middle of the scene, both 3DGS and LightGS perform better than our GeoGaussian.

\begin{table}[]
\resizebox{\linewidth}{!}{
\begin{tabular}{l|ccc|ccc|ccc}
\toprule
\multicolumn{1}{c|}{Method}   & \multicolumn{3}{c|}{3DGS~\cite{kerbl3Dgaussians}}  & \multicolumn{3}{c|}{LightGS~\cite{fan2023lightgaussian}}  & \multicolumn{3}{c}{GeoGaussian}         \\ \hline
\multicolumn{1}{c|}{Metric}   & \multicolumn{1}{c}{PSNR$\uparrow$} &  SSIM$\uparrow$ & LPIPS$\downarrow$ & \multicolumn{1}{c}{PSNR$\uparrow$} &  SSIM$\uparrow$ & LPIPS$\downarrow$ & \multicolumn{1}{c}{PSNR$\uparrow$} &  SSIM$\uparrow$ & LPIPS$\downarrow$ \\ \hline
f3/cabinet (100\%)  &27.24 &0.907 & 0.125& \cellcolor[HTML]{c7f9cc}27.35 &\cellcolor[HTML]{c7f9cc} 0.908 &\cellcolor[HTML]{c7f9cc}0.124      &\cellcolor[HTML]{57cc99}28.17 &\cellcolor[HTML]{57cc99}0.916 &\cellcolor[HTML]{57cc99}0.106       \\
f3/cabinet (10\%) &\cellcolor[HTML]{c7f9cc}22.43 &\cellcolor[HTML]{c7f9cc}0.850 &\cellcolor[HTML]{c7f9cc}0.186  & 22.36 &0.849 &0.190  &\cellcolor[HTML]{57cc99}25.59 &\cellcolor[HTML]{57cc99}0.887 &\cellcolor[HTML]{57cc99}0.138 \\
f3/strtex-far (100\%) &27.85 &\cellcolor[HTML]{c7f9cc}0.896 &\cellcolor[HTML]{c7f9cc}0.073  &\cellcolor[HTML]{c7f9cc}27.90 &\cellcolor[HTML]{c7f9cc}0.896 &\cellcolor[HTML]{c7f9cc}0.073  &\cellcolor[HTML]{57cc99}28.64 &\cellcolor[HTML]{57cc99}0.907 &\cellcolor[HTML]{57cc99}0.064   \\
f3/strtex-far (10\%) &\cellcolor[HTML]{c7f9cc}21.41 &\cellcolor[HTML]{c7f9cc}0.737 &\cellcolor[HTML]{c7f9cc}0.192 & 21.39 &0.734 &0.194  &\cellcolor[HTML]{57cc99}22.84 &\cellcolor[HTML]{57cc99}0.779 &\cellcolor[HTML]{57cc99}0.137 \\
\multicolumn{1}{c|}{Avg.} &24.73 &\cellcolor[HTML]{c7f9cc}0.848 &\cellcolor[HTML]{c7f9cc}0.144  &\cellcolor[HTML]{c7f9cc}24.75 &0.847 &0.145 &\cellcolor[HTML]{57cc99}26.31 &\cellcolor[HTML]{57cc99}0.872 &\cellcolor[HTML]{57cc99}0.111 \\
\bottomrule
\end{tabular}}
\caption{Rendering performance comparison on the TUM RGB-D datasets. }
\label{tab:tum-icl}
\end{table}

\subsection{Ablation Studies}
\noindent \textbf{Sparse Views for Training.}
In this section, we present additional settings for sparse view rendering evaluation as listed in Table~\ref{tab:sparse_view_new}. As the number of training images decreases from 50\% to 10\%, the performance of 3DGS deteriorates significantly. In contrast, the GeoGaussian approach, leveraging the proposed geometry-aware densification operations and constraints, demonstrates more robust rendering performance.

\begin{table}[]
\centering
\resizebox{.98\linewidth}{!}{%
\begin{tabular}{@{}ll|cccccc@{}}
\toprule
Method & Metric 
& \begin{tabular}[c]{@{}c@{}}10\%\\(train/test) \end{tabular}  
& \begin{tabular}[c]{@{}c@{}}12.5\%\\(train/test) \end{tabular} 
&\begin{tabular}[c]{@{}c@{}}16.6\%\\(train/test) \end{tabular} 
& \begin{tabular}[c]{@{}c@{}}25\%\\(train/test) \end{tabular}  
& \begin{tabular}[c]{@{}c@{}}50\%\\(train/test) \end{tabular} \\ \hline
 \begin{tabular}[c]{@{}l@{}} \\3DGS~\cite{kerbl3Dgaussians}\\ \end{tabular} & \begin{tabular}[r]{@{}l@{}}PSNR$\uparrow$\\ SSIM$\uparrow$\\ LPIPS$\downarrow$\end{tabular} 
& \begin{tabular}[c]{@{}c@{}} 30.49\\  0.932\\ 0.051 \end{tabular} 
& \begin{tabular}[c]{@{}c@{}} 33.15\\ 0.944\\ 0.039 \end{tabular} 
& \begin{tabular}[c]{@{}c@{}} 33.98\\ 0.951\\  0.035\end{tabular} 
& \begin{tabular}[c]{@{}c@{}} 36.58\\ 0.961\\ 0.028\end{tabular} 
& \begin{tabular}[c]{@{}c@{}} 37.45 \\0.964\\  0.028\end{tabular}  \\
\hline
\begin{tabular}[c]{@{}l@{}}GeoGaussian\\(ours)\end{tabular} 
& \begin{tabular}[r]{@{}l@{}}PSNR$\uparrow$\\ SSIM$\uparrow$\\ LPIPS$\downarrow$\end{tabular} 
& \begin{tabular}[c]{@{}c@{}} \cellcolor[HTML]{57cc99}31.65\\ \cellcolor[HTML]{57cc99}0.942\\ \cellcolor[HTML]{57cc99}0.041 \end{tabular} 
& \begin{tabular}[c]{@{}c@{}} \cellcolor[HTML]{57cc99}33.95\\ \cellcolor[HTML]{57cc99}0.949\\ \cellcolor[HTML]{57cc99}0.032 \end{tabular} 
& \begin{tabular}[c]{@{}c@{}} \cellcolor[HTML]{57cc99}35.17\\ \cellcolor[HTML]{57cc99}0.957\\ \cellcolor[HTML]{57cc99}0.027 \end{tabular} 
& \begin{tabular}[c]{@{}c@{}} \cellcolor[HTML]{57cc99}36.83\\ \cellcolor[HTML]{57cc99}0.964 \\ \cellcolor[HTML]{57cc99}0.023 \end{tabular} 
& \begin{tabular}[c]{@{}c@{}} \cellcolor[HTML]{57cc99}38.00 \\ \cellcolor[HTML]{57cc99}0.968 \\ \cellcolor[HTML]{57cc99}0.022\end{tabular} \\
\bottomrule
\end{tabular}}
\caption{Sparse view rendering on the R1 sequence of Replica dataset. The position and orientation of viewpoints used in training and evaluation are given in Appendix. }
\label{tab:sparse_view_new}
\end{table}

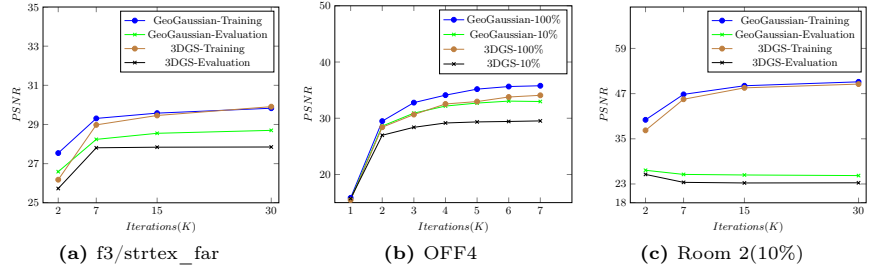
\begin{figure}
     \centering
    \subfloat[f3/strtex\_far]{
    \resizebox{0.30\linewidth}{!}{
   \begin{tikzpicture}
   \begin{axis}[
       xlabel=$Iterations (K)$,
       ylabel=$PSNR$,
       xmin=0, xmax=31,
       ymin=25, ymax=35,
       xtick={2,7,15,30},
       ytick={25,27,...,35}]
   \addplot[mark=*,blue] plot coordinates {
     (2,  27.54)
     (7,  29.31)
     (15,  29.58)
     (30,  29.83)};
  \addlegendentry{GeoGaussian-Training}
   \addplot[color=green,mark=x]
       plot coordinates {
       (2,  26.59)
       (7,  28.24)
       (15,  28.55)
       (30,  28.70)};
   \addlegendentry{GeoGaussian-Evaluation}
   \addplot[mark=*,brown] plot coordinates {
       (2,  26.18)
       (7,  28.98)
       (15, 29.46)
       (30, 29.91)};
  \addlegendentry{3DGS-Training}
   \addplot[color=black,mark=x]
       plot coordinates {
       (2,  25.73)
       (7,  27.81)
       (15, 27.84)
       (30, 27.85)};
   \addlegendentry{3DGS-Evaluation}
   \end{axis}
   \end{tikzpicture}}
    \label{subfig:short-10-c}}
   \subfloat[OFF4]{
    \resizebox{0.3\linewidth}{!}{
   \begin{tikzpicture}
   \begin{axis}[
       xlabel=$Iterations (K)$,
       ylabel=$PSNR$,
       xmin=0.5, xmax=8,
       ymin=15, ymax=50,
       xtick={0.01,1,2,3,4,5,6,7},
       ytick={0,10,20,30,40,50}]
   \addplot[mark=*,blue] plot coordinates {
     (1, 15.85)
     (2,  29.49)
     (3,  32.78)
     (4,  34.11)
     (5,  35.18)
     (6,  35.64)
     (7,  35.78)};
  \addlegendentry{GeoGaussian-100\%}
   \addplot[color=green,mark=x]
       plot coordinates {
    (1, 15.56)
     (2, 28.61)
     (3,  30.91)
     (4,  32.17)
     (5,  32.70)
     (6,  33.05)
     (7,  32.98)};
   \addlegendentry{GeoGaussian-10\%}
   \addplot[mark=*,brown] plot coordinates {
      (1, 15.47)
     (2, 28.41)
     (3,  30.66)
     (4,  32.54)
     (5,  32.97)
     (6,  33.78)
     (7,  34.08)};
  \addlegendentry{3DGS-100\%}
   \addplot[color=black,mark=x]
       plot coordinates {
     (1, 15.72)
     (2, 26.98)
     (3,  28.39)
     (4,  29.16)
     (5,  29.35)
     (6,  29.43)
     (7,  29.53)};
   \addlegendentry{3DGS-10\%}
   \end{axis}
   \end{tikzpicture}}
   \label{subfig:short-10-d}}
      \subfloat[Room 2(10\%)]{
    \label{subfig:short-10}
    \resizebox{0.3\linewidth}{!}{
   \begin{tikzpicture}
   \begin{axis}[
       xlabel=$Iterations (K)$,
       ylabel=$PSNR$,
       xmin=0, xmax=31,
       ymin=18, ymax=70,
       xtick={2,7,15,30},
       ytick={18,23,35,...,70}]
   \addplot[mark=*,blue] plot coordinates {
     (2,  40.04)
     (7,  46.80)
     (15,  49.09)
     (30,  50.13)};
  \addlegendentry{GeoGaussian-Training}
   \addplot[color=green,mark=x]
       plot coordinates {
       (2,  26.64)
       (7,  25.54)
       (15,  25.39)
       (30,  25.23)};
   \addlegendentry{GeoGaussian-Evaluation}
   \addplot[mark=*,brown] plot coordinates {
       (2,  37.24)
       (7,  45.50)
       (15, 48.53)
       (30, 49.53)};
  \addlegendentry{3DGS-Training}
   \addplot[color=black,mark=x]
       plot coordinates {
       (2,  25.54)
       (7,  23.43)
       (15, 23.27)
       (30, 23.31)};
   \addlegendentry{3DGS-Evaluation}
   \end{axis}
   \end{tikzpicture}}}
     \caption{Rendering performance in training and evaluation using TUM RGB-D (a), Replica (b), and ICL-NUIM (c) datasets.}
     \label{fig:training}
 \end{figure}

\vspace{3mm}
\noindent \textbf{Training and Evaluation.}
Figure~\ref{fig:training} illustrates the PSNR results collected during training and evaluation iterations. In Figure \ref{subfig:short-10-c}, our GeoGaussian achieves significantly better results in the first 15K iterations than 3DGS. This improvement can be attributed to the accuracy provided by our proposed initialization and densification modules, which help to achieve better convergence. Similar phenomena can be seen in Figure~\ref{subfig:short-10-d}, the performance of GeoGaussian is more robust in both sparse (10\%) and full (100\%) training data scenarios. In Figure~\ref{subfig:short-10}, 3DGS tends to overfit the training views, while our approach limits the extent of overfitting and prevents test results from degrading rapidly.

\section{Conclusion}


In this paper, we introduce a novel approach called GeoGaussian, which emphasizes the importance of preserving accurate geometry in Gaussian models to enhance their representation in 3D space. Our method first gives Gaussian parameters a clearer geometric meaning. The third parameter of the scale vector is used to control thickness, while the third column of the rotation matrix specifies the normal direction of the thin ellipsoid, where the initial normal direction is extracted from the initial point clouds.
Additionally, we propose a carefully designed densification approach to effectively organize the newly generated ellipsoids. In the optimization module, we encourage ellipsoids in the neighborhood to lie in a co-planar area, further enhancing the representation quality.
Experimental results on public datasets demonstrate that our method achieves superior performance in terms of geometry accuracy and photo-realistic novel view rendering compared to state-of-the-art approaches.
In the future, we plan to explore more comprehensive solutions for improving the geometry of Gaussian models. This will involve incorporating depth, normals, and camera poses into the 3D Gaussian Splatting optimization process, reducing the reliance on point cloud normal vectors.

\vspace{3mm}
\noindent \textbf{Acknowledgement.} This research work is supported by the Agency for Science, Technology and Research (A*STAR) under its MTC Programmatic Funds (Grant No. M23L7b0021).

\clearpage

\appendix
\setcounter{table}{4}
\setcounter{figure}{5}

\section{Novel View Rendering}
In the evaluation section, we present a comprehensive set of experiments to evaluate the performance of our method. Here, we provide additional quantitative and qualitative results to further demonstrate the effectiveness of our approach.   

\subsection{Replica Sequences}
In Table \textcolor{red}{2} (see Section \textcolor{red}{4.1}), we compare novel view rendering performance using the R1, R2, OFF3, and OFF4 sequences. Additionally, we conduct experiments on four additional sequences (R0, OFF0, OFF1, and OFF2) in this section. Following the same setup as described in the section of \textit{Gaussian Splatting in NVS} (see Section \textcolor{red}{4.4}), we evaluate the rendering performance of 3DGS, LightGS, and GeoGaussian using three metrics, as shown in Table~\ref{tab:sparse-replica-appen}.

\begin{table}[]
\centering
\resizebox{.95\linewidth}{!}{%
\begin{tabular}{ll|cccc|cccc|cccc}
\toprule
\multicolumn{2}{c|}{methods} & \multicolumn{4}{c|}{3DGS} & \multicolumn{4}{c|}{LightGS} & \multicolumn{4}{c}{GeoGaussian} \\
\multicolumn{2}{c|}{data} & 10\% & 16.6\% & 50\% & 100\% & 10\% & 16.6\% & 50\% & 100\% & 10\% & 16.6\% & 50\% & 100\% \\ \hline

R0 & \begin{tabular}[r]{@{}r@{}}PSNR$\uparrow$\\ SSIM$\uparrow$\\ LPIPS$\downarrow$\end{tabular} 
& \begin{tabular}[c]{@{}c@{}} \cellcolor[HTML]{c7f9cc}27.24 \\ \cellcolor[HTML]{c7f9cc}0.875 \\ \cellcolor[HTML]{c7f9cc}0.070\end{tabular}
& \begin{tabular}[c]{@{}c@{}} 31.14 \\ 0.928 \\ \cellcolor[HTML]{c7f9cc}0.040\end{tabular}
& \begin{tabular}[c]{@{}c@{}} 34.54 \\ \cellcolor[HTML]{c7f9cc}0.949 \\ \cellcolor[HTML]{c7f9cc}0.031\end{tabular}
& \begin{tabular}[c]{@{}c@{}} 35.10 \\ \cellcolor[HTML]{c7f9cc}0.952 \\ 0.032 \end{tabular}

& \begin{tabular}[c]{@{}c@{}} 27.23 \\\cellcolor[HTML]{c7f9cc}0.875\\ \cellcolor[HTML]{c7f9cc}0.070 \end{tabular}
& \begin{tabular}[c]{@{}c@{}} \cellcolor[HTML]{c7f9cc}31.21 \\\cellcolor[HTML]{c7f9cc}0.929\\ \cellcolor[HTML]{c7f9cc}0.040 \end{tabular}
& \begin{tabular}[c]{@{}c@{}} \cellcolor[HTML]{57cc99}34.68 \\\cellcolor[HTML]{57cc99}0.950\\ \cellcolor[HTML]{c7f9cc}0.031 \end{tabular}
& \begin{tabular}[c]{@{}c@{}} \cellcolor[HTML]{57cc99}35.28 \\\cellcolor[HTML]{57cc99}0.953\\ \cellcolor[HTML]{c7f9cc}0.031 \end{tabular}

& \begin{tabular}[c]{@{}c@{}} \cellcolor[HTML]{57cc99}28.84 \\ \cellcolor[HTML]{57cc99}0.899 \\ \cellcolor[HTML]{57cc99}0.057\end{tabular}
& \begin{tabular}[c]{@{}c@{}} \cellcolor[HTML]{57cc99}31.77 \\ \cellcolor[HTML]{57cc99}0.931 \\ \cellcolor[HTML]{57cc99}0.036\end{tabular}
& \begin{tabular}[c]{@{}c@{}} \cellcolor[HTML]{c7f9cc}34.66 \\ \cellcolor[HTML]{c7f9cc}0.949 \\ \cellcolor[HTML]{57cc99}0.028\end{tabular}  
& \begin{tabular}[c]{@{}c@{}} \cellcolor[HTML]{c7f9cc}35.23 \\ \cellcolor[HTML]{c7f9cc}0.952 \\ \cellcolor[HTML]{57cc99}0.029\end{tabular} \\

OFF0 & \begin{tabular}[r]{@{}r@{}}PSNR$\uparrow$\\ SSIM$\uparrow$ \\ LPIPS$\downarrow$\end{tabular} 
& \begin{tabular}[c]{@{}c@{}} 31.61 \\ \cellcolor[HTML]{57cc99}0.927 \\ \cellcolor[HTML]{c7f9cc}0.062 \end{tabular} 
& \begin{tabular}[c]{@{}c@{}} 35.99 \\ \cellcolor[HTML]{c7f9cc}0.959 \\ \cellcolor[HTML]{c7f9cc}0.034 \end{tabular} 
& \begin{tabular}[c]{@{}c@{}} 41.70 \\ \cellcolor[HTML]{c7f9cc}0.979 \\ \cellcolor[HTML]{c7f9cc}0.018 \end{tabular} 
& \begin{tabular}[c]{@{}c@{}} 42.67 \\ \cellcolor[HTML]{57cc99}0.981 \\ \cellcolor[HTML]{c7f9cc}0.017 \end{tabular} 

& \begin{tabular}[c]{@{}c@{}} \cellcolor[HTML]{c7f9cc}31.63 \\\cellcolor[HTML]{57cc99}0.927\\ \cellcolor[HTML]{c7f9cc}0.062 \end{tabular}
& \begin{tabular}[c]{@{}c@{}} \cellcolor[HTML]{c7f9cc}36.00 \\\cellcolor[HTML]{c7f9cc}0.959\\ \cellcolor[HTML]{c7f9cc}0.034 \end{tabular}
& \begin{tabular}[c]{@{}c@{}} \cellcolor[HTML]{c7f9cc}41.82 \\\cellcolor[HTML]{c7f9cc}0.979\\ \cellcolor[HTML]{c7f9cc}0.018 \end{tabular}
& \begin{tabular}[c]{@{}c@{}} \cellcolor[HTML]{57cc99}43.08 \\\cellcolor[HTML]{57cc99}0.981\\ \cellcolor[HTML]{57cc99}0.016 \end{tabular}

& \begin{tabular}[c]{@{}c@{}} \cellcolor[HTML]{57cc99}32.42 \\  \cellcolor[HTML]{c7f9cc}0.924 \\  \cellcolor[HTML]{57cc99}0.053 \end{tabular} 
& \begin{tabular}[c]{@{}c@{}} \cellcolor[HTML]{57cc99}36.34 \\  \cellcolor[HTML]{57cc99}0.960 \\  \cellcolor[HTML]{57cc99}0.030 \end{tabular} 
& \begin{tabular}[c]{@{}c@{}} \cellcolor[HTML]{57cc99}42.11 \\  \cellcolor[HTML]{57cc99}0.980 \\  \cellcolor[HTML]{57cc99}0.016 \end{tabular} 
& \begin{tabular}[c]{@{}c@{}} \cellcolor[HTML]{c7f9cc}42.74 \\ \cellcolor[HTML]{57cc99} 0.981 \\  \cellcolor[HTML]{c7f9cc}0.017 \end{tabular}  \\

OFF1 & \begin{tabular}[r]{@{}r@{}}PSNR$\uparrow$\\ SSIM$\uparrow$\\ LPIPS$\downarrow$\end{tabular} 
& \begin{tabular}[c]{@{}c@{}}34.81\\  \cellcolor[HTML]{57cc99}0.943 \\  \cellcolor[HTML]{57cc99}0.060\end{tabular} 
& \begin{tabular}[c]{@{}c@{}}\cellcolor[HTML]{c7f9cc}38.39\\  \cellcolor[HTML]{57cc99}0.956 \\   \cellcolor[HTML]{c7f9cc}0.046\end{tabular} 
& \begin{tabular}[c]{@{}c@{}}\cellcolor[HTML]{c7f9cc}42.27\\   \cellcolor[HTML]{57cc99}0.972 \\  \cellcolor[HTML]{57cc99}0.036\end{tabular} 
& \begin{tabular}[c]{@{}c@{}}\cellcolor[HTML]{c7f9cc}42.88\\   \cellcolor[HTML]{57cc99}0.974 \\  \cellcolor[HTML]{57cc99}0.035\end{tabular} 

& \begin{tabular}[c]{@{}c@{}} \cellcolor[HTML]{c7f9cc}34.83 \\\cellcolor[HTML]{57cc99}0.943\\ \cellcolor[HTML]{57cc99}0.060 \end{tabular}
& \begin{tabular}[c]{@{}c@{}} \cellcolor[HTML]{57cc99}38.43 \\\cellcolor[HTML]{57cc99}0.956\\ \cellcolor[HTML]{57cc99}0.045 \end{tabular}
& \begin{tabular}[c]{@{}c@{}} \cellcolor[HTML]{57cc99}42.43 \\\cellcolor[HTML]{57cc99}0.972\\ \cellcolor[HTML]{57cc99}0.036 \end{tabular}
& \begin{tabular}[c]{@{}c@{}} \cellcolor[HTML]{57cc99}43.01 \\\cellcolor[HTML]{57cc99}0.974\\ \cellcolor[HTML]{57cc99}0.035 \end{tabular}

& \begin{tabular}[c]{@{}c@{}} \cellcolor[HTML]{57cc99}34.86\\  \cellcolor[HTML]{c7f9cc}0.939\\   \cellcolor[HTML]{c7f9cc}0.061\end{tabular}
& \begin{tabular}[c]{@{}c@{}} 38.35\\  \cellcolor[HTML]{c7f9cc}0.952\\   \cellcolor[HTML]{57cc99}0.045\end{tabular} 
& \begin{tabular}[c]{@{}c@{}} 41.89\\  \cellcolor[HTML]{c7f9cc}0.968\\   \cellcolor[HTML]{c7f9cc}0.039\end{tabular} 
& \begin{tabular}[c]{@{}c@{}} 42.17\\  \cellcolor[HTML]{c7f9cc}0.970\\  \cellcolor[HTML]{c7f9cc}0.039\end{tabular} \\

OFF2 & \begin{tabular}[r]{@{}r@{}}PSNR$\uparrow$\\ SSIM$\uparrow$\\ LPIPS$\downarrow$\end{tabular} 
& \begin{tabular}[c]{@{}c@{}} \cellcolor[HTML]{c7f9cc}28.27\\ \cellcolor[HTML]{c7f9cc}0.911\\   0.074\end{tabular} 
& \begin{tabular}[c]{@{}c@{}}\cellcolor[HTML]{c7f9cc}32.91\\  \cellcolor[HTML]{c7f9cc}0.948\\   \cellcolor[HTML]{c7f9cc}0.043\end{tabular} 
& \begin{tabular}[c]{@{}c@{}}36.72\\  0.964\\  0.033\end{tabular} 
& \begin{tabular}[c]{@{}c@{}} \cellcolor[HTML]{c7f9cc}37.34 \\  0.966\\   \cellcolor[HTML]{c7f9cc}0.033\end{tabular} 

& \begin{tabular}[c]{@{}c@{}} 28.25 \\\cellcolor[HTML]{c7f9cc}0.911\\ \cellcolor[HTML]{c7f9cc}0.073 \end{tabular}
& \begin{tabular}[c]{@{}c@{}} 32.85 \\\cellcolor[HTML]{c7f9cc}0.948\\ 0.045 \end{tabular}
& \begin{tabular}[c]{@{}c@{}} \cellcolor[HTML]{57cc99}36.95 \\\cellcolor[HTML]{c7f9cc}0.965\\ \cellcolor[HTML]{c7f9cc}0.032 \end{tabular}
& \begin{tabular}[c]{@{}c@{}} \cellcolor[HTML]{57cc99}37.49 \\\cellcolor[HTML]{c7f9cc}0.967\\ \cellcolor[HTML]{c7f9cc}0.033 \end{tabular}

& \begin{tabular}[c]{@{}c@{}}  \cellcolor[HTML]{57cc99}29.52 \\  \cellcolor[HTML]{57cc99}0.925 \\  \cellcolor[HTML]{57cc99}0.054\end{tabular}
& \begin{tabular}[c]{@{}c@{}}  \cellcolor[HTML]{57cc99}33.97 \\  \cellcolor[HTML]{57cc99}0.955 \\  \cellcolor[HTML]{57cc99}0.032\end{tabular}
& \begin{tabular}[c]{@{}c@{}}  \cellcolor[HTML]{c7f9cc}36.90 \\  \cellcolor[HTML]{57cc99}0.968 \\  \cellcolor[HTML]{57cc99}0.028\end{tabular}
& \begin{tabular}[c]{@{}c@{}}  37.32 \\  \cellcolor[HTML]{57cc99}0.970 \\  \cellcolor[HTML]{57cc99}0.029\end{tabular}  \\
 \hline

Avg. & \begin{tabular}[r]{@{}r@{}}PSNR$\uparrow$\\ SSIM$\uparrow$\\ LPIPS$\downarrow$\end{tabular} 
& \begin{tabular}[c]{@{}c@{}} 30.48 \\\cellcolor[HTML]{c7f9cc}0.914\\ 0.067 \end{tabular}
& \begin{tabular}[c]{@{}c@{}} 34.61 \\\cellcolor[HTML]{c7f9cc}0.948\\ \cellcolor[HTML]{c7f9cc}0.041 \end{tabular}
& \begin{tabular}[c]{@{}c@{}} 38.81 \\\cellcolor[HTML]{c7f9cc}0.966\\ 0.030 \end{tabular}
& \begin{tabular}[c]{@{}c@{}} \cellcolor[HTML]{c7f9cc}39.50 \\\cellcolor[HTML]{c7f9cc}0.968\\ \cellcolor[HTML]{57cc99}0.029 \end{tabular}

& \begin{tabular}[c]{@{}c@{}} \cellcolor[HTML]{c7f9cc}30.49 \\\cellcolor[HTML]{c7f9cc}0.914\\ \cellcolor[HTML]{c7f9cc}0.066 \end{tabular}
& \begin{tabular}[c]{@{}c@{}} \cellcolor[HTML]{c7f9cc}34.87 \\\cellcolor[HTML]{c7f9cc}0.948\\ \cellcolor[HTML]{c7f9cc}0.041 \end{tabular}
& \begin{tabular}[c]{@{}c@{}} \cellcolor[HTML]{57cc99}38.97 \\\cellcolor[HTML]{57cc99}0.967\\ \cellcolor[HTML]{c7f9cc}0.029 \end{tabular}
& \begin{tabular}[c]{@{}c@{}} \cellcolor[HTML]{57cc99}39.72 \\\cellcolor[HTML]{57cc99}0.969\\ \cellcolor[HTML]{57cc99}0.029 \end{tabular}

& \begin{tabular}[c]{@{}c@{}} \cellcolor[HTML]{57cc99}31.41 \\\cellcolor[HTML]{57cc99}0.922\\ \cellcolor[HTML]{57cc99}0.056 \end{tabular}
& \begin{tabular}[c]{@{}c@{}} \cellcolor[HTML]{57cc99}35.11 \\\cellcolor[HTML]{57cc99}0.950\\ \cellcolor[HTML]{57cc99}0.036 \end{tabular}
& \begin{tabular}[c]{@{}c@{}} \cellcolor[HTML]{c7f9cc}38.89 \\\cellcolor[HTML]{c7f9cc}0.966\\ \cellcolor[HTML]{57cc99}0.028 \end{tabular}
& \begin{tabular}[c]{@{}c@{}} 39.37 \\\cellcolor[HTML]{c7f9cc}0.968\\ \cellcolor[HTML]{57cc99}0.029 \end{tabular}   \\

\bottomrule
\end{tabular}}
\caption{Comparison of rendering on the Replica dataset.}
\label{tab:sparse-replica-appen}
\end{table}

\begin{figure}
    \centering
     \resizebox{0.04\linewidth}{!}{
     	\begin{tikzpicture}
        \draw [white] (0,-1.5) -- (0,2.5);
        \node[rotate=270,font=\bf] at (0.2, 0.7) {GeoGaussian 3DGS};
    	\end{tikzpicture}}
    \begin{subfigure}{0.23\linewidth}
    \includegraphics[width=\linewidth]{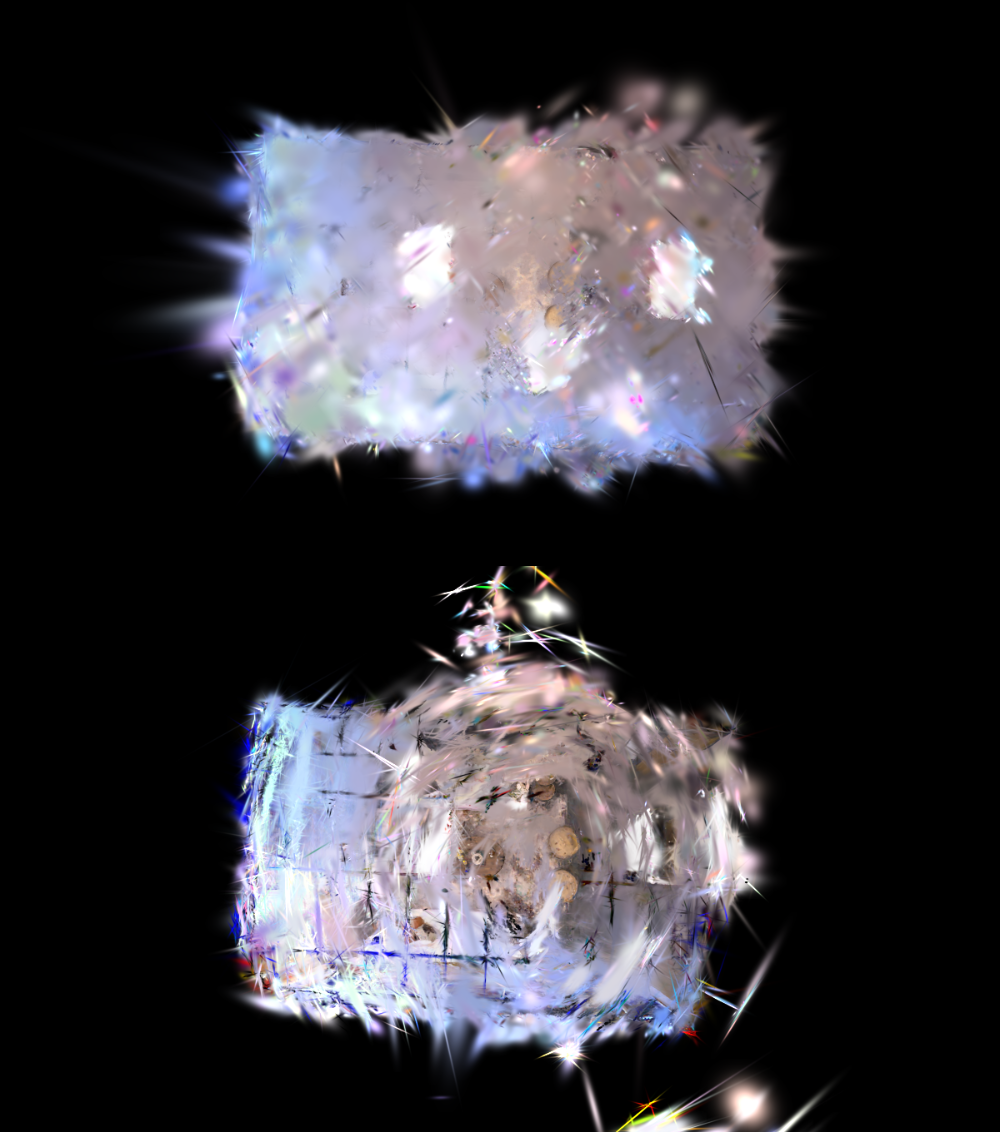}
    \caption{R0 (100\%)}
    \label{fig:short-a}
  \end{subfigure}
  \begin{subfigure}{0.23\linewidth}
    \includegraphics[width=\linewidth]{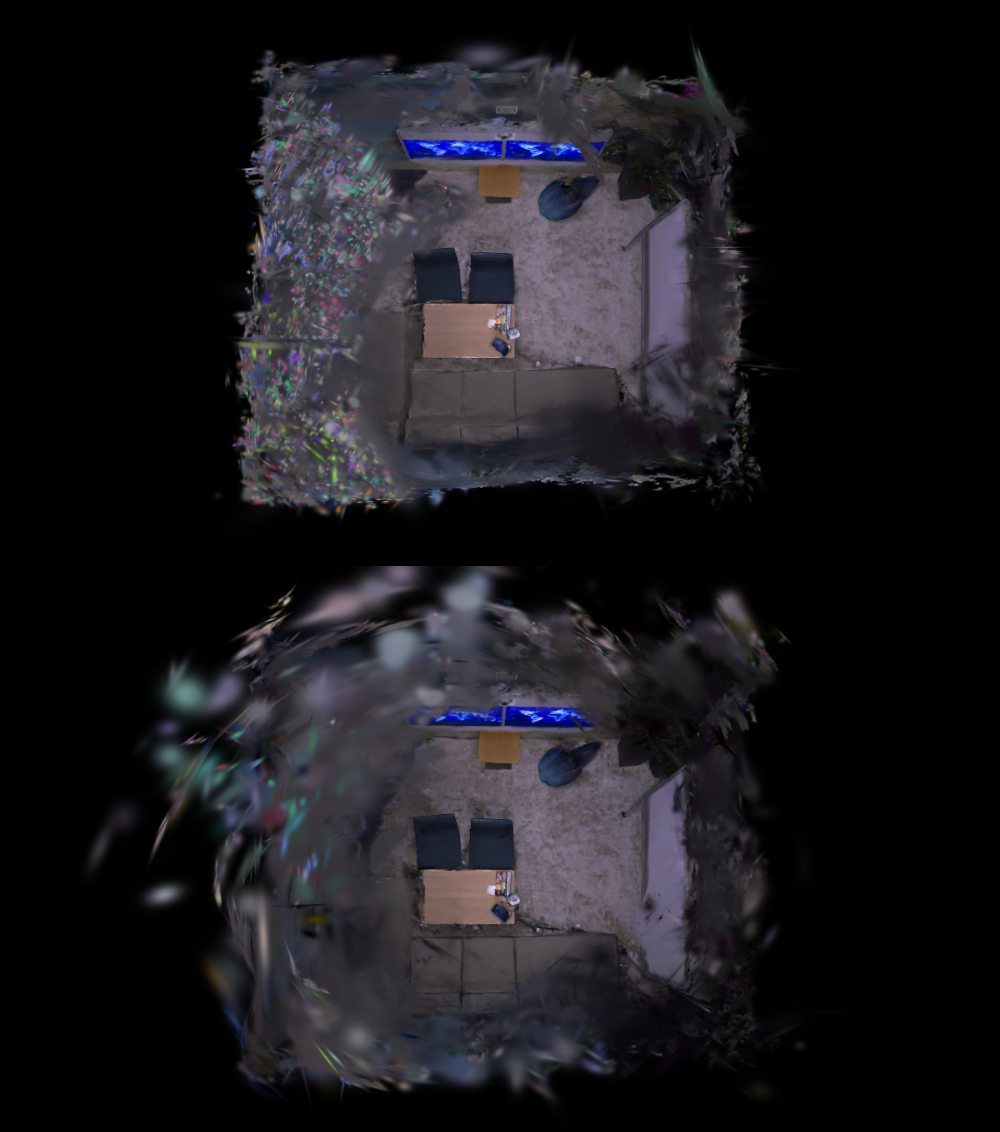}
    \caption{OFF0(100\%)}
    \label{fig:short-b}
  \end{subfigure}
  \begin{subfigure}{0.23\linewidth}
    \includegraphics[width=\linewidth]{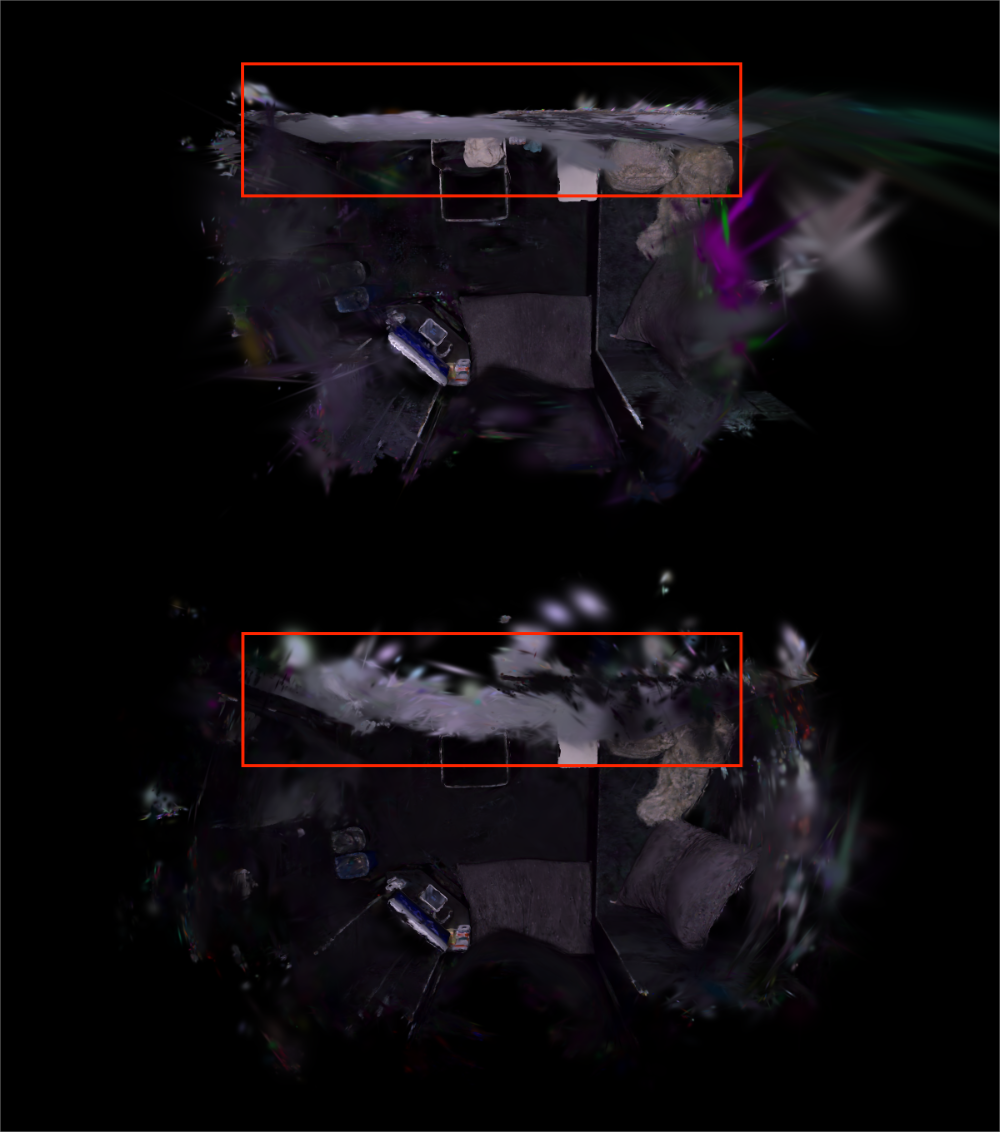}
    \caption{OFF1(100\%)}
    \label{fig:short-c}
  \end{subfigure}
    \begin{subfigure}{0.23\linewidth}
    \includegraphics[width=\linewidth]{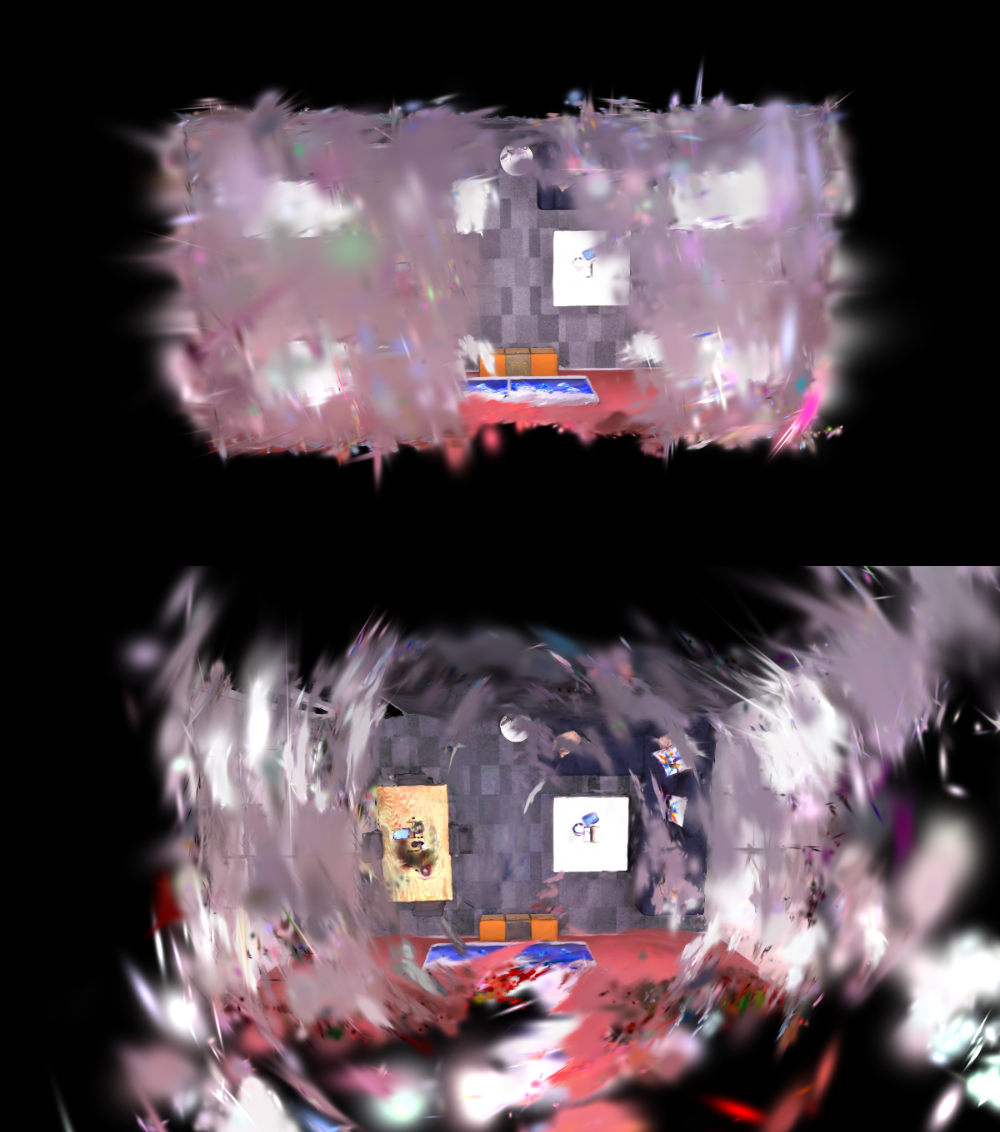}
    \caption{OFF2(100\%)}
    \label{fig:short-d}
  \end{subfigure}
    \caption{Gaussian models generated by GeoGaussian (ours) and 3DGS on the Replica sequences.}
    \label{fig:replica-models-appen}
\end{figure}
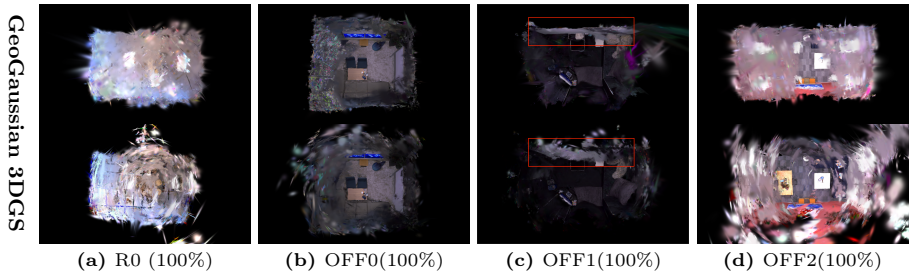

The 3D Gaussian models obtained by GeoGaussian and 3DGS are represented in Figure~\ref{fig:replica-models-appen}. Benefiting from the proposed densification method and geometric constraints, our method, GeoGaussian, preserves the reasonable geometry of environments. For example, the walls in Figure~\ref{fig:short-b} and \ref{fig:short-c} are very thin, but the geometry is very noisy in the corresponding models of 3DGS.

\begin{figure}
    \centering
    \begin{subfigure}{0.22\linewidth}
    \includegraphics[width=\linewidth]{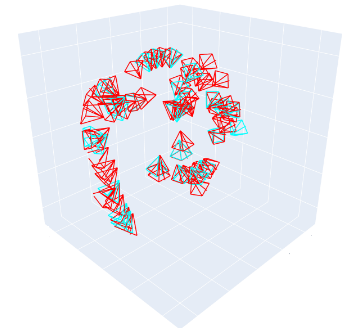}
    \caption{OFF3 (10\%)}
    \label{fig:short-model-a}
  \end{subfigure}
  \begin{subfigure}{0.22\linewidth}
    \includegraphics[width=\linewidth]{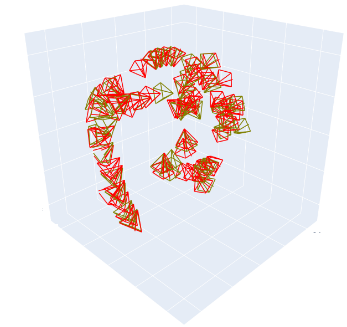}
    \caption{OFF3 (16.6\%)}
    \label{fig:short-model-b}
  \end{subfigure}
  \begin{subfigure}{0.22\linewidth}
    \includegraphics[width=\linewidth]{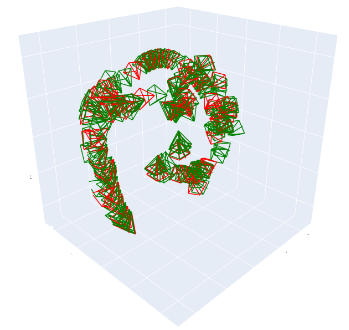}
    \caption{OFF3 (50\%)}
    \label{fig:short-model-c}
  \end{subfigure}
  \begin{subfigure}{0.22\linewidth}
    \includegraphics[width=\linewidth]{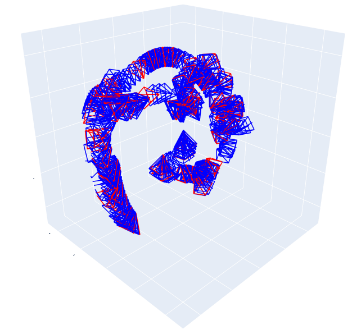}
    \caption{OFF3 (100\%)}
    \label{fig:short-model-d}
  \end{subfigure}
    \caption{Viewpoints are used during training and evaluation, where the red camera is used for evaluation.}
    \label{fig:sparse-view-appen}
\end{figure}

\paragraph{\textbf{Viewpoints in training and evaluation.}}
In Table \textcolor{red}{2}, we also evaluate the relationship between rendering performance and the sparsity of training views. Therefore, we visualize the position and orientation of viewpoints used in training and evaluation in Figure~\ref{fig:sparse-view-appen}.

As shown in Figure~\ref{fig:sparse-view-appen}, when the views used in training become sparse from OFF3(100\%) to OFF3(10\%), the scenarios seen from the training frames are difficult to cover from evaluation viewpoints. Therefore, the structure of the models plays an important role in maintaining the rendering performance as illustrated in Table \textcolor{red}{2} (see Section \textcolor{red}{4.4}).

\subsection{ICL-NUIM Sequences}

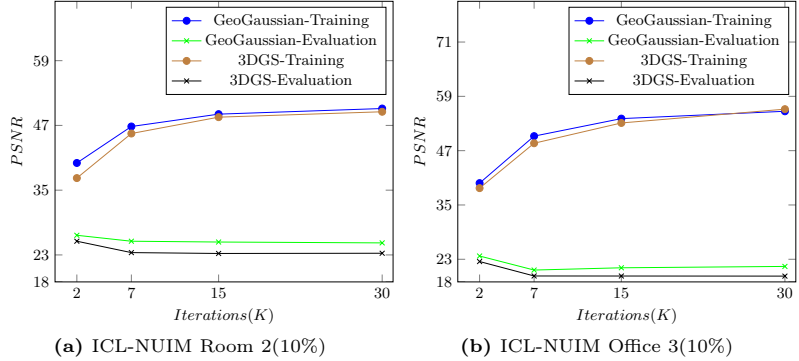
\begin{figure}
    \centering
    \subfloat[ICL-NUIM Room 2(10\%)]{
    \label{subfig:short-10-a}
    \resizebox{0.43\linewidth}{!}{
   \begin{tikzpicture}
   \begin{axis}[
       xlabel=$Iterations (K)$,
       ylabel=$PSNR$,
       xmin=0, xmax=31,
       ymin=18, ymax=70,
       xtick={2,7,15,30},
       ytick={18,23,35,...,70}]
   \addplot[mark=*,blue] plot coordinates {
     (2,  40.04)
     (7,  46.80)
     (15,  49.09)
     (30,  50.13)};
  \addlegendentry{GeoGaussian-Training}
   \addplot[color=green,mark=x]
       plot coordinates {
       (2,  26.64)
       (7,  25.54)
       (15,  25.39)
       (30,  25.23)};
   \addlegendentry{GeoGaussian-Evaluation}
   \addplot[mark=*,brown] plot coordinates {
       (2,  37.24)
       (7,  45.50)
       (15, 48.53)
       (30, 49.53)};
  \addlegendentry{3DGS-Training}
   \addplot[color=black,mark=x]
       plot coordinates {
       (2,  25.54)
       (7,  23.43)
       (15, 23.27)
       (30, 23.31)};
   \addlegendentry{3DGS-Evaluation}
   \end{axis}
   \end{tikzpicture}}}
    \subfloat[ICL-NUIM Office 3(10\%)]{
    \resizebox{0.43\linewidth}{!}{
   \begin{tikzpicture}
   \begin{axis}[
       xlabel=$Iterations (K)$,
       ylabel=$PSNR$,
       xmin=0, xmax=31,
       ymin=18, ymax=80,
       xtick={2,7,15,30},
       ytick={18,23,35,...,80}]
   \addplot[mark=*,blue] plot coordinates {
     (2,  39.83)
     (7,  50.21)
     (15,  54.08)
     (30,  55.70)};
  \addlegendentry{GeoGaussian-Training}
   \addplot[color=green,mark=x]
       plot coordinates {
       (2,  23.72)
       (7, 20.62)
       (15,  21.12)
       (30,  21.42)};
   \addlegendentry{GeoGaussian-Evaluation}
   \addplot[mark=*,brown] plot coordinates {
       (2,  38.72)
       (7,  48.64)
       (15, 53.12)
       (30, 56.17)};
  \addlegendentry{3DGS-Training}
   \addplot[color=black,mark=x]
       plot coordinates {
       (2,  22.50)
       (7,  19.31)
       (15, 19.30)
       (30, 19.28)};
   \addlegendentry{3DGS-Evaluation}
   \end{axis}
   \end{tikzpicture}}}
    \caption{Comparison of Rendering performance in training and evaluation datasets.}
    \label{fig:icl-10sparse}
\end{figure}

\begin{table}[]
\centering
\resizebox{.95\linewidth}{!}{%
\begin{tabular}{ll|cccc|cccc|cccc}
\toprule
\multicolumn{2}{c|}{methods} & \multicolumn{4}{c|}{3DGS} & \multicolumn{4}{c|}{LightGS} & \multicolumn{4}{c}{GeoGaussian} \\
\multicolumn{2}{c|}{data} & 10\% & 16.6\% & 50\% & 100\% & 10\% & 16.6\% & 50\% & 100\% & 10\% & 16.6\% & 50\% & 100\% \\ \hline

Room 1 & \begin{tabular}[r]{@{}r@{}}PSNR$\uparrow$\\ SSIM$\uparrow$\\ LPIPS$\downarrow$\end{tabular} 
& \begin{tabular}[c]{@{}c@{}} 40.09 \\ 0.971 \\ 0.022\end{tabular}
& \begin{tabular}[c]{@{}c@{}} 40.39 \\ 0.973 \\ 0.023 \end{tabular}
& \begin{tabular}[c]{@{}c@{}} 40.59 \\ 0.973 \\ 0.024 \end{tabular}
& \begin{tabular}[c]{@{}c@{}} 40.79 \\ 0.973 \\ 0.025 \end{tabular}

& \begin{tabular}[c]{@{}c@{}} \cellcolor[HTML]{c7f9cc}40.25 \\\cellcolor[HTML]{c7f9cc}0.972\\ \cellcolor[HTML]{c7f9cc}0.021 \end{tabular}
& \begin{tabular}[c]{@{}c@{}} \cellcolor[HTML]{c7f9cc}40.53 \\\cellcolor[HTML]{c7f9cc}0.974\\ \cellcolor[HTML]{c7f9cc}0.022 \end{tabular}
& \begin{tabular}[c]{@{}c@{}} \cellcolor[HTML]{c7f9cc}41.03 \\\cellcolor[HTML]{c7f9cc}0.974\\ \cellcolor[HTML]{c7f9cc}0.023 \end{tabular}
& \begin{tabular}[c]{@{}c@{}} \cellcolor[HTML]{c7f9cc}41.26 \\\cellcolor[HTML]{c7f9cc}0.974\\ \cellcolor[HTML]{c7f9cc}0.023 \end{tabular}

& \begin{tabular}[c]{@{}c@{}} \cellcolor[HTML]{57cc99}40.58 \\ \cellcolor[HTML]{57cc99}0.974 \\ \cellcolor[HTML]{57cc99}0.019\end{tabular}
& \begin{tabular}[c]{@{}c@{}} \cellcolor[HTML]{57cc99}41.21 \\ \cellcolor[HTML]{57cc99}0.976 \\ \cellcolor[HTML]{57cc99}0.019\end{tabular}
& \begin{tabular}[c]{@{}c@{}} \cellcolor[HTML]{57cc99}41.51 \\ \cellcolor[HTML]{57cc99}0.977 \\ \cellcolor[HTML]{57cc99}0.019\end{tabular}  
& \begin{tabular}[c]{@{}c@{}} \cellcolor[HTML]{57cc99}41.43 \\ \cellcolor[HTML]{57cc99}0.976 \\ \cellcolor[HTML]{57cc99}0.019\end{tabular} \\

Room 2 & \begin{tabular}[r]{@{}r@{}}PSNR$\uparrow$\\ SSIM$\uparrow$ \\ LPIPS$\downarrow$\end{tabular} 
& \begin{tabular}[c]{@{}c@{}} \cellcolor[HTML]{c7f9cc}23.31 \\ \cellcolor[HTML]{c7f9cc}0.777 \\ \cellcolor[HTML]{c7f9cc}0.251 \end{tabular} 
& \begin{tabular}[c]{@{}c@{}} 30.15 \\ \cellcolor[HTML]{c7f9cc}0.906 \\ \cellcolor[HTML]{c7f9cc}0.080 \end{tabular} 
& \begin{tabular}[c]{@{}c@{}} 37.33 \\ \cellcolor[HTML]{c7f9cc}0.965 \\ 0.023 \end{tabular} 
& \begin{tabular}[c]{@{}c@{}} 39.10 \\ \cellcolor[HTML]{c7f9cc}0.974 \\ \cellcolor[HTML]{c7f9cc}0.017 \end{tabular} 

& \begin{tabular}[c]{@{}c@{}} 23.28 \\0.775\\ 0.252 \end{tabular}
& \begin{tabular}[c]{@{}c@{}} \cellcolor[HTML]{c7f9cc}30.18 \\\cellcolor[HTML]{c7f9cc}0.906\\ \cellcolor[HTML]{c7f9cc}0.080 \end{tabular}
& \begin{tabular}[c]{@{}c@{}} \cellcolor[HTML]{c7f9cc}37.51 \\\cellcolor[HTML]{c7f9cc}0.965\\\cellcolor[HTML]{c7f9cc} 0.022 \end{tabular}
& \begin{tabular}[c]{@{}c@{}} \cellcolor[HTML]{c7f9cc}39.23 \\\cellcolor[HTML]{c7f9cc}0.974\\ \cellcolor[HTML]{c7f9cc}0.017 \end{tabular}

& \begin{tabular}[c]{@{}c@{}} \cellcolor[HTML]{57cc99}25.23 \\ \cellcolor[HTML]{57cc99}0.841 \\ \cellcolor[HTML]{57cc99}0.142 \end{tabular} 
& \begin{tabular}[c]{@{}c@{}} \cellcolor[HTML]{57cc99}31.26 \\ \cellcolor[HTML]{57cc99}0.925 \\ \cellcolor[HTML]{57cc99}0.059 \end{tabular} 
& \begin{tabular}[c]{@{}c@{}} \cellcolor[HTML]{57cc99}37.70 \\ \cellcolor[HTML]{57cc99}0.968 \\ \cellcolor[HTML]{c7f9cc}0.023 \end{tabular} 
& \begin{tabular}[c]{@{}c@{}} \cellcolor[HTML]{57cc99}39.46 \\ \cellcolor[HTML]{57cc99}0.975 \\ \cellcolor[HTML]{c7f9cc}0.018 \end{tabular}  \\

Office 2 & \begin{tabular}[r]{@{}r@{}}PSNR$\uparrow$\\ SSIM$\uparrow$\\ LPIPS$\downarrow$\end{tabular} 
& \begin{tabular}[c]{@{}c@{}}\cellcolor[HTML]{c7f9cc}26.24 \\ \cellcolor[HTML]{c7f9cc}0.844 \\ \cellcolor[HTML]{c7f9cc}0.145 \end{tabular} 
& \begin{tabular}[c]{@{}c@{}}\cellcolor[HTML]{c7f9cc}29.82 \\ \cellcolor[HTML]{c7f9cc}0.896 \\ \cellcolor[HTML]{c7f9cc}0.078\end{tabular} 
& \begin{tabular}[c]{@{}c@{}}35.54 \\ \cellcolor[HTML]{c7f9cc}0.943 \\ 0.036 \end{tabular} 
& \begin{tabular}[c]{@{}c@{}}37.88 \\ \cellcolor[HTML]{c7f9cc}0.962 \\ 0.024 \end{tabular} 

& \begin{tabular}[c]{@{}c@{}} 26.22 \\0.843\\ 0.146 \end{tabular}
& \begin{tabular}[c]{@{}c@{}} 29.77 \\\cellcolor[HTML]{c7f9cc}0.896\\ \cellcolor[HTML]{c7f9cc}0.078 \end{tabular}
& \begin{tabular}[c]{@{}c@{}} \cellcolor[HTML]{c7f9cc}35.58 \\\cellcolor[HTML]{c7f9cc}0.943\\ \cellcolor[HTML]{c7f9cc}0.035 \end{tabular}
& \begin{tabular}[c]{@{}c@{}} \cellcolor[HTML]{c7f9cc}37.99 \\\cellcolor[HTML]{c7f9cc}0.962\\ \cellcolor[HTML]{c7f9cc}0.023 \end{tabular}

& \begin{tabular}[c]{@{}c@{}}\cellcolor[HTML]{57cc99}28.35\\ \cellcolor[HTML]{57cc99}0.874\\  \cellcolor[HTML]{57cc99}0.100\end{tabular}
& \begin{tabular}[c]{@{}c@{}}\cellcolor[HTML]{57cc99}32.34\\ \cellcolor[HTML]{57cc99}0.917\\  \cellcolor[HTML]{57cc99}0.055\end{tabular} 
& \begin{tabular}[c]{@{}c@{}}\cellcolor[HTML]{57cc99}36.76\\ \cellcolor[HTML]{57cc99}0.952\\  \cellcolor[HTML]{57cc99}0.026\end{tabular} 
& \begin{tabular}[c]{@{}c@{}}\cellcolor[HTML]{57cc99}38.54\\ \cellcolor[HTML]{57cc99}0.967\\ \cellcolor[HTML]{57cc99}0.017\end{tabular} \\

Office 3 & \begin{tabular}[r]{@{}r@{}}PSNR$\uparrow$\\ SSIM$\uparrow$\\ LPIPS$\downarrow$\end{tabular} 
& \begin{tabular}[c]{@{}c@{}}\cellcolor[HTML]{c7f9cc}19.28\\ \cellcolor[HTML]{c7f9cc}0.718\\  \cellcolor[HTML]{c7f9cc}0.219\end{tabular} 
& \begin{tabular}[c]{@{}c@{}}\cellcolor[HTML]{c7f9cc}22.86\\ \cellcolor[HTML]{c7f9cc}0.848 \\ \cellcolor[HTML]{c7f9cc}0.143 \end{tabular} 
& \begin{tabular}[c]{@{}c@{}}32.20\\ \cellcolor[HTML]{c7f9cc}0.950\\  \cellcolor[HTML]{c7f9cc}0.044 \end{tabular} 
& \begin{tabular}[c]{@{}c@{}} 36.04 \\ \cellcolor[HTML]{c7f9cc}0.975 \\ 0.017 \end{tabular} 

& \begin{tabular}[c]{@{}c@{}} 19.19 \\0.716\\ \cellcolor[HTML]{c7f9cc}0.219 \end{tabular}
& \begin{tabular}[c]{@{}c@{}} 22.82 \\0.847\\ \cellcolor[HTML]{c7f9cc}0.143 \end{tabular}
& \begin{tabular}[c]{@{}c@{}} \cellcolor[HTML]{c7f9cc}32.21 \\0.949\\ \cellcolor[HTML]{c7f9cc}0.044 \end{tabular}
& \begin{tabular}[c]{@{}c@{}} \cellcolor[HTML]{c7f9cc}36.06 \\\cellcolor[HTML]{c7f9cc}0.975\\ \cellcolor[HTML]{c7f9cc}0.016 \end{tabular}

& \begin{tabular}[c]{@{}c@{}} \cellcolor[HTML]{57cc99}21.42 \\ \cellcolor[HTML]{57cc99}0.769 \\ \cellcolor[HTML]{57cc99}0.145\end{tabular}
& \begin{tabular}[c]{@{}c@{}} \cellcolor[HTML]{57cc99}27.06 \\ \cellcolor[HTML]{57cc99}0.907 \\ \cellcolor[HTML]{57cc99}0.082\end{tabular}
& \begin{tabular}[c]{@{}c@{}} \cellcolor[HTML]{57cc99}33.52 \\ \cellcolor[HTML]{57cc99}0.966 \\ \cellcolor[HTML]{57cc99}0.024\end{tabular}
& \begin{tabular}[c]{@{}c@{}} \cellcolor[HTML]{57cc99}36.19 \\ \cellcolor[HTML]{57cc99}0.977 \\ \cellcolor[HTML]{57cc99}0.015\end{tabular}  \\
 \hline

Avg. & \begin{tabular}[r]{@{}r@{}}PSNR$\uparrow$\\ SSIM$\uparrow$\\ LPIPS$\downarrow$\end{tabular} 
& \begin{tabular}[c]{@{}c@{}} \cellcolor[HTML]{c7f9cc}27.23 \\\cellcolor[HTML]{c7f9cc}0.827\\ \cellcolor[HTML]{c7f9cc}0.159 \end{tabular}
& \begin{tabular}[c]{@{}c@{}} 30.80 \\\cellcolor[HTML]{c7f9cc}0.905\\ \cellcolor[HTML]{c7f9cc}0.081 \end{tabular}
& \begin{tabular}[c]{@{}c@{}} 36.41 \\\cellcolor[HTML]{c7f9cc}0.967\\ \cellcolor[HTML]{c7f9cc}0.031 \end{tabular}
& \begin{tabular}[c]{@{}c@{}} 38.45\\\cellcolor[HTML]{c7f9cc}0.971\\  0.020\end{tabular}

& \begin{tabular}[c]{@{}c@{}} \cellcolor[HTML]{c7f9cc}27.23 \\0.826\\ \cellcolor[HTML]{c7f9cc}0.159 \end{tabular}
& \begin{tabular}[c]{@{}c@{}} \cellcolor[HTML]{c7f9cc}30.82 \\\cellcolor[HTML]{c7f9cc}0.905\\ 0.087 \end{tabular}
& \begin{tabular}[c]{@{}c@{}} \cellcolor[HTML]{c7f9cc}36.58 \\0.957\\ \cellcolor[HTML]{c7f9cc}0.031 \end{tabular}
& \begin{tabular}[c]{@{}c@{}} \cellcolor[HTML]{c7f9cc}38.63\\\cellcolor[HTML]{c7f9cc}0.971\\ \cellcolor[HTML]{c7f9cc}0.019 \end{tabular}

& \begin{tabular}[c]{@{}c@{}} \cellcolor[HTML]{57cc99}28.89 \\\cellcolor[HTML]{57cc99}0.864\\ \cellcolor[HTML]{57cc99}0.101 \end{tabular}
& \begin{tabular}[c]{@{}c@{}} \cellcolor[HTML]{57cc99}32.96 \\\cellcolor[HTML]{57cc99}0.931\\ \cellcolor[HTML]{57cc99}0.053 \end{tabular}
& \begin{tabular}[c]{@{}c@{}} \cellcolor[HTML]{57cc99}37.37 \\\cellcolor[HTML]{57cc99}0.965\\ \cellcolor[HTML]{57cc99}0.024 \end{tabular}
& \begin{tabular}[c]{@{}c@{}} \cellcolor[HTML]{57cc99}38.90\\ \cellcolor[HTML]{57cc99}0.973\\ \cellcolor[HTML]{57cc99}0.017\end{tabular}   \\

\bottomrule
\end{tabular}}
\caption{Comparison of rendering on the ICL-NUIM dataset.}
\label{tab:sparse-icl-appen}
\end{table}

As we mentioned in the Experiment section (see Section~\textcolor{red}{4.2}), the four sequences are used to evaluate these Gaussian Splatting approaches. As listed in Table~\ref{tab:sparse-icl-appen}, the proposed method shows robust rendering performance in different training settings compared with state-of-the-art Gaussian Splatting methods. When the number of training frames is reduced, our method shows more robust performance since our models have better geometry to alleviate overfitting problems in rendering tasks. For example, our model trained on ICL-O3 (10\%) achieves a PSNR of $21.42$, while models of 3DGS and LightGS obtain $19.28$ and $19.19$, respectively.

\begin{figure} 
	\centering
\subfloat[3DGS]{
 \begin{minipage}[b]{0.22\textwidth}
 \begin{tikzpicture}
     \node (img1) at (0,1.6*7) {\includegraphics[width=\linewidth]{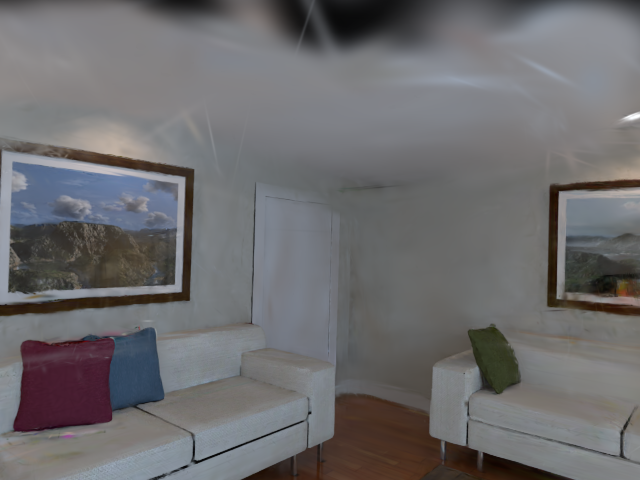}};
     \node (img1) at (0,1.6*6) {\includegraphics[width=\linewidth]{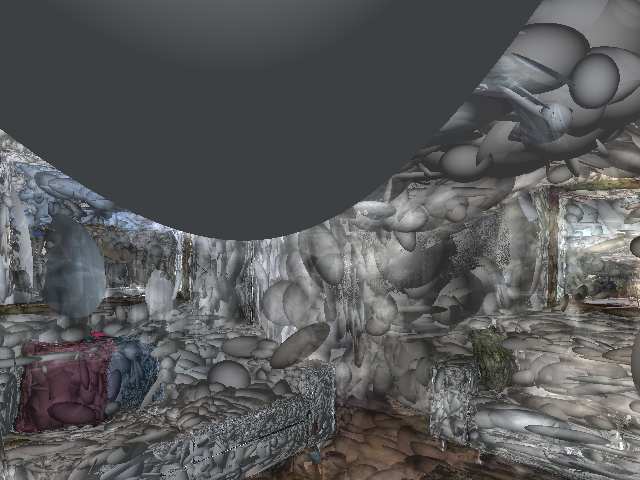}}; 

    \draw[red, thick] (-1.2,1.643*7) rectangle +(1.6,0.7);
    \draw[red, thick] (-1.2,1.643*6) rectangle +(1.6,0.7);

     
 \end{tikzpicture}
\end{minipage}}
 \subfloat[LightGS]{
 \begin{minipage}[b]{0.22\textwidth}
 \begin{tikzpicture}

     \node (img1) at (0,1.6*7) {\includegraphics[width=\linewidth]{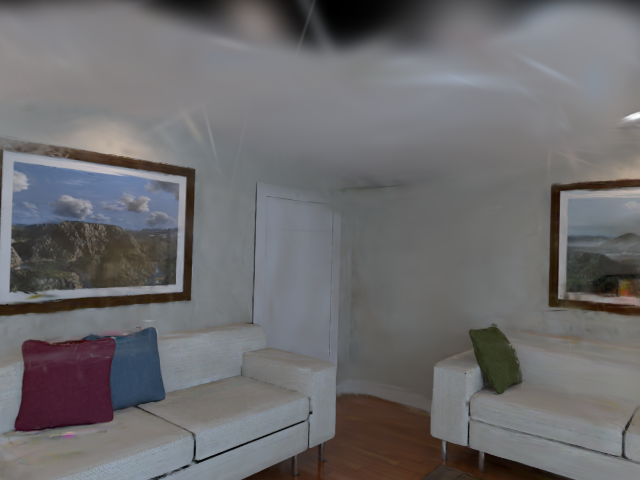}};
     \node (img1) at (0,1.6*6) {\includegraphics[width=\linewidth]{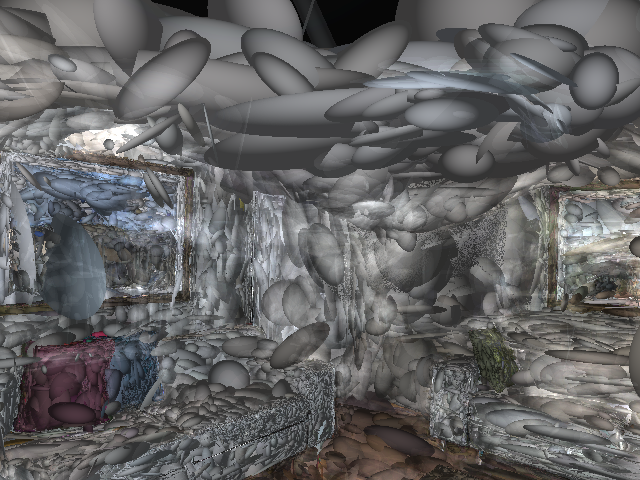}}; 

    \draw[red, thick] (-1.2,1.643*7) rectangle +(1.6,0.7);
    \draw[red, thick] (-1.2,1.643*6) rectangle +(1.6,0.7);

    
 \end{tikzpicture}
\end{minipage}}
 \subfloat[GeoGaussian]{
 \begin{minipage}[b]{0.22\textwidth}
 \begin{tikzpicture}
    \node(img7) at (0,1.6*7){\includegraphics[width=\linewidth]{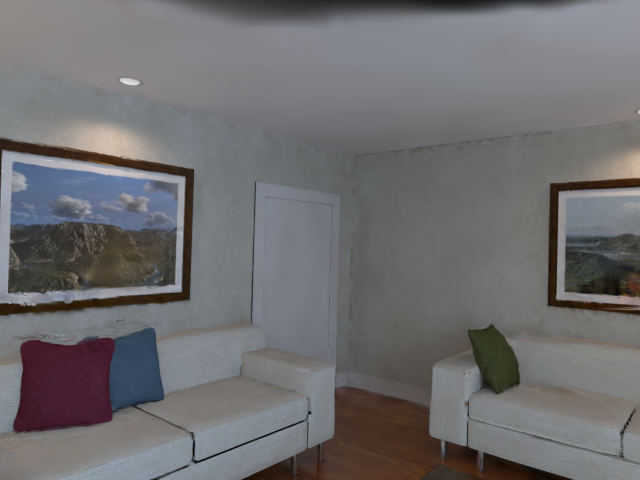}};
    \node(img8) at (0,1.6*6){\includegraphics[width=\linewidth]{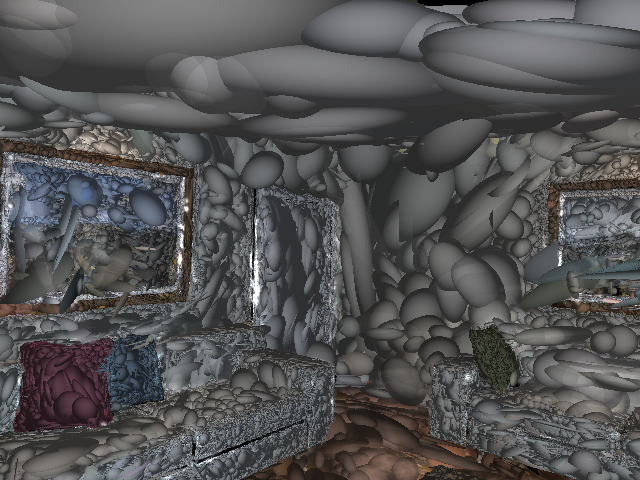}};
    
    \draw[red, thick] (-1.2,1.643*7) rectangle +(1.6,0.7);
    \draw[red, thick] (-1.2,1.643*6) rectangle +(1.6,0.7);

    
 \end{tikzpicture}
	\end{minipage}}
\subfloat[Reference]{
	\begin{minipage}[b]{0.22\textwidth}
 \begin{tikzpicture}

     \node (img1) at (0,1.6*7){\includegraphics[width=\linewidth]{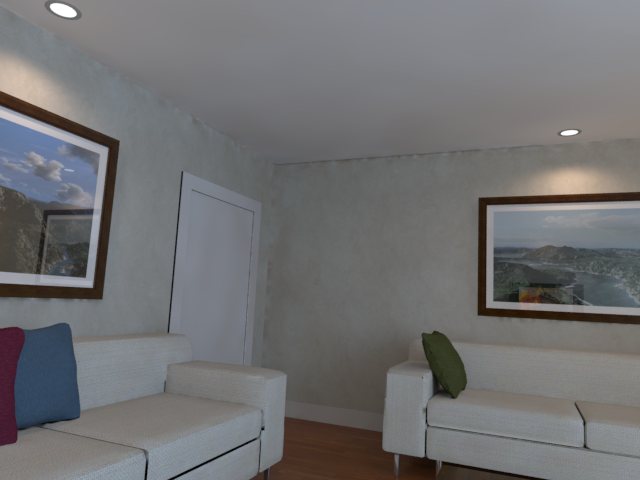}};
     \node (img1) at (0,1.6*6){\includegraphics[width=\linewidth]{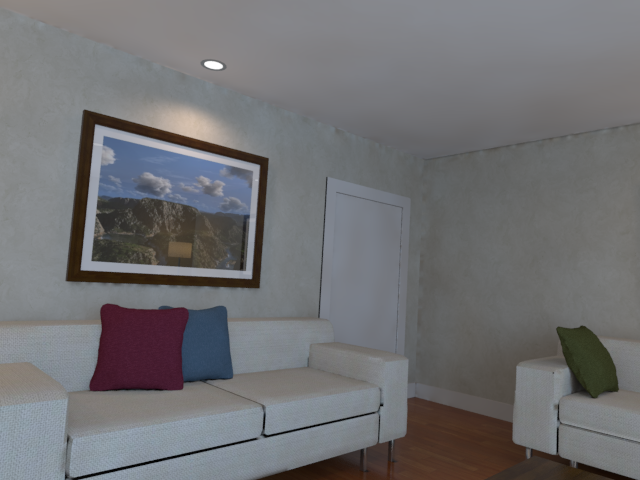}};

    \draw[red, thick] (-1.2,1.643*7) rectangle +(1.2,0.7);
    \draw[red, thick] (-0.8,1.643*6) rectangle +(1.2,0.7);

     \end{tikzpicture}
	\end{minipage}}
\caption{Comparisons of novel view rendering on the ICL-NUIM datasets. In these scenarios, 3DGS and LightGS struggle with photorealistic rendering. As depicted in (d), the two training views surrounding the novel view.}
\label{fig:icl-qualitative}
\end{figure}
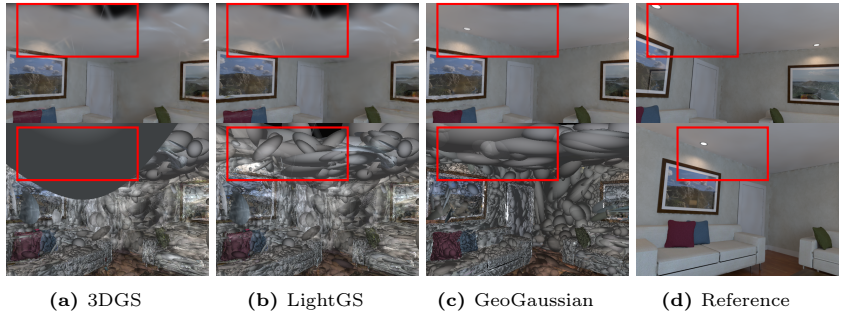

As we mentioned in the Experiment section (see Section~\textcolor{red}{4.2}), the four sequences are used to evaluate these Gaussian Splatting approaches. As listed in Table~\ref{tab:sparse-icl-appen}, the proposed method shows robust rendering performance in different training settings compared with state-of-the-art Gaussian Splatting methods. When the number of training frames is reduced, our method shows more robust performance since our models have better geometry to alleviate overfitting problems in rendering tasks. For example, our model trained on ICL-O3 (10\%) achieves a PSNR of $21.42$, while models of 3DGS and LightGS obtain $19.28$ and $19.19$, respectively.

In an extreme case, when we observe the training process of these methods on ICL-O3 (10\%), as shown in Figure~\ref{fig:icl-10sparse}, it is evident that Gaussian Splatting methods tend to overfit the training views due to the limited number of training views. However, our method maintains a certain level of generalization ability based on geometry constraints. For example, as the model improves its rendering results on the training dataset, the performance on the test dataset does not decrease significantly, as shown in Figure~\ref{subfig:short-10-a}.





%

\subsection{TUM RGB-D Sequences}

\begin{table}[]
\centering
\resizebox{.95\linewidth}{!}{%
\begin{tabular}{ll|cccc|cccc|cccc}
\toprule
\multicolumn{2}{c|}{methods} & \multicolumn{4}{c|}{3DGS} & \multicolumn{4}{c|}{LightGS} & \multicolumn{4}{c}{GeoGaussian} \\
\multicolumn{2}{c|}{data} & 10\% & 16.6\% & 50\% & 100\% & 10\% & 16.6\% & 50\% & 100\% & 10\% & 16.6\% & 50\% & 100\% \\ \hline

f3/lag-cabinet & \begin{tabular}[r]{@{}r@{}}PSNR$\uparrow$\\ SSIM$\uparrow$\\ LPIPS$\downarrow$\end{tabular} 
& \begin{tabular}[c]{@{}c@{}} 20.05 \\ 0.772 \\ 0.226 \end{tabular}
& \begin{tabular}[c]{@{}c@{}} \cellcolor[HTML]{c7f9cc}23.27 \\ 0.831 \\ \cellcolor[HTML]{c7f9cc}0.143 \end{tabular}
& \begin{tabular}[c]{@{}c@{}} \cellcolor[HTML]{c7f9cc}24.63 \\ \cellcolor[HTML]{c7f9cc}0.864 \\ \cellcolor[HTML]{57cc99}0.123 \end{tabular}
& \begin{tabular}[c]{@{}c@{}} \cellcolor[HTML]{c7f9cc}24.87 \\ 0.866 \\ 0.127 \end{tabular}

& \begin{tabular}[c]{@{}c@{}} \cellcolor[HTML]{c7f9cc}20.09 \\ \cellcolor[HTML]{c7f9cc}0.773 \\ \cellcolor[HTML]{c7f9cc}0.225 \end{tabular}
& \begin{tabular}[c]{@{}c@{}} \cellcolor[HTML]{57cc99}23.32 \\ \cellcolor[HTML]{c7f9cc}0.832 \\ \cellcolor[HTML]{57cc99}0.142 \end{tabular}
& \begin{tabular}[c]{@{}c@{}} \cellcolor[HTML]{57cc99}24.74 \\ \cellcolor[HTML]{57cc99}0.865 \\ \cellcolor[HTML]{57cc99}0.123 \end{tabular}
& \begin{tabular}[c]{@{}c@{}} \cellcolor[HTML]{57cc99}24.89 \\ \cellcolor[HTML]{c7f9cc}0.867 \\ \cellcolor[HTML]{c7f9cc}0.126 \end{tabular}

& \begin{tabular}[c]{@{}c@{}} \cellcolor[HTML]{57cc99}20.85 \\ \cellcolor[HTML]{57cc99}0.804 \\ \cellcolor[HTML]{57cc99}0.178 \end{tabular}
& \begin{tabular}[c]{@{}c@{}} 22.22 \\ \cellcolor[HTML]{57cc99}0.833 \\ 0.158 \end{tabular}
& \begin{tabular}[c]{@{}c@{}} 24.15 \\ 0.863 \\ \cellcolor[HTML]{c7f9cc}0.130 \end{tabular}  
& \begin{tabular}[c]{@{}c@{}} 24.79 \\ \cellcolor[HTML]{57cc99}0.868 \\ \cellcolor[HTML]{57cc99}0.125 \end{tabular} \\

f3/long-office & \begin{tabular}[r]{@{}r@{}}PSNR$\uparrow$\\ SSIM$\uparrow$ \\ LPIPS$\downarrow$\end{tabular} 
& \begin{tabular}[c]{@{}c@{}} 16.67 \\ 0.598 \\ 0.336 \end{tabular} 
& \begin{tabular}[c]{@{}c@{}} 18.88 \\ 0.686 \\ 0.246 \end{tabular} 
& \begin{tabular}[c]{@{}c@{}} 23.54 \\ 0.812 \\ 0.148 \end{tabular} 
& \begin{tabular}[c]{@{}c@{}} 24.33 \\ 0.832 \\ 0.139 \end{tabular} 

& \begin{tabular}[c]{@{}c@{}} \cellcolor[HTML]{c7f9cc}16.69 \\ \cellcolor[HTML]{c7f9cc}0.600 \\ \cellcolor[HTML]{c7f9cc}0.335 \end{tabular}
& \begin{tabular}[c]{@{}c@{}} \cellcolor[HTML]{c7f9cc}18.93 \\ \cellcolor[HTML]{c7f9cc}0.689 \\ \cellcolor[HTML]{c7f9cc}0.245 \end{tabular}
& \begin{tabular}[c]{@{}c@{}} \cellcolor[HTML]{c7f9cc}23.66 \\ \cellcolor[HTML]{c7f9cc}0.813 \\ \cellcolor[HTML]{c7f9cc}0.146 \end{tabular}
& \begin{tabular}[c]{@{}c@{}} \cellcolor[HTML]{c7f9cc}24.43 \\ \cellcolor[HTML]{c7f9cc}0.835 \\ \cellcolor[HTML]{c7f9cc}0.138 \end{tabular}

& \begin{tabular}[c]{@{}c@{}} \cellcolor[HTML]{57cc99}18.89 \\ \cellcolor[HTML]{57cc99}0.643 \\ \cellcolor[HTML]{57cc99}0.279 \end{tabular} 
& \begin{tabular}[c]{@{}c@{}} \cellcolor[HTML]{57cc99}20.55 \\ \cellcolor[HTML]{57cc99}0.736 \\ \cellcolor[HTML]{57cc99}0.196 \end{tabular} 
& \begin{tabular}[c]{@{}c@{}} \cellcolor[HTML]{57cc99}24.56 \\ \cellcolor[HTML]{57cc99}0.833 \\ \cellcolor[HTML]{57cc99}0.128 \end{tabular} 
& \begin{tabular}[c]{@{}c@{}} \cellcolor[HTML]{57cc99}25.12 \\ \cellcolor[HTML]{57cc99}0.843 \\ \cellcolor[HTML]{57cc99}0.125 \end{tabular}  \\
\hline

Avg. & \begin{tabular}[r]{@{}r@{}}PSNR$\uparrow$\\ SSIM$\uparrow$\\ LPIPS$\downarrow$\end{tabular} 
& \begin{tabular}[c]{@{}c@{}} 18.31 \\0.685\\ 0.281 \end{tabular}
& \begin{tabular}[c]{@{}c@{}} 21.07 \\ \cellcolor[HTML]{c7f9cc}0.758\\ 0.194 \end{tabular}
& \begin{tabular}[c]{@{}c@{}} 24.08 \\0.838\\ 0.135 \end{tabular}
& \begin{tabular}[c]{@{}c@{}} 24.60 \\0.849\\ 0.133 \end{tabular}

& \begin{tabular}[c]{@{}c@{}} \cellcolor[HTML]{c7f9cc}18.39 \\\cellcolor[HTML]{c7f9cc}0.686\\ \cellcolor[HTML]{c7f9cc}0.280 \end{tabular}
& \begin{tabular}[c]{@{}c@{}} \cellcolor[HTML]{c7f9cc}21.12 \\ \cellcolor[HTML]{57cc99}0.760\\\cellcolor[HTML]{c7f9cc}0.193 \end{tabular}
& \begin{tabular}[c]{@{}c@{}} \cellcolor[HTML]{c7f9cc}24.20 \\\cellcolor[HTML]{c7f9cc}0.839\\ \cellcolor[HTML]{c7f9cc}0.134 \end{tabular}
& \begin{tabular}[c]{@{}c@{}} \cellcolor[HTML]{c7f9cc}24.66 \\ \cellcolor[HTML]{57cc99}0.851\\ \cellcolor[HTML]{c7f9cc}0.132 \end{tabular}

& \begin{tabular}[c]{@{}c@{}} \cellcolor[HTML]{57cc99}19.87 \\ \cellcolor[HTML]{57cc99}0.723\\ \cellcolor[HTML]{57cc99}0.228 \end{tabular}
& \begin{tabular}[c]{@{}c@{}} \cellcolor[HTML]{57cc99}21.38 \\0.748\\ \cellcolor[HTML]{57cc99}0.177 \end{tabular}
& \begin{tabular}[c]{@{}c@{}} \cellcolor[HTML]{57cc99}24.35 \\ \cellcolor[HTML]{57cc99}0.848\\ \cellcolor[HTML]{57cc99}0.129 \end{tabular}
& \begin{tabular}[c]{@{}c@{}} \cellcolor[HTML]{57cc99}24.95 \\\cellcolor[HTML]{c7f9cc}0.850\\ \cellcolor[HTML]{57cc99}0.125\end{tabular}   \\

\bottomrule
\end{tabular}}
\caption{Comparison of rendering on the non-structured sequences on the TUM RGB-D dataset.}
\label{tab:sparse-TUM-appen}
\end{table}

\begin{figure}
    \centering
    \begin{subfigure}{0.3\linewidth}
    \includegraphics[width=\linewidth]{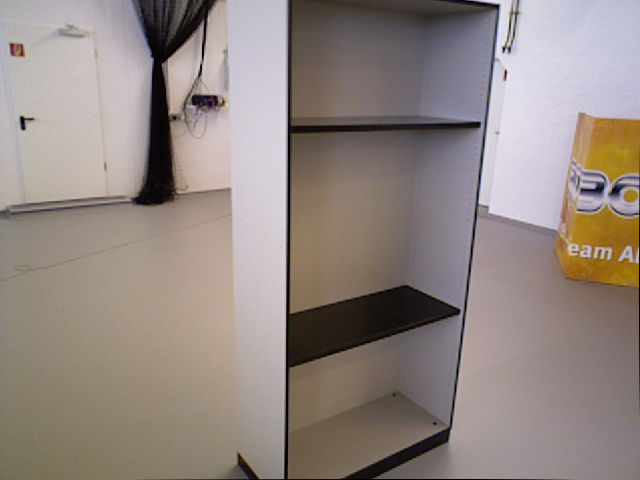}
    \caption{Real scene}
    \label{fig:short-lag-a}
  \end{subfigure}
  \begin{subfigure}{0.3\linewidth}
    \includegraphics[width=\linewidth]{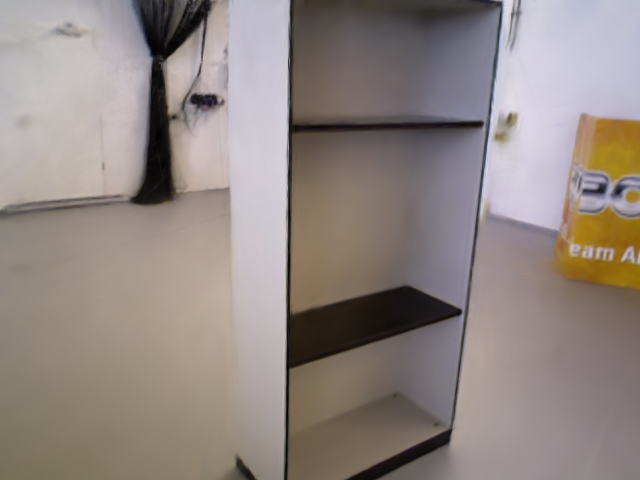}
    \caption{3DGS}
    \label{fig:short-lag-b}
  \end{subfigure}
  \begin{subfigure}{0.3\linewidth}
    \includegraphics[width=\linewidth]{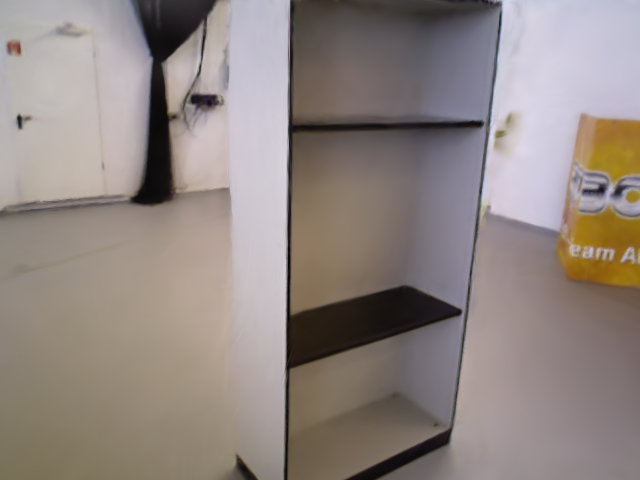}
    \caption{GeoGaussian(ours)}
    \label{fig:short-lag-c}
  \end{subfigure}
    \caption{Scene of the lag-cabinet sequence and render quality.}
    \label{fig:lag}
\end{figure}
Structured environments from the TUM RGB-D dataset are utilized to evaluate rendering performance in Section~\textcolor{red}{4.4}. For a comprehensive evaluation of our method in general scenarios, we test it on the f3/lag-cabinet and f3/long-office sequences, which are non-structured Scenes, as shown in Figure~\ref{fig:lag} and Table~\ref{tab:sparse-TUM-appen}.

As mentioned in Section~\textcolor{red}{4.4}, the proposed method is not as effective as 3DGS and LightGS in the lag-cabinet sequence. This is because the far walls in this sequence cannot be accurately captured based on point clouds, making it challenging to obtain good normal vectors from the point clouds.

\section{Model Reconstruction}

In this section, we present more quantitative results in the comparison of 3D Gaussian models. Based on the ground truth mesh models provided by the Replica dataset, we align these mesh models with point clouds from Gaussian models, where we randomly sample three points in each Gaussian ellipsoid.

\begin{table}[]
\centering
\resizebox{.95\linewidth}{!}{%
\begin{tabular}{ll|cccccccc|c}  
\toprule
\multicolumn{2}{c|}{methods} & R0 & R1 & R2 & OFF0& OFF1 & OFF2 & OFF3 & OFF4 & Avg. \\ \hline

3DGS & 
\begin{tabular}[c]{@{}c@{}} mean \\ std \end{tabular} 
& \begin{tabular}[c]{@{}c@{}} 0.026 \\ 0.066 \end{tabular} 
& \begin{tabular}[c]{@{}c@{}} 0.025 \\ 0.081 \end{tabular} 
& \begin{tabular}[c]{@{}c@{}} 0.042 \\ 0.146 \end{tabular} 
& \begin{tabular}[c]{@{}c@{}} \cellcolor[HTML]{57cc99}0.017 \\ 0.050 \end{tabular} 
& \begin{tabular}[c]{@{}c@{}} \cellcolor[HTML]{57cc99}0.019 \\ \cellcolor[HTML]{57cc99}0.055 \end{tabular}  
& \begin{tabular}[c]{@{}c@{}} 0.039 \\ 0.201 \end{tabular} 
& \begin{tabular}[c]{@{}c@{}} 0.032 \\ 0.066 \end{tabular} 
& \begin{tabular}[c]{@{}c@{}} 0.032 \\ 0.112 \end{tabular} 
& \begin{tabular}[c]{@{}c@{}} 0.029 \\ 0.097 \end{tabular} 
\\

\begin{tabular}[c]{@{}l@{}} GeoGaussian \\ (ours) \end{tabular} 
 & 
\begin{tabular}[c]{@{}c@{}} mean \\ std \end{tabular} 
& \begin{tabular}[c]{@{}c@{}} \cellcolor[HTML]{57cc99}0.018 \\ \cellcolor[HTML]{57cc99}0.032 \end{tabular} 
& \begin{tabular}[c]{@{}c@{}} \cellcolor[HTML]{57cc99}0.014 \\ \cellcolor[HTML]{57cc99}0.016 \end{tabular} 
& \begin{tabular}[c]{@{}c@{}} \cellcolor[HTML]{57cc99}0.015 \\ \cellcolor[HTML]{57cc99}0.028 \end{tabular} 
& \begin{tabular}[c]{@{}c@{}} 0.020 \\ \cellcolor[HTML]{57cc99}0.042 \end{tabular} 
& \begin{tabular}[c]{@{}c@{}} 0.029 \\ 0.067 \end{tabular}  
& \begin{tabular}[c]{@{}c@{}} \cellcolor[HTML]{57cc99}0.013 \\ \cellcolor[HTML]{57cc99}0.024 \end{tabular} 
& \begin{tabular}[c]{@{}c@{}} \cellcolor[HTML]{57cc99}0.018 \\ \cellcolor[HTML]{57cc99}0.020 \end{tabular} 
& \begin{tabular}[c]{@{}c@{}} \cellcolor[HTML]{57cc99}0.014 \\ \cellcolor[HTML]{57cc99}0.023 \end{tabular} 
& \begin{tabular}[c]{@{}c@{}} \cellcolor[HTML]{57cc99}0.018 \\ \cellcolor[HTML]{57cc99}0.031 \end{tabular} 
\\

\bottomrule
\end{tabular}}
\caption{Comparison of reconstruction performance on the Replica dataset.}
\label{tab:sparse-replica-recon-appen}
\end{table}

\begin{figure}
  \centering
  \subfloat[R0]
  {\includegraphics[width=0.3\linewidth]{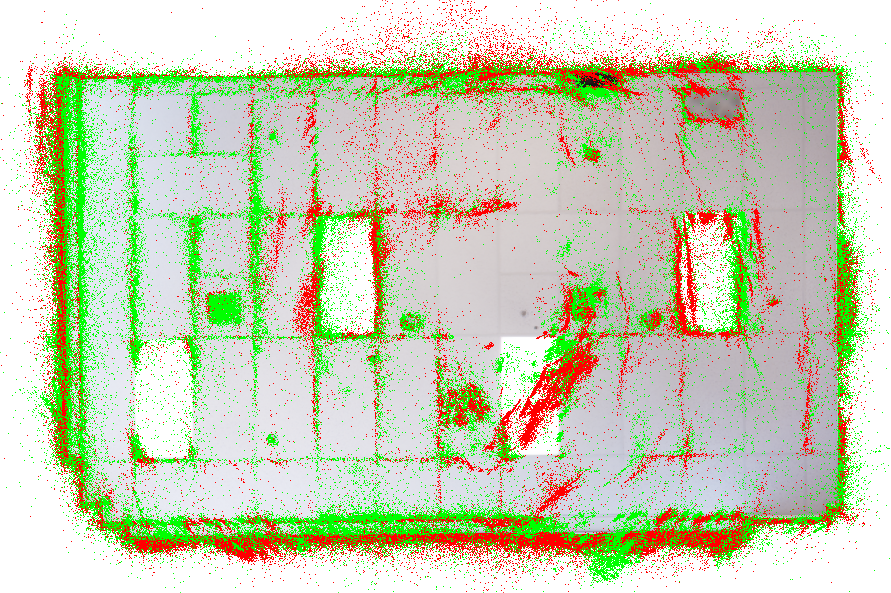}
  }
  \subfloat[R1]
   {\includegraphics[width=0.3\linewidth]{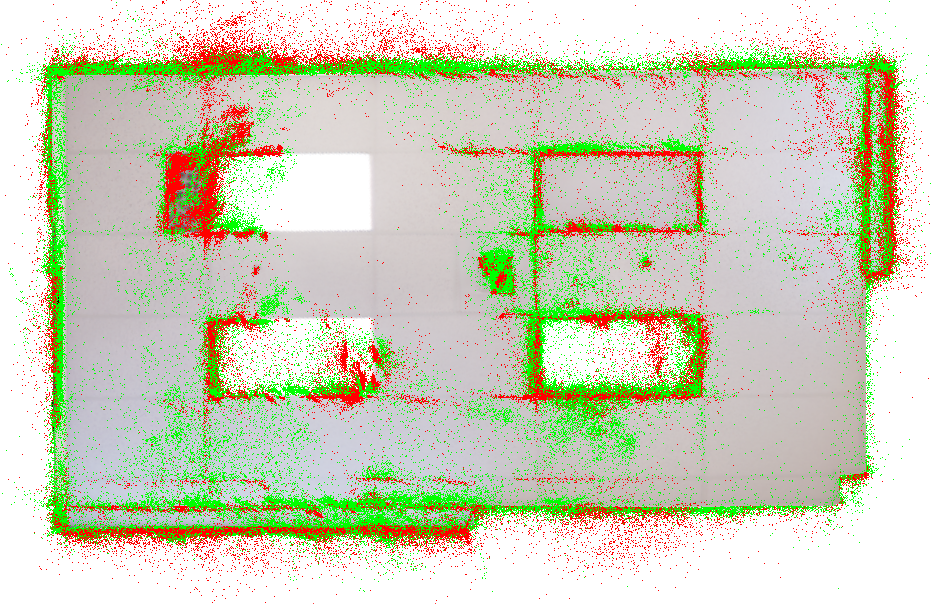}
  }
  \subfloat[R2]
  {\includegraphics[width=0.3\linewidth]{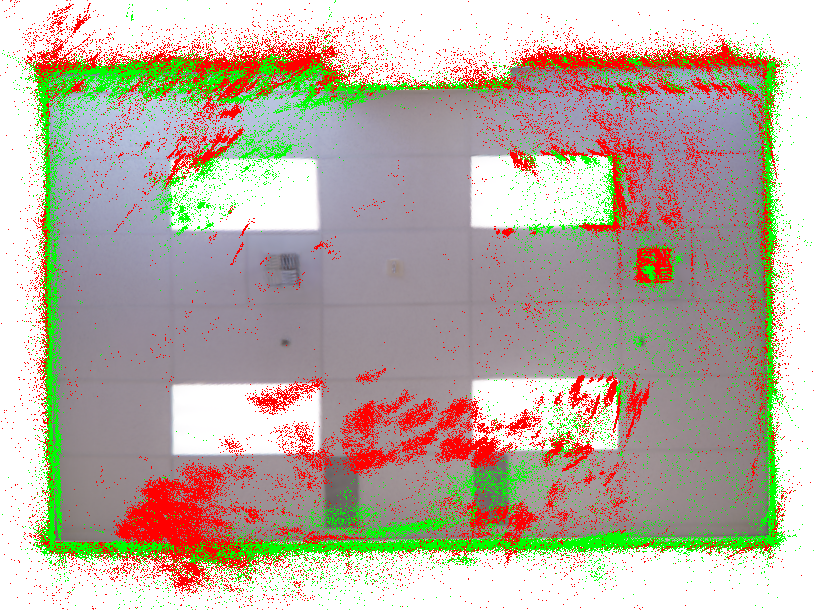}
  }
  
  \subfloat[OFF0]
  {\includegraphics[width=0.3\linewidth]{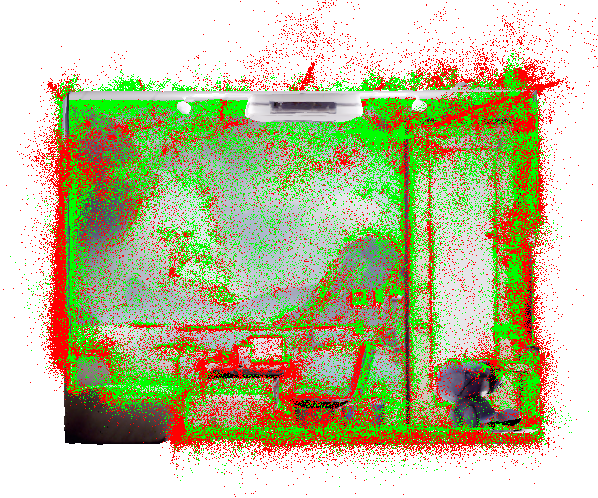}
  }
  \subfloat[OFF1]
  {\includegraphics[width=0.3\linewidth]{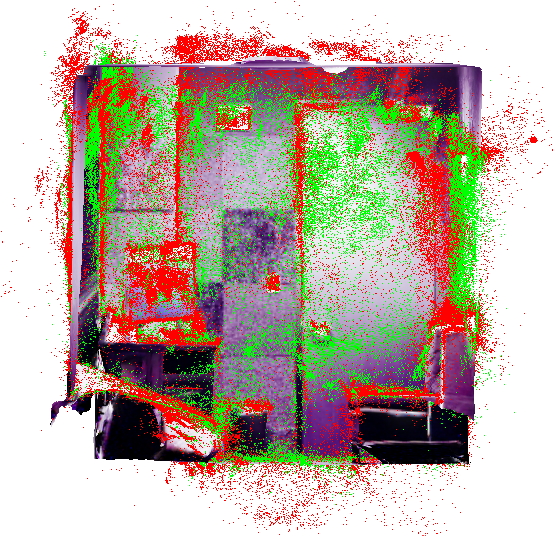}
  }
  \subfloat[OFF2]
   {\includegraphics[width=0.3\linewidth]{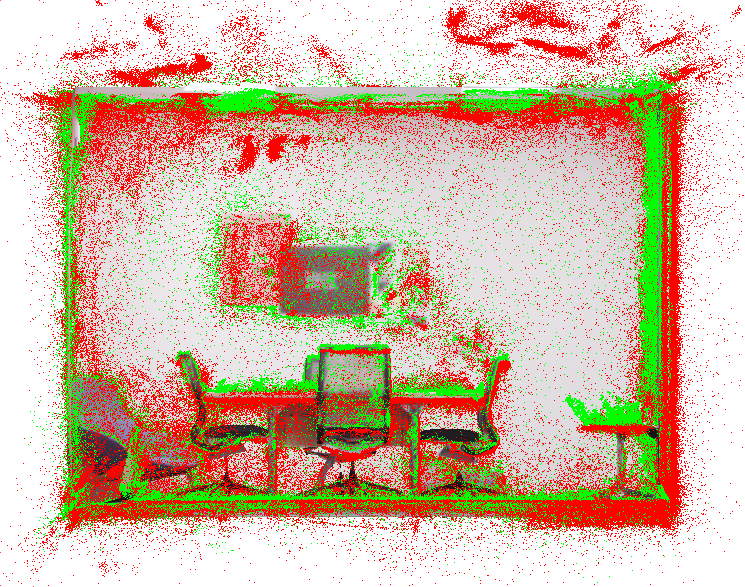}
  }
  
  \subfloat[OFF3]
  {\includegraphics[width=0.3\linewidth]{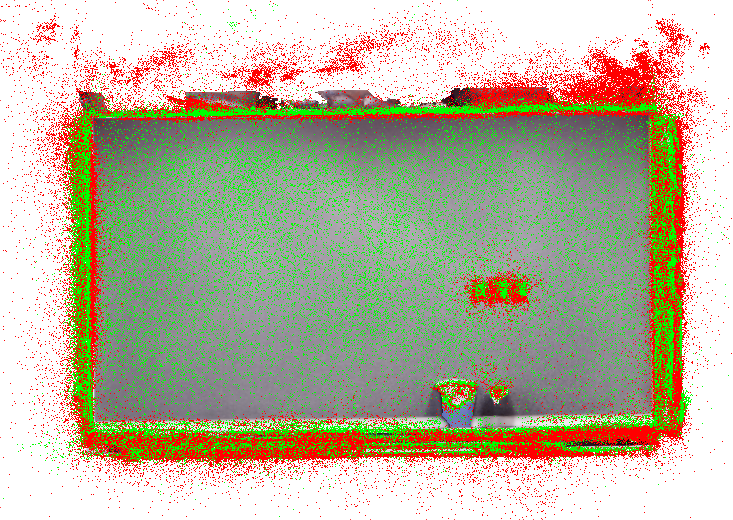}
  }
  \subfloat[OFF4]
  {\includegraphics[width=0.3\linewidth]{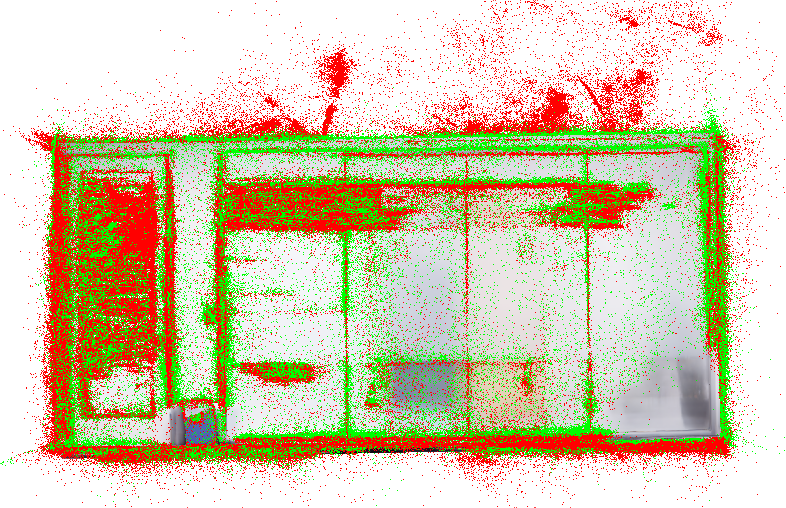}
  }
  \subfloat[Description]
  {\resizebox{0.3\linewidth}{!}{
\begin{tikzpicture}[spy using outlines={red,magnification=2,size=1cm}, connect spies]
    \node[draw,align=left] at (1.8,1.5) {
    Ground truth mesh \\
    Red points from 3DGS\\
    Green points from GeoGaussian};
\end{tikzpicture}}
  }
  \caption{Reconstruction error visualization.}
\label{fig:reconstruction}
\end{figure}
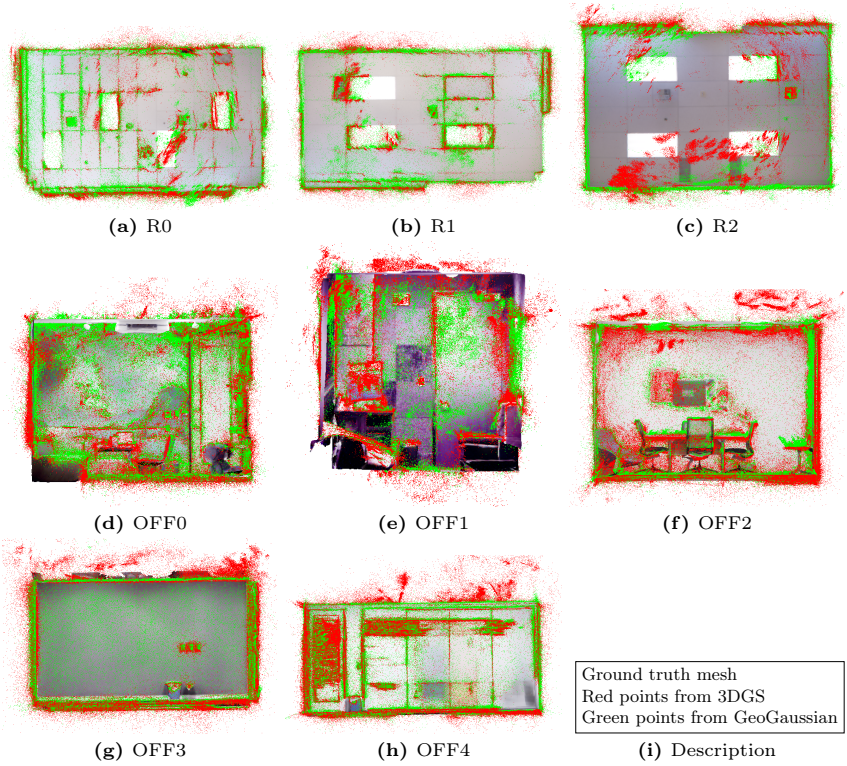

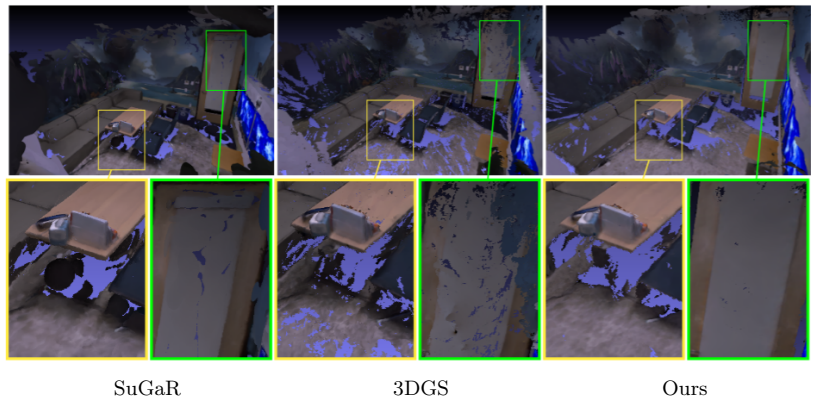
\begin{figure}
\centering
\resizebox{0.9\textwidth}{!}{
\begin{tikzpicture}[spy using outlines={red,magnification=3,size=2cm}, connect spies] 
\node (img1) at (2.5,-0.9){ \includegraphics[width=\linewidth]{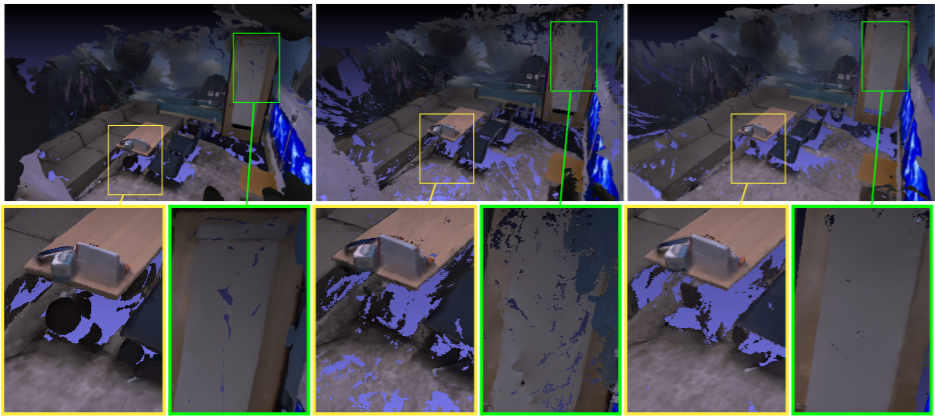}};
\node[] at (0,-4.0) {};
\node[] at (-1.45,-4.0) {SuGaR};
\node[] at (2.7,-4.0) {3DGS};
\node[] at (6.7,-4.0) {Ours};
\end{tikzpicture}}
\caption{Comparison in dense mesh modelling of Replica OFF0.
}
\label{fig:mesh_recon}
\end{figure}


As illustrated in Table~\ref{tab:sparse-replica-recon-appen}, the proposed method achieves better performance in mean and standard errors compared to 3DGS. Specifically, in the R2 sequence, the standard error of our method is $0.028$, while the corresponding value for 3DGS is $0.146$, which is $5$ times worse than ours. Additionally, in OFF2, the mean error of 3DGS is $3$ times greater than ours.

As shown in Figure~\ref{fig:reconstruction}, the point clouds from 3D Gaussians are aligned with mesh models, where the green and red points represent GeoGaussian and 3DGS, respectively. It is evident that points from our method align well with the model surface, indicating that our Gaussians are distributed according to the structure and texture of the training views. In contrast, the points from 3DGS are distributed more uniformly around the training perspective. This indicates that 3DGS is more inclined to perform densification along the angle of the training line of sight, while our method is more focused on training a highly versatile model rather than overfitting to the training perspectives.

In Figure~\ref{fig:mesh_recon}, the mesh reconstruction results of 3DGS and our method are obtained by feeding Gaussian points into the Poisson reconstruction and texturing algorithms~\cite{ye2024gaustudio}. In low-textured regions, such as door and floor areas, the proposed method demonstrates robust and accurate performance in reconstructing dense models.

\section{Performance in Wild Scenes and Limitations}

\begin{figure}
    \centering
    \includegraphics[width=0.48\linewidth]{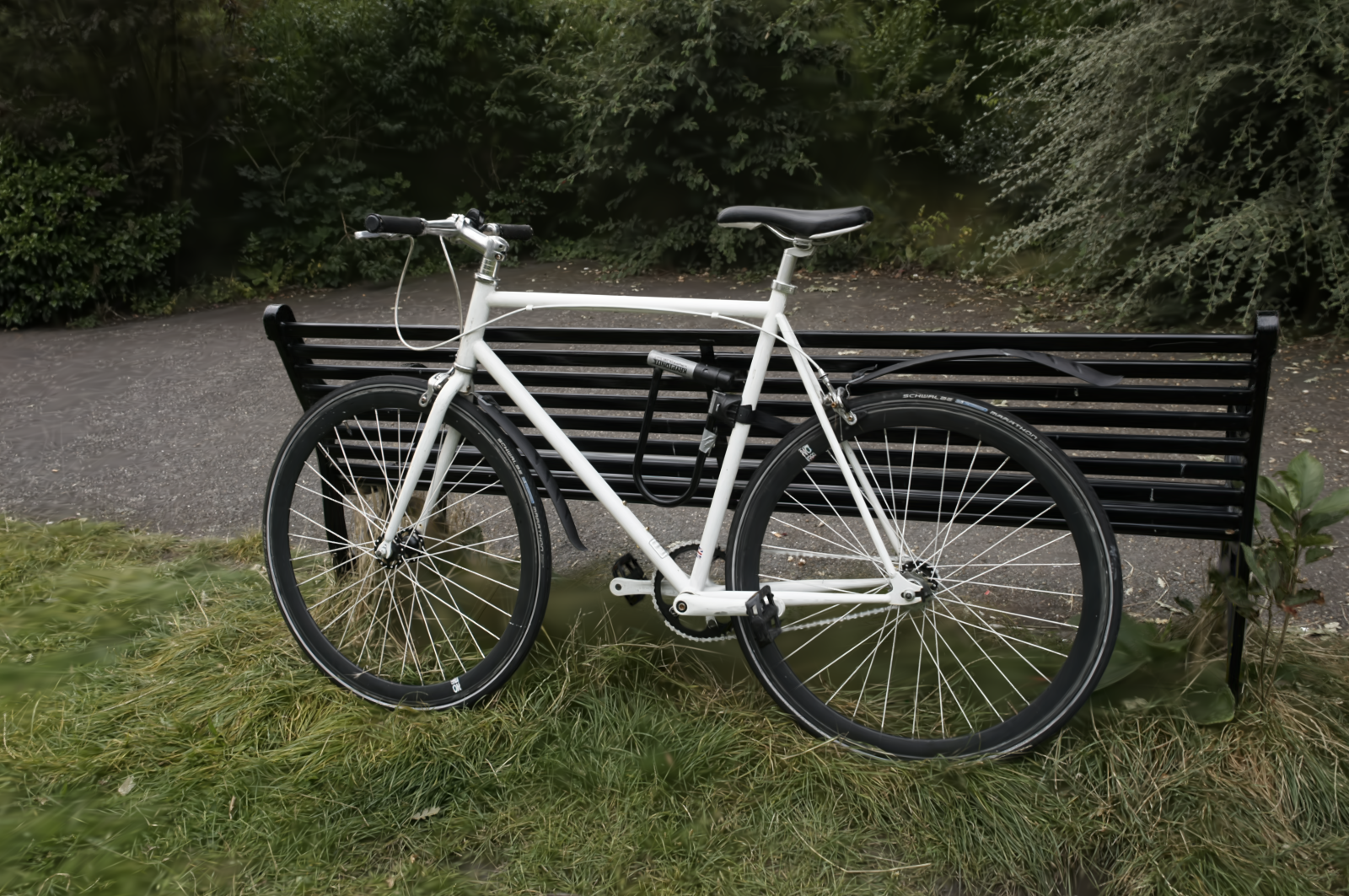}
    \includegraphics[width=0.48\linewidth]{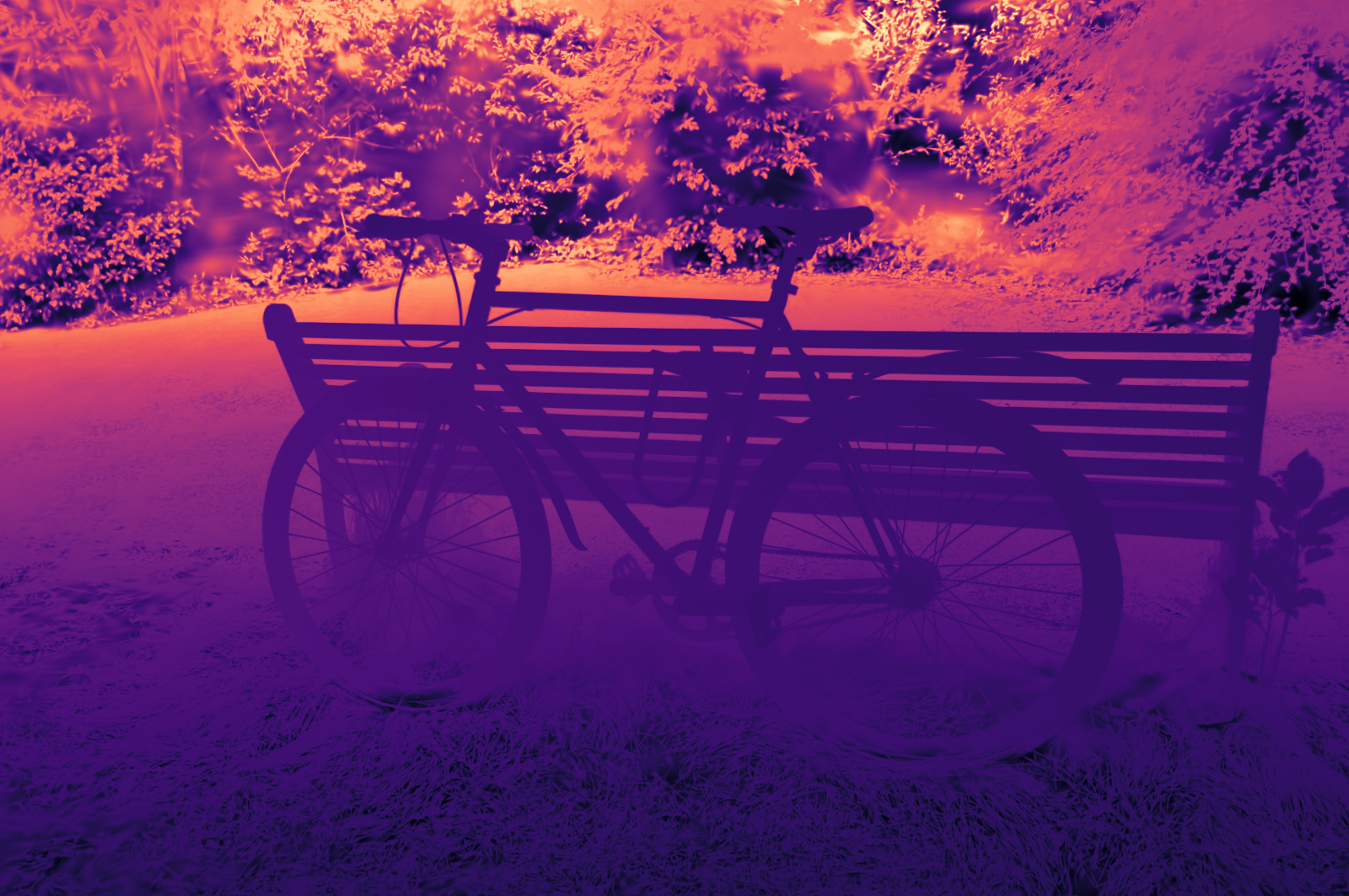}
    \caption{\small Our depth render result in Mip-NeRF360 Bicycle.}
    \label{fig:depth_render}
\end{figure}

\begin{figure}
\centering
\includegraphics[width=\linewidth]{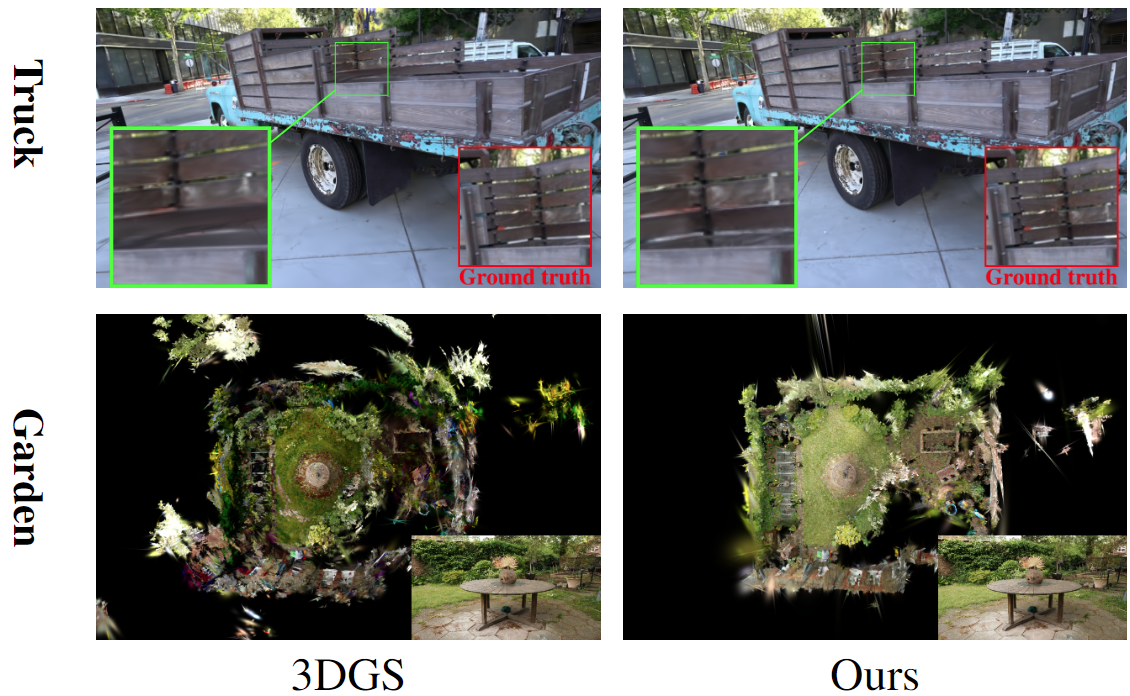}
\caption{2D and 3D results in Truck~\cite{knapitsch2017tanks} and Garden scenes~\cite{barron2022mip}. }
\label{fig:wild_scenes}
\end{figure}

To evaluate the performance of the proposed approach in wild scenes, additional popular datasets~\cite{barron2022mip,knapitsch2017tanks} are used to test GeoGaussian and 3DGS in this section. As shown in Figure~\ref{fig:depth_render} and~\ref{fig:wild_scenes}, we compare our method with 3DGS in wild scenarios, demonstrating that GeoGaussian is more robust and converges faster since our method requires only local smooth regions instead of large plane features. In 3DGS, 3D Gaussians are optimized via photometric residuals, resulting in floaters in texture-less regions. As shown in Figure~\ref{fig:wild_scenes}, while the rendering result of 3DGS is acceptable, the 3D geometry is clearly not well reconstructed. This phenomenon has been witnessed in other indoor sequences (see Figure~\ref{fig:replica-models-appen} and~\ref{fig:reconstruction}).

To solve the problem, our system introduces geometric constraints in the initialization, densification, and optimization modules. This design enhances the quality of scene geometry, leading to robust rendering performance (Figure \textcolor{red}{3}). However, there are limitations to this approach, summarized as follows: 
\begin{itemize}
    \item The geometric constraints degenerate if few smooth regions are detected; 
    \item Incorrect geometric residuals can occur when fake smooth regions are initialized as thin Gaussians, although the effect of outliers is controllable.
\end{itemize}

\section{Initialization and Densification}

\begin{figure}
     \centering
  \subfloat[ICL-NUIM Room 2]{
    \resizebox{0.48\linewidth}{!}{
   \begin{tikzpicture}
   \begin{axis}[
       xlabel=$Iterations (K)$,
       ylabel=$PSNR$,
       xmin=-0.5, xmax=11,
       ymin=15, ymax=55,
       xtick={0,1,2,3,4,5,6,7,10},
       ytick={0,10,20,30,40,50}]
   \addplot[mark=*,blue] plot coordinates {
   (0,  0.01)
     (1, 30.79)
     (2,  33.99)
     (3,  35.52)
     (4,  36.10)
     (5,  36.81)
     (6,  37.21)
     (7,  37.65)
     (10,  38.19)};
  \addlegendentry{GeoGaussian-100\%}
   \addplot[color=green,mark=x]
       plot coordinates {
       (0,  0.01)
     (1, 26.05)
     (2, 26.64)
     (3,  26.85)
     (4,  25.46)
     (5,  25.46)
     (6,  25.50)
     (7,  25.54)
     (10,  25.47)};
   \addlegendentry{GeoGaussian-10\%}
   \addplot[mark=*,brown] plot coordinates {
      (0,  0.01)
      (1, 22.91)
     (2, 32.19)
     (3,  34.34)
     (4,  35.54)
     (5,  36.17)
     (6,  36.90)
     (7,  36.96)
     (10,  36.99)};
  \addlegendentry{3DGS-100\%}
   \addplot[color=black,mark=x]
       plot coordinates {
       (0,  0.01)
     (1, 20.60)
     (2, 25.54)
     (3,  25.63)
     (4,  23.87)
     (5,  23.71)
     (6,  23.66)
     (7,  23.43)
     (10,  23.45)};
   \addlegendentry{3DGS-10\%}
   \end{axis}
   \end{tikzpicture}}}
  \subfloat[ICL NUIM Office 2]{
    \resizebox{0.48\linewidth}{!}{
   \begin{tikzpicture}
   \begin{axis}[
       xlabel=$Iterations (K)$,
       ylabel=$PSNR$,
       xmin=-0.5, xmax=11,
       ymin=15, ymax=55,
       xtick={0,1,2,3,4,5,6,7,10},
       ytick={0,10,20,30,40,50}]
   \addplot[mark=*,blue] plot coordinates {
    (0,  0.01)
     (1, 32.78)
     (2,  35.30)
     (3,  36.12)
     (4,  36.75)
     (5,  36.96)
     (6,  37.23)
     (7,  37.60)
     (10,  38.04)};
  \addlegendentry{GeoGaussian-100\%}
   \addplot[color=green,mark=x]
       plot coordinates {
    (0,  0.01)
      (1, 29.46)
     (2, 29.50)
     (3,  29.70)
     (4,  28.75)
     (5,  28.75)
     (6,  28.79)
     (7,  28.92)
     (10,  28.93)};
   \addlegendentry{GeoGaussian-10\%}
   \addplot[mark=*,brown] plot coordinates {
    (0,  0.01)
     (1, 24.32)
     (2,  32.00)
     (3,  33.73)
     (4,  34.98)
     (5,  35.70)
     (6,  36.40)
     (7,  36.50)
     (10,  36.65)};
  \addlegendentry{3DGS-100\%}
   \addplot[color=black,mark=x]
       plot coordinates {
    (0,  0.01)
      (1, 23.33)
     (2, 27.88)
     (3,  27.81)
     (4,  27.32)
     (5,  27.25)
     (6,  27.22)
     (7,  27.03)
     (10,  27.21)};
   \addlegendentry{3DGS-10\%}
   \end{axis}
   \end{tikzpicture}}}
  \caption{Rendering performance in the first 10,000 iterations.}
  \label{fig:initial-densi-appen}
 \end{figure}
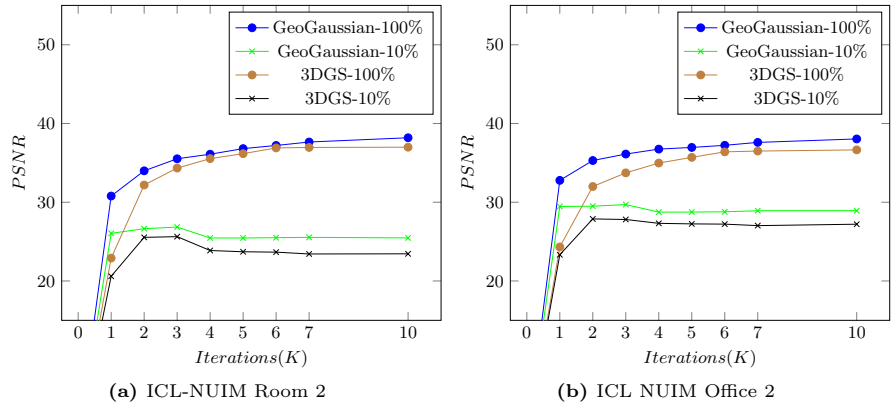   

In Figure~\ref{fig:initial-densi-appen}, the rendering performance of 3DGS and GeoGaussian approaches in the first 10,000 iterations is presented. Our method stops increasing Gaussians after 10,000 iterations, while 3DGS continues to increase Gaussians until 15,000 iterations. Therefore, the densification progress mainly occurs in the first 10,000 iterations, where Gaussians are initialized and dramatically densified under the supervision of photometric and geometric constraints.

As shown in Figure~\ref{fig:initial-densi-appen}, the PSNR performance increases rapidly in the first 10,000 iterations for both 3DGS and GeoGaussian methods. But, our method is easy to train, especially in sparse view rendering tasks like ICL-NUIM Room 2(10\%) and ICL-NUIM Office 2(10\%).

%
%
\bibliographystyle{splncs04}
\bibliography{main}

\begin{thebibliography}{10}
\providecommand{\url}[1]{\texttt{#1}}
\providecommand{\urlprefix}{URL }
\providecommand{\doi}[1]{https://doi.org/#1}

\bibitem{allene2008seamless}
Allene, C., Pons, J.P., Keriven, R.: Seamless image-based texture atlases using multi-band blending. In: 2008 19th international conference on pattern recognition. pp.~1--4. IEEE (2008)

\bibitem{barron2022mip}
Barron, J.T., Mildenhall, B., Verbin, D., Srinivasan, P.P., Hedman, P.: Mip-nerf 360: Unbounded anti-aliased neural radiance fields. In: Proceedings of the IEEE/CVF Conference on Computer Vision and Pattern Recognition. pp. 5470--5479 (2022)

\bibitem{barron2023zip}
Barron, J.T., Mildenhall, B., Verbin, D., Srinivasan, P.P., Hedman, P.: Zip-nerf: Anti-aliased grid-based neural radiance fields. arXiv preprint arXiv:2304.06706  (2023)

\bibitem{dai2017bundlefusion}
Dai, A., Nie{\ss}ner, M., Zollh{\"o}fer, M., Izadi, S., Theobalt, C.: Bundlefusion: Real-time globally consistent 3d reconstruction using on-the-fly surface reintegration. ACM Transactions on Graphics (ToG)  \textbf{36}(4), ~1 (2017)

\bibitem{debevec2023modeling}
Debevec, P.E., Taylor, C.J., Malik, J.: Modeling and rendering architecture from photographs: A hybrid geometry-and image-based approach. In: Seminal Graphics Papers: Pushing the Boundaries, Volume 2, pp. 465--474 (2023)

\bibitem{deng2020jaxnerf}
Deng, B., Barron, J.T., Srinivasan, P.P.: Jaxnerf: an efficient jax implementation of nerf. URL http://github. com/googleresearch/google-research/tree/master/jaxnerf  (2020)

\bibitem{detone2018superpoint}
DeTone, D., Malisiewicz, T., Rabinovich, A.: Superpoint: Self-supervised interest point detection and description. In: Proceedings of the IEEE conference on computer vision and pattern recognition workshops. pp. 224--236 (2018)

\bibitem{dong2018learning}
Dong, X., Dong, J., Sun, G., Duan, Y., Qi, L., Yu, H.: Learning-based texture synthesis and automatic inpainting using support vector machines. IEEE Transactions on Industrial Electronics  \textbf{66}(6),  4777--4787 (2018)

\bibitem{fan2023lightgaussian}
Fan, Z., Wang, K., Wen, K., Zhu, Z., Xu, D., Wang, Z.: Lightgaussian: Unbounded 3d gaussian compression with 15x reduction and 200+ fps. arXiv preprint arXiv:2311.17245  (2023)

\bibitem{feng2019meshnet}
Feng, Y., Feng, Y., You, H., Zhao, X., Gao, Y.: Meshnet: Mesh neural network for 3d shape representation. In: Proceedings of the AAAI conference on artificial intelligence. vol.~33, pp. 8279--8286 (2019)

\bibitem{gao2020learning}
Gao, J., Chen, W., Xiang, T., Jacobson, A., McGuire, M., Fidler, S.: Learning deformable tetrahedral meshes for 3d reconstruction. Advances In Neural Information Processing Systems  \textbf{33},  9936--9947 (2020)

\bibitem{goesele2007multi}
Goesele, M., Snavely, N., Curless, B., Hoppe, H., Seitz, S.M.: Multi-view stereo for community photo collections. In: 2007 IEEE 11th International Conference on Computer Vision. pp.~1--8. IEEE (2007)

\bibitem{guedon2023sugar}
Gu{\'e}don, A., Lepetit, V.: Sugar: Surface-aligned gaussian splatting for efficient 3d mesh reconstruction and high-quality mesh rendering. arXiv preprint arXiv:2311.12775  (2023)

\bibitem{handa2014benchmark}
Handa, A., Whelan, T., McDonald, J., Davison, A.J.: A benchmark for rgb-d visual odometry, 3d reconstruction and slam. In: 2014 IEEE international conference on Robotics and automation (ICRA). pp. 1524--1531. IEEE (2014)

\bibitem{izadi2011kinectfusion}
Izadi, S., Kim, D., Hilliges, O., Molyneaux, D., Newcombe, R., Kohli, P., Shotton, J., Hodges, S., Freeman, D., Davison, A., et~al.: Kinectfusion: real-time 3d reconstruction and interaction using a moving depth camera. In: Proceedings of the 24th annual ACM symposium on User interface software and technology. pp. 559--568 (2011)

\bibitem{karnewar2022relu}
Karnewar, A., Ritschel, T., Wang, O., Mitra, N.: Relu fields: The little non-linearity that could. In: ACM SIGGRAPH 2022 Conference Proceedings. pp.~1--9 (2022)

\bibitem{kazhdan2006poisson}
Kazhdan, M., Bolitho, M., Hoppe, H.: Poisson surface reconstruction. In: Proceedings of the fourth Eurographics symposium on Geometry processing. vol.~7, p.~0 (2006)

\bibitem{keetha2023splatam}
Keetha, N., Karhade, J., Jatavallabhula, K.M., Yang, G., Scherer, S., Ramanan, D., Luiten, J.: Splatam: Splat, track \& map 3d gaussians for dense rgb-d slam. arXiv preprint arXiv:2312.02126  (2023)

\bibitem{kerbl3Dgaussians}
Kerbl, B., Kopanas, G., Leimk{\"u}hler, T., Drettakis, G.: 3d gaussian splatting for real-time radiance field rendering. ACM Transactions on Graphics  \textbf{42}(4) (July 2023), \url{https://repo-sam.inria.fr/fungraph/3d-gaussian-splatting/}

\bibitem{kerl2013dense}
Kerl, C., Sturm, J., Cremers, D.: Dense visual slam for rgb-d cameras. In: 2013 IEEE/RSJ International Conference on Intelligent Robots and Systems. pp. 2100--2106. IEEE (2013)

\bibitem{knapitsch2017tanks}
Knapitsch, A., Park, J., Zhou, Q.Y., Koltun, V.: Tanks and temples: Benchmarking large-scale scene reconstruction. ACM Transactions on Graphics (ToG)  \textbf{36}(4),  1--13 (2017)

\bibitem{lempitsky2007seamless}
Lempitsky, V., Ivanov, D.: Seamless mosaicing of image-based texture maps. In: 2007 IEEE conference on computer vision and pattern recognition. pp.~1--6. IEEE (2007)

\bibitem{liu2020neural}
Liu, L., Gu, J., Zaw~Lin, K., Chua, T.S., Theobalt, C.: Neural sparse voxel fields. Advances in Neural Information Processing Systems  \textbf{33},  15651--15663 (2020)

\bibitem{lorensen1998marching}
Lorensen, W.E., Cline, H.E.: Marching cubes: A high resolution 3d surface construction algorithm. In: Seminal graphics: pioneering efforts that shaped the field, pp. 347--353 (1998)

\bibitem{Matsuki:Murai:etal:CVPR2024}
Matsuki, H., Murai, R., Kelly, P.H.J., Davison, A.J.: {G}aussian {S}platting {SLAM}  (2024)

\bibitem{mescheder2019occupancy}
Mescheder, L., Oechsle, M., Niemeyer, M., Nowozin, S., Geiger, A.: Occupancy networks: Learning 3d reconstruction in function space. In: Proceedings of the IEEE/CVF conference on computer vision and pattern recognition. pp. 4460--4470 (2019)

\bibitem{mildenhall2021nerf}
Mildenhall, B., Srinivasan, P.P., Tancik, M., Barron, J.T., Ramamoorthi, R., Ng, R.: Nerf: Representing scenes as neural radiance fields for view synthesis. Communications of the ACM  \textbf{65}(1),  99--106 (2021)

\bibitem{muja2009fast}
Muja, M., Lowe, D.G.: Fast approximate nearest neighbors with automatic algorithm configuration. VISAPP (1)  \textbf{2}(331-340), ~2 (2009)

\bibitem{muller2022instant}
M{\"u}ller, T., Evans, A., Schied, C., Keller, A.: Instant neural graphics primitives with a multiresolution hash encoding. ACM Transactions on Graphics (ToG)  \textbf{41}(4),  1--15 (2022)

\bibitem{niemeyer2020differentiable}
Niemeyer, M., Mescheder, L., Oechsle, M., Geiger, A.: Differentiable volumetric rendering: Learning implicit 3d representations without 3d supervision. In: Proceedings of the IEEE/CVF Conference on Computer Vision and Pattern Recognition. pp. 3504--3515 (2020)

\bibitem{oechsle2019texture}
Oechsle, M., Mescheder, L., Niemeyer, M., Strauss, T., Geiger, A.: Texture fields: Learning texture representations in function space. In: Proceedings of the IEEE/CVF International Conference on Computer Vision. pp. 4531--4540 (2019)

\bibitem{yunus2021manhattanslam}
R.~Yunus, Y.L., Tombari, F.: Manhattanslam: Robust planar tracking and mapping leveraging mixture of manhattan frames. In: 2021 IEEE international conference on Robotics and automation (ICRA) (2021)

\bibitem{rebain2021derf}
Rebain, D., Jiang, W., Yazdani, S., Li, K., Yi, K.M., Tagliasacchi, A.: Derf: Decomposed radiance fields. In: Proceedings of the IEEE/CVF Conference on Computer Vision and Pattern Recognition. pp. 14153--14161 (2021)

\bibitem{Sandstrom_2023_ICCV}
Sandstr\"om, E., Li, Y., Van~Gool, L., Oswald, M.R.: Point-slam: Dense neural point cloud-based slam. In: Proceedings of the IEEE/CVF International Conference on Computer Vision (ICCV). pp. 18433--18444 (October 2023)

\bibitem{schoenberger2016sfm}
Sch\"{o}nberger, J.L., Frahm, J.M.: Structure-from-motion revisited. In: Conference on Computer Vision and Pattern Recognition (CVPR) (2016)

\bibitem{seitz2006comparison}
Seitz, S.M., Curless, B., Diebel, J., Scharstein, D., Szeliski, R.: A comparison and evaluation of multi-view stereo reconstruction algorithms. In: 2006 IEEE computer society conference on computer vision and pattern recognition (CVPR'06). vol.~1, pp. 519--528. IEEE (2006)

\bibitem{simonyan2014very}
Simonyan, K., Zisserman, A.: Very deep convolutional networks for large-scale image recognition. arXiv preprint arXiv:1409.1556  (2014)

\bibitem{straub2019replica}
Straub, J., Whelan, T., Ma, L., Chen, Y., Wijmans, E., Green, S., Engel, J.J., Mur-Artal, R., Ren, C., Verma, S., et~al.: The replica dataset: A digital replica of indoor spaces.(2019). arXiv preprint arXiv:1906.05797  \textbf{2}(8), ~9 (2019)

\bibitem{sturm2012benchmark}
Sturm, J., Engelhard, N., Endres, F., Burgard, W., Cremers, D.: A benchmark for the evaluation of rgb-d slam systems. In: 2012 IEEE/RSJ international conference on intelligent robots and systems. pp. 573--580. IEEE (2012)

\bibitem{waechter2014let}
Waechter, M., Moehrle, N., Goesele, M.: Let there be color! large-scale texturing of 3d reconstructions. In: Computer Vision--ECCV 2014: 13th European Conference, Zurich, Switzerland, September 6-12, 2014, Proceedings, Part V 13. pp. 836--850. Springer (2014)

\bibitem{wang2023f2}
Wang, P., Liu, Y., Chen, Z., Liu, L., Liu, Z., Komura, T., Theobalt, C., Wang, W.: F2-nerf: Fast neural radiance field training with free camera trajectories. In: Proceedings of the IEEE/CVF Conference on Computer Vision and Pattern Recognition. pp. 4150--4159 (2023)

\bibitem{wang2019forknet}
Wang, Y., Tan, D.J., Navab, N., Tombari, F.: Forknet: Multi-branch volumetric semantic completion from a single depth image. In: Proceedings of the IEEE/CVF international conference on computer vision. pp. 8608--8617 (2019)

\bibitem{westoby2012structure}
Westoby, M.J., Brasington, J., Glasser, N.F., Hambrey, M.J., Reynolds, J.M.: Structure-from-motion photogrammetry: A low-cost, effective tool for geoscience applications. Geomorphology  \textbf{179},  300--314 (2012)

\bibitem{xu2019disn}
Xu, Q., Wang, W., Ceylan, D., Mech, R., Neumann, U.: Disn: Deep implicit surface network for high-quality single-view 3d reconstruction. Advances in neural information processing systems  \textbf{32} (2019)

\bibitem{voxfusion}
Yang, X., Li, H., Zhai, H., Ming, Y., Liu, Y., Zhang, G.: Vox-fusion: Dense tracking and mapping with voxel-based neural implicit representation. In: 2022 IEEE International Symposium on Mixed and Augmented Reality (ISMAR). pp. 499--507. IEEE Computer Society, Los Alamitos, CA, USA (oct 2022)

\bibitem{ye2024gaustudio}
Ye, C., Nie, Y., Chang, J., Chen, Y., Zhi, Y., Han, X.: Gaustudio: A modular framework for 3d gaussian splatting and beyond. arXiv preprint arXiv:2403.19632  (2024)

\bibitem{yugay2023gaussian}
Yugay, V., Li, Y., Gevers, T., Oswald, M.R.: Gaussian-slam: Photo-realistic dense slam with gaussian splatting. arXiv preprint arXiv:2312.10070  (2023)

\end{thebibliography}
\end{document}